\documentclass[12pt]{article}
\newcommand{\ABSTRACT}[1]{\begin{abstract}#1\end{abstract}}
\newcommand{\ACKNOWLEDGMENT}[1]{\section*{Acknowledgments}#1}
\usepackage{amsthm,amsmath,amssymb}
\theoremstyle{plain}
\newtheorem{theorem}{Theorem}
\newtheorem{lemma}{Lemma}
\newtheorem{corollary}[theorem]{Corollary}
\theoremstyle{definition}
\newtheorem{assumption}{Assumption}
\newtheorem{definition}{Definition}

\newtheorem{proposition}[theorem]{Proposition}
\newcommand{\argmax}{\operatornamewithlimits{argmax}}
\newcommand{\argmin}{\operatornamewithlimits{argmin}}
\newcommand{\ECSwitch}{\clearpage\appendix}
\newcommand{\ECHead}[1]{\noindent{\LARGE\textbf{Appendix}}}
\usepackage[margin=1in]{geometry}
\usepackage{authblk}
\allowdisplaybreaks
\usepackage{subcaption} 
\usepackage[utf8]{inputenc} %
\usepackage[T1]{fontenc}    %
\usepackage{hyperref}       %
\usepackage{url}            %
\usepackage{booktabs}       %
\usepackage{amsfonts}       %
\usepackage{nicefrac}       %
\usepackage{microtype}      %
\usepackage{amsmath, latexsym}
\usepackage{amscd}
\usepackage{amsbsy}
\usepackage{amssymb}
\usepackage{todonotes}
\usepackage{comment} 
\usepackage{shortcuts-ms} 
\usepackage[Algorithm,ruled]{algorithm}
\usepackage[noend]{algpseudocode} 
\usepackage{hyperref}
\usepackage{cleveref}
\usepackage{nicefrac}
\usepackage{graphicx}
\usepackage{adjustbox}
\usepackage{wrapfig}

\newcommand{\Bern}{\operatornamewithlimits{Bern}}

\newcommand{\norm}[1]{\left\lVert#1\right\rVert}

\newcommand\numberthis{\addtocounter{equation}{1}\tag{\theequation}}
\newcommand{\pr}{\mathbb{P}}

\newcommand{\blockedit}{}
\newcommand*{\defeq}{\mathrel{\vcenter{\baselineskip0.5ex \lineskiplimit0pt
			\hbox{\scriptsize.}\hbox{\scriptsize.}}}%
	=}
\newcommand{\ttmentindic}{\indic{T=t}}
\newcommand{\ccpscalar}{\psi}

\newcommand{\sumexpregdenom}{ \E[  \indic{T=t}W ]}

\newcommand{\suma}{\sum_{t=0}^{m-1} }
\newcommand{\Ugammand}{\mathcal{U}^\Gamma_{\mathrm{nd}} }
\newcommand{\Und}{\mathcal{U}_{\mathrm{nd}} }

\newcommand{\CR}{\mathrm{CR}}
\newcommand{\piAX}{\pi(t \mid X)}
\newcommand{\piAXbaseline}{\pi_0(t\mid X)}

\newcommand{\Ugamma}{\mathcal{U}^\Gamma }

\newcommand{\Wgamma}{\mathcal{W}^\Gamma}

\newcommand{\indt}{\mathcal{I}_t}
\newcommand{\weightuncertainty}{W}
\usepackage[sort]{natbib}
 \bibpunct[, ]{(}{)}{,}{a}{}{,}%
\begin{document}

\title{Confounding-Robust Policy Improvement}
\author[1,2]{Nathan Kallus\thanks{kallus@cornell.edu}}
\author[1,2]{Angela Zhou\thanks{az434@cornell.edu}}
\affil[1]{School of Operations Research and Information Engineering, Cornell University}
\affil[2]{Cornell Tech, Cornell University}
\renewcommand\Authands{ and }
\date{}
\maketitle\vspace{-\baselineskip}

\ABSTRACT{%
 {We study the problem of learning personalized decision policies from observational data while accounting for possible unobserved confounding. Previous approaches, which assume unconfoundedness, i.e., that no unobserved confounders affect both the treatment assignment as well as outcome, can lead to policies that introduce harm rather than benefit when some unobserved confounding is present, as is generally the case with observational data. Instead, since policy value and regret may not be point-identifiable, we study a method that minimizes the worst-case estimated regret of a candidate policy against a baseline policy over an uncertainty set for propensity weights that controls the extent of unobserved confounding. We prove generalization guarantees that ensure our policy will be safe when applied in practice and will in fact obtain the best-possible uniform control on the range of all possible population regrets that agree with the possible extent of confounding. We develop efficient algorithmic solutions to compute this confounding-robust policy. Finally, we assess and compare our methods on synthetic and semi-synthetic data. In particular, we consider a case study on personalizing hormone replacement therapy based on observational data, where we validate our results on a randomized experiment. We demonstrate that hidden confounding can hinder existing policy learning approaches and lead to unwarranted harm, while our robust approach guarantees safety and focuses on well-evidenced improvement, a necessity for making personalized treatment policies learned from observational data reliable in practice.}

\vspace{0.5\baselineskip}\noindent\textbf{History}: First version: May 22, 2018. This version: November 4, 2019.
}%

\section{Introduction}\label{intro} %
The problem of learning personalized decision policies to study ``what works and for whom'' in areas such as medicine, e-commerce, and civics often endeavors to draw insights from increasingly rich and plentiful observational data, such as electronic medical records (EMRs), since data from randomized controlled experiments may be scarce, costly, or unethical to acquire. A variety of methods have been proposed to address the corresponding problem of \emph{policy learning from observational data} \citep{k16,bl09,wager17,dell2014,kt15,kallus2017balanced,kallus2018policy}. These methods, as well as approaches to predict individual-level causal effects from observational data
\citep{wager2017estimation,nie2017learning,kunzel2017meta,shalit2017estimating}, operate under the 
controversial assumption of \textit{unconfoundedness}, which requires that the data are sufficiently informative such that no confounders that jointly affect treatment assignment and individual response are unobserved \citep{r74}, effectively requiring that assignment is \emph{as if at random} once we control for observables. This key assumption may be always made to hold \textit{ex ante} by directly controlling the treatment assignment policy as in a randomized controlled experiment, but in other domains of key interest such as personalized medicine where EMRs are increasingly being analyzed \textit{ex post}, unconfoundedness is an assumption that may never truly fully hold in fact. {Even in randomized controlled trials, in practice, challenges such as compliance, censoring, or even site selection bias may lead to confounding.}

Assuming unconfoundedness, also called \textit{ignorability}, \textit{conditional exogeneity}, or \textit{selection on observables}, is controversial because it is fundamentally unverifiable since the counterfactual distribution is never identified from the data \citep{ir15}. Thus, insights from observational studies, which passively study treatment-outcome data without intervening on treatment, are always vulnerable to this fundamental critique. 
For example, studying drug efficacy by assessing outcomes of those prescribed the drug during the course of normal clinical practice may make a drug look less clinically effective if those who were prescribed the drug were sicker to begin with and therefore would have had worse outcomes regardless. Conversely, if the drug was correctly prescribed only to the patients who would most benefit from it, it may make the drug appear to be falsely effective for all patients. These issues can potentially be alleviated by controlling for more baseline factors that may have affected treatment choices but they can never really be fully eliminated in practice.

Conclusions drawn from healthcare databases such as 
claims data are particularly vulnerable to unobserved confounding because although they record administrative interactions and diagnostic codes, they are uninformative about medical histories, notes on patient severity, observations, nor monitoring of clinical outcomes, \textit{i.e.}, the key clinical information which may drive a physician's treatment choices. EMRs provide great promise for enabling richer personalized medicine from observational data because they record the entire patient treatment and diagnostic history, past medical history and comorbidities, as well as fine-grained information regarding patient response such as vital signs
\citep{hoffman2011electronic}.
The growing adoption of richer EMRs can both provide higher precision for personalized treatment and render unconfoundedness more plausible, since the data includes more of the information regarding patient history and outcomes that informs physician decision-making, yet unconfoundedness, an ideal stylized assumption, still may never be fully satisfied in practice.

{
The challenges of observational data are of course not new to the modern era of data-driven decision-making, but have been widely recognized. 
One high-profile example is the case of the parallel WHI observational study and clinical trial, which illustrates how confounding factors can lead to dramatic discrepancies in drawing clinically relevant prescriptions from randomized trial versus observational data. The parallel WHI observational study and clinical trials studied whether hormone replacement therapy (HRT) had therapeutic benefits for chronic disease prevention.
While HRT was known to be clinically effective for vasomotor symptoms of menopause, 
earlier observational epidemiological studies additionally suggested a protective benefit against coronary heart disease (CHD) which lead to the increasing clinical practice of prescribing HRT in menopause for preventive purposes (without clinical trial evidence) \citep{to-whi-03} . The parallel WHI observational study and clinical trial were designed to evaluate the efficacy of HRT in a preventive context on chronic disease, such as coronary heart disease (CHD) and breast cancer, among other clinical endpoints. 
Ultimately, the WHI clinical trial dramatically repudiated these purported therapeutic benefits. In fact, while the observational study suggested a protective benefit of HRT against CHD, showing a 40-50\% reduction in CHD incidence, the HRT arm of the clinical trial had to be stopped early due to a dangerously elevated incidence of CHD \citep{prentice-whi-05}. After the WHI study, the new evidence that arose not only dramatically changed the standard of care, spurring an 80\% reduction in the prescription of HRT, but also sparked a broader methodological debate about the clinical credibility of observational studies \citep{Lawlor04}. Later in \Cref{sec-whi}, we build a case-study with semi-synthetic data from the observational study and clinical trial to illustrate potential harms of policy learning from realistically confounded data. This case study, as well as others, illustrate the challenges of unobserved confounders that would continue to plague richer data-driven decision-making strategies such as personalized policy learning
}

Because unconfoundedness may fail to hold, existing policy learning methods that operate under this assumption can lead to personalized decision policies that seek to exploit individual-level effects that are not really there, may intervene where not necessary, and may in fact lead to net harm rather than net good. 
Such dangers constitute obvious impediments to the use of policy learning to enhance decision making in such sensitive applications as medicine, public policy, and civics, where reliable and safe algorithms are critical to implementation. 
Clearly, a policy that could potentially introduce additional harm, toxicity, or risk to patients compared to current standards of care is an unacceptable replacement, and an algorithm that could potentially give rise to such a policy is unusable in medical and other sensitive settings. { Furthermore, unobserved confounders remain a key issue for data-driven decision-making in e-commerce; for example, a randomized trial of the effectiveness of search ads on eBay \citep{blake2015consumer} revealed the spurious efficacy of advertising, based on observational studies of user search queries, which did not account for unobserved intent or customer loyalty. }

To address this deficiency, in this paper we develop a framework for confounding-robust policy learning and improvement that can ensure that the personalized decision policy derived from observational data, 
which inevitably will have \emph{some} unobserved confounding, 
will do no worse than a current policy such as the current standard of care and, in fact, will do better if the data can indeed support it. We do so by recognizing and accounting for the potential confounding in the data and requiring that the learned policy improve upon the baseline no matter the direction of confounding. Thus, we calibrate personalized decision policies to address sensitivity to realistic violations of the unconfoundedness assumption. For the purposes of informing reliable and personalized decision-making that leverages modern machine learning, our work highlights that statistical point identification of individual-level causal effects, which previous approaches crucially rely on, may not at all be necessary for successfully learning effective policies that reliably improve on unpersonalized standards of care, but accounting for the lack of point identification is necessary. 

Functionally, our approach is to optimize a policy to achieve the best worst-case improvement relative to a baseline treatment assignment policy (such as treat all or treat none), where the improvement is measured using a weighted average of outcomes and weights take values in an uncertainty set around the nominal inverse propensity weights (IPW). 
This generalizes the popular class of IPW-based approaches to policy learning, which optimize an unbiased estimator for policy value under unconfoundedness \citep{lwlw11,sj15,js15norm,bl09,kt15}.
Unlike standard approaches,
in our approach the choice of baseline is material and changes the resulting policy chosen by our method.
This framing supports reliable decision-making in practice, as often a practitioner is seeking evidence of substantial improvement upon the standard of care or a default option, and/or the intervention under consideration introduces risk of toxicity or adverse effects and should not be applied without strong evidence.

{
Our contributions are as follows. We provide a framework for performing \textit{policy improvement} that is robust in the face of unobserved confounding by using a robust optimization formulation. 
Our framework allows for the specification of data-driven uncertainty sets based on a sensitivity parameter 
describing a pointwise bound on the odds ratio between true and observed propensities as well as uncertainty sets with a global budget-of-uncertainty parameter. 
Whereas previous approaches for sensitivity analysis in causal inference focus on evaluating the \textit{range} of inferential procedures (e.g. effect estimation or hypothesis tests), we focus on the question of learning \textit{optimal decision policies} in the presence of unmeasured confounding. Sensitivity models in causal inference, which introduce ambiguity in the space of inverse propensity weights which does not vanish with increasing data. Thus, learning decision policies under sensitivity models introduces analytical challenges in ensuring convergence. 
}
{
We prove a uniform convergence result both over the space of policies of restricted complexity and over the possible confounded data-generating distributions in our uncertainty set: therefore, our approach is asymptotically optimal for the population minimax regret. These results also imply an
appealing improvement guarantee that shows that, up to
vanishing factors that depend
on the complexity of the policy class, 
our approach will not do worse than the baseline
and, moreover, will do better, as can be easily validated by simply evaluating the objective value of our optimization problem.
Leveraging the structure of our optimization problem and characterizing the inner subproblem, we provide a set of efficient algorithms for performing robust policy optimization over parameterized policy classes and over decision trees. 
We assess performance on a synthetic example that illustrates the benefits of our approach and the effect of the uncertainty parameters.
We then show, in a case study drawing on the unique simultaneous WHI observational study and clinical trial, that in regimes with realistic confounding, for a variety of possible treatment effect profiles, our approach can lead to improvement upon a baseline while learning from confounded data causes harm. This case study allows us to uniquely learn from observational data with unobserved confounding, yet assess out of sample performance on an unconfounded clinical trial. 
}

\section{Problem Statement and Preliminaries}\label{probstatementsec}
{ 
We first summarize the setup. 
	We consider policy learning from observational data consisting of tuples of random variables $\{(X_i,T_i,Y_i):i=1,\dots,n\}$, comprising of covariates $X_i \in \mathcal{X}$, assigned treatment level out of $m$ discrete treatments $T_i \in \{0, \dots, m-1\}$, and real-valued outcomes $Y_i\in\mathbb R$. 
	We suppose that these constitute iid (independent and identically distributed) observations from a population and we drop subscripts to denote a generic draw from this population.
	We allow $m\geq2$, so that we accommodate the case of multiple, discrete treatment levels. 
	We let $Y_i(0),\dots,Y_i(m-1)$ denote the potential outcomes of applying each treatment option, respectively, and we assume that $Y_i=Y_i(T_i)$ so that the observed outcome corresponds to the potential outcome of the observed treatment.}\footnote{The equation $Y_i=Y_i(T_i)$ captures two important 
	features. 
	One is that the observed outcomes are consistent with the 
	hypothetical potential outcomes. Another is that the outcome of an 
	individual only depends on the treatment assignment of that individual
	and there is no interference between units. This two assumptions 
	together are also known as the \emph{stable unit treatment value assumption} \citep{rubin1980randomization}.}.
{We let $\E_n$ denote the empirical expectation, i.e. taking a sample average over the data. We define the index set for treatment value $t$ as $\mathcal I_t = \{ i\leq n: T_i = t \}$. }
{
We use the convention that the outcomes $Y_i$ corresponds to losses so that lower outcomes are better.}

{We denote the \emph{nominal} propensity function by $\tilde e_t(x)=\Prb{T=t\mid X=x}$ and the nominal generalized propensity score by $\tilde e_{T_i}(X_i)$. This can be estimated directly from the data using a probabilistic classification model such as logistic regression or a neural network. When it is estimated, we denote the estimated nominal propensity function by $\hat{e}_t(x)$. Since we do not assume unconfoundedness, the nominal propensity is insufficient to account for confounding. We therefore additionally define the \emph{true} propensity function as $e_t(x,y)=\Prb{T=t\mid X=x,Y(t)=y}$ and the true generalized propensity score as $e_{T}(X,Y)$. Note that these \emph{cannot} be estimated from the data. Unconfoundedness (weak ignorability) is the assumption that $\tilde e_t(x)=e_t(x,y)$ as functions (i.e., $\indic{T=t}\indep Y(t)\mid X$). Here, we do \emph{not} assume unconfoundedness and will generally have that $e_t(x)\neq e_t(x,y)$.}

{
	We consider evaluating and learning a (potentially) randomized policy mapping covariates to the probability of assigning treatment, $\pi: \mathcal{X}\to \Delta^{m}$, where $\Delta^m$ denotes the $m$-simplex. Given a policy $\pi$, we use the notation $\pi(t\mid x)$ to denote the probability $\pi$ assigns to treatment $t$ when observing covariates $x$. {It is also convenient to also define the random treatment variable $Z^\pi$ that, given $X$, is independent of all else, and has the distribution $\Prb{Z^\pi=t\mid X}=\pi(t\mid X)$.} { 
The policy value of $\pi$ is
$V(\pi)=\Eb{\sum_{t=0}^{m-1}\pi(t\mid X_i)Y(t)}=\Eb{Y(Z^\pi)}$. 
}
	{As is common for policy learning \citep[e.g.,][]{kallus2017balanced,wager17}, we focus on a restricted policy class $\Pi\subseteq[\mathcal{X}\to \Delta^{m}]$.}
{Examples 
	include 
	deterministic linear policies, 
	$\pi_{\alpha_{0:m-1},\beta_{0:m-1}}(t(x)\mid x)=1$ where
	$t(x)\in\argmax_{t=0,\dots,m-1}\alpha_t+\beta_t^\intercal x$;
	logistic policies, $\pi_{\alpha_{0:m-1},\beta_{0:m-1}}(t\mid x)\propto \exp(\alpha_t+\beta_t^\intercal x)$;
	or decision trees of a bounded depth,
	which assign any probability vector to each leaf of the tree. }}

\section{Related Work}\label{sec:relatedwork}

Our work builds upon several strands of literatures, notably policy learning from observational data as well as sensitivity analysis in causal inference. 

\textbf{Causal inference for personalization from observational data under unconfoundedness. }
The key difficulty in learning \textit{interventional} effects from observational data is that the outcome $Y_i(T_i)$ is only observed for the treatment actually administered historically to the unit, $T_i$, whose assignment can itself be correlated with the potential outcomes, obfuscating differences in them. Since the data is observational and the treatment assignment procedure was not under the control of the experimenter, the distribution of covariates may be systematically different between treatment and control groups due to self selection of the individuals into treatments, medical imperatives trading off treatment risk vs. patient severity, or business imperatives to offer discounts or target advertising not completely at random. Thus, the systematic differences in covariates in the population $\Prb{X=x,Y=y\mid T=1},\,\Prb{X=x,Y=y\mid T=0}$, also known as \textit{covariate shift}, make the treated and untreated populations incomparable for the purpose of assessing effect.

{When \emph{all} covariates needed to ensure unconfoundedness are assumed to be observed, \textit{i.e}, {$\tilde e_t(x)=e_t(x,y)$}, then a variety of approaches for learning personalized intervention policies that maximize causal effect have been proposed. These fall under regression-based strategies \citep{qian2011performance,bertsimas2016personalized}, reweighting-based strategies \citep{bl09,k16,kt15,sj15}, or doubly robust combinations thereof \citep{dell2014,wager17}. {Regression-based strategies estimate the conditional average outcomes, $\E[Y(t)\mid X]$, which under unconfoundedness are equal to $\E[Y\mid X,T=t]$, a regression of outcome on covariates in the $t$-treated group. These estimates are either used directly to treat by picking the smallest value (known as \emph{direct comparison}) or to score policies and pick the best in a restricted class (known as the \emph{direct method}). For binary treatments, we can directly fit the difference $\E[Y(1)-Y(0) \mid X]$, known as the conditional average treatment effect \citep{wager2017estimation}. If the regression functions are} ill-specified, we are not guaranteed to find the best policy, even if the class is amenable to the estimation method (\textit{e.g.}, the best linear policy does not arise from comparing the best linear CATE estimator to zero). 
{Without unconfoundedness, the regression functions or CATE are not identifiable from the data (parametrically or non-parametrically) and these methods have no guarantees.}

Reweighting-based strategies use 
inverse propensity weighting (IPW) \citep{bl09,k16,sj15,kt15}
or covariate-balancing weights \citep{kallus2017balanced}
to change measure from the distribution induced by a historical logging policy to that induced by any new policy $\pi$.
Specifically, these methods use the fact \citep{lwlw11} that, under unconfoundedness,
$\hat{V}^\text{IPW}(\pi)$ is unbiased for $V(\pi)$, where
\begin{equation}\label{Vipw}
{\hat{V}^\text{IPW}(\pi; \tilde{e}_T)=  \suma \E_n\left[  \frac{  \piAX  \ttmentindic Y}{\tilde{e}_t(X)}
\right]}
\end{equation}
Optimizing $\hat{V}^\text{IPW}(\pi)$ for deterministic policies can be phrased as a weighted classification problem \citep{bl09}.
\citet{dell2014} suggest to augment eq.~\eqref{Vipw} by using
the doubly-robust estimator \citep{robins1994estimation}, which centers the outcomes using a regression estimate. \citet{wager17} show that since this estimate is semiparametrically efficient when using cross-fold fitting, as shown by \citet{chernozhukov2016double}, this leads to better regret bounds.
Since dividing by propensities can lead to extreme weights and high variance estimates,
clipping the probabilities are typically necessary for good performance \citep{js15norm,wad2017} or the use of weights that directly optimize for balance \citep{kallus2017balanced}.
{With or without any of those fixes, if there are unobserved confounders, then,
 neither a policy's value nor the optimal policy are identifiable, and any of these methods may lead to learned policies that may well introduce more harm than good.} 
Under unconfoundedness, such reweighting-based methods are notable for 
being able to find best-in-class policies regardless of specification
of an outcome model 
\citep[or with outcome models learned at 
sub-parametric rates;][]{wager17}.
Specifically, they focus directly on the policy learning problem
rather than a prediction problem and 
on finding a policy that performs as the
best in a given class. This leads to strong
generalization guarantees
\citep{kt15,kallus2017balanced,wager17} and can also allow one
to incorporate domain-specific constraints that favor simple prescriptive decision policies that are interpretable, implementable, and/or satisfy operational constraints, such as scorecards or decision-trees \citep{ur15}. These constraints and approaches for training optimal constrained policies can be composed directly with the policy optimization problem by restricting the policy class. 
Because of these unique properties, our approach will also be
based on a reweighting approach that directly optimizes a policy
rather than a predictor.

{The literature on optimal policy learning in econometrics has also considered a minimax regret criterion. 
	\cite{manski08,manski05} consider the optimal decision policy obtained by minimax regret bounds on conditional average outcomes, which arise from partial identification bounds on arbitrary confounding from the unidentified counterfactual probabilities: this approach is highly conservative and does not use available information on selection based on observables (namely, $\tilde{e}_t(x)$ which exists despite additional unobserved confounding.). \cite{stoye12,stoye09} consider minimax regret from a decision-theoretic point of view, where a closed form is available under limiting asymptotic assumptions on an experimental sampling design generating treatment assignments.
}
{
In contrast to these lines of work, we focus on handling the commonplace yet detrimental issue of
unobserved confounding by designing an uncertainty set around the inverse propensity weights, while our minimax approach is data-driven and assesses \textit{reasonable} violations of unconfoundedness.}

\textbf{Policy improvement. }
A separate literature within reinforcement learning, unrelated to causal inference, considers the idea of safe policy improvement by forming an uncertainty set around the presumed unknown transition probabilities between states as in \citet{ttg15} or forming a trust region for safe policy exploration via concentration inequalities on  estimates of policy risk as in \citet{pgc16}. None of these consider the issue of confounding or observational data. 
This general approach of safely improving upon 
another policy using a robust or minimax formulation
is related to, and inspires the name of, our method.

\textbf{Sensitivity analysis. }
{Sensitivity analysis in causal inference provides means to test the robustness of inferences about an average treatment effect made based on observational data to assumptions such as unconfoundedness. In contrast, our work focuses on personalized policy learning in the presence of unobserved confounding from an infinite family of potential policies.}
{Some approaches from sensitivity analysis} for assessing unconfoundedness require auxiliary data or additional structural assumptions, which we do not assume here \citep{ir15}. 
Other approaches consider how large confounding must be to invalidate the conclusions of statistical inference, and consider assumptions restricting the strength of unobserved confounding on the selection process, or on the outcome model. under the Rosenbaum's sensitivity model \citet[Ch. 4]{r02} for hypothesis testing in randomization inference. 
{ For example, sensitivity analysis would assess the range of extremal $p$-values 
on the hypothesis of no effect for randomization inference, 
depending on the value of $\Gamma$
so that consequent binary conclusions can be couched in terms
of the level of unobserved confounding required to overturn a nominal conclusion
\citep{r02,fogarty2016sensitivity,hasegawa2017sensitivity}.}
{Our approach borrows the marginal sensitivity model from sensitivity analysis \citep{tan}, assuming bounds on the strength of unobserved confounding on selection into treatment, and focuses on the implications for personalized treatment decisions.}

The Rosenbaum model for sensitivity analysis assesses the robustness of randomization inference to the presence of unobserved confounding by considering a uniform bound $\Gamma$ on the \textit{odds ratio} between $e_t(x,y)$ and $e_t(x,y')$, i.e., between the treatment propensities of any two units with equal covariates \citep{r02}. 
{The closely related \textit{marginal} sensitivity model, introduced by \cite{tan}, considers a uniform bound $\Gamma$ on the odds-ratio between the \textit{nominal} propensity $e_t(x)$ and the true propensity $e_t(x,y)$. \cite{zhaosmall17} provides further discussion on the relationship between the two sensitivity models. They are generally different and incomparable for equal values of $\Gamma$.}
The value of $\Gamma$ can be calibrated against the discrepancies induced by omitting observed variables; then determining $\Gamma$ can be phrased in terms of whether one thinks one has omitted a variable that could have increased or decreased the probability of treatment by as much as, say, gender or age can in the observed data \citep{hsu2013calibrating}. 

In the sampling literature, 
the H\'ajek estimator
for population mean \citep{hajek1971comment} is an extension of
the classic Horvitz-Thompson estimator \citep{ht52}
that adds weight normalization. The objective of the minimax game we define between policy optimizer and possible confounding is a H\'ajek estimator for the policy value.
\citet{aronowlee12} derive sharp bounds on the estimator arising from a uniform bound on the sampling weights, showing a closed-form for the solution for a \textit{uniform} bound on the sampling probabilities. \citet{zhaosmall17} consider bounds on the H\'ajek estimator, but impose a parametric model on the treatment assignment probability. \citet{mwz18} consider tightening the bounds from the H\'ajek estimator by adding shape constraints, such as log-concavity, on the cumulative distribution of outcomes. \citet{mp18} consider sup-norm
bounds on propensity differences and show sharp partial identification of bounds for CATE  and ATE by integrating partially identified bounds on the conditional quantile treatment effect. 
{ In contrast to the sensitivity analysis literature in causal inference, we focus on the implications of sensitivity analysis for learning a robust personalized policy function which poses additional analytical challenges in ensuring convergence of data-driven robust policies.}

\section{Robust policy evaluation and improvement}\label{sec-estimator}

We now present our framework for confounding-robust policy improvement. Our approach minimizes a bound on policy regret against a specified baseline policy $\pi_0$, $R_{\pi_0}(\pi) = V(\pi) - V(\pi_0)$. Our bound is achieved by maximizing a reweighting-based regret estimate over an uncertainty set around the nominal propensities. This ensures that we cannot do any worse than $\pi_0$ and may in fact do better, even if the data is confounded.

The baseline policy $\pi_0$ can be any fixed policy that we want to make sure not to do worse than or deviate from unnecessarily. This is usually the current standard of care, established from prior evidence, and we would not want any algorithmic solution to personalization to introduce any harm relative to current standards. {Generally, this is the policy that always assigns control, $\pi_0(0\mid X)=1$.}
Alternatively, if reliable clinical guidelines exist for 
some limited personalization, then $\piAXbaseline$ represent the non-constant
function that encodes these.

\subsection{Confounding-robust policy learning by optimizing minimax regret}\label{sec:optminimaxregret}

{If we had oracle access to the true inverse propensities $W_i^*=1/{e}_{T_i}(X_i,Y_i)$ we could form the correct IPW estimate by replacing nominal with true propensities in eq.~\eqref{Vipw}. 
We may go a step further and, recognizing that $\E[W_i^* \ttmentindic]=1$, use the empirical sum of true propensities as a control variate by normalizing our IPW estimate by them. 
This gives rise to the H\'ajek regret estimator
		\begin{align*}
\hat{R}^*_{\pi_0}(\pi)&=\hat{R}_{\pi_0}(\pi;W^*),\qquad\text{where}\\
\hat{R}_{\pi_0}(\pi;W)&=\suma\hat{R}^{(t)}_{\pi_0}(\pi;W),\quad \hat{R}^{(t)}_{\pi_0}(\pi;W)=\frac{\E_n [(\piAX - \piAXbaseline) \ttmentindic Y W  ] }{   \E_n [  W \ttmentindic ] }
\end{align*}
These estimators introduce the denominator $\E[W_i^* \ttmentindic]$ as a ratio control variate within each treatment group. It follows by Slutsky's theorem that these estimates remain consistent 
(\emph{if} we know $W_i^*$).
Note that the choice of $\pi_0$ amounts to a constant shift to $\hat{R}^*_{\pi_0}(\pi)$ and does not change which policy $\pi$ minimizes the regret estimate. This will not be true of our bound, where the 
choice of $\pi_0$ will be material to the success of the method.}

Since the oracle weights $W_i^*$ are unknown, we instead minimize 
the worst-case possible value of our regret estimate, by
ranging over the space of possible values {for $W_i^*$ that} are consistent with the observed data and our assumptions about the confounded data-generating process. 
Specifically, we restrict the extent to which unobserved confounding may affect assignment probabilities.

{\blockedit
{We first consider an uncertainty set motivated by the odds-ratio bounds of the marginal sensitivity model,
which restricts how far the weights can vary pointwise from the nominal propensities \cite{tan}.} {Given a bound $\Gamma\geq1$, the marginal sensitivity model posits the following restriction: 
\begin{equation}\label{oddsratio}
\Gamma^{-1}\leq
\frac{ (1-\tilde{e}_T(X) ) e_T(X, Y)  }{ \tilde{e}_T(X)  (1-e_T(X,Y))  } 
\leq \Gamma.
\end{equation}
The choice of $\Gamma$ can be calibrated using, \textit{e.g.}, the method of \citet{hsu2013calibrating}, and we discuss other approaches in \Cref{apx:rbars}. Note that $\Gamma=1$ corresponds to unconfoundedness (weak ignorability and $\Gamma=\infty$ to no restriction at all.
}

{
The restriction in eq.~\eqref{oddsratio} 
leads to an uncertainty set 
for the true inverse propensity weights of each unit
centered around the nominal inverse propensity weights, $\tilde W_i = {1}/{\tilde{e}_{T_i}(X_i)}$:}
\begin{align} \label{uncertaintyset1eq}
W_{1:n}^*\in\mathcal{W}_n^{\Gamma} &= \left\{ W \in \R n \colon  a_{i}^\Gamma  \leq W_i \leq   b_{i}^\Gamma,\;~\forall i=1,\dots,n\right\}
,~~\text{where}\\\notag
a_{i}^\Gamma&=
1+\Gamma^{-1}(\tilde W_i - 1)
,~
b_{i}^\Gamma=
1+\Gamma(\tilde W_i - 1).
\end{align}
We assume for now that $\tilde W_i$ is known and phrase our method in terms of it. In practice, when $\tilde e_t(x)$ is unknown, we suggest to estimate it (e.g., using regression) and plug in the corresponding estimates of $\tilde W_i$ in their place. In \Cref{sec:estprop}, we will show that this approach is asymptotically equivalent and provide explicit finite-sample bounds.

{
Given this uncertainty set, we obtain the following bound on the empirical regret H\'ajek estimator:
}
\begin{equation}\label{rbareq-prod}
\hat{\overline{R}}_{\pi_0}(\pi;\mathcal{W}_n^\Gamma) 
=\sup_{W \in \mathcal{W}_n^{\Gamma}}\hat R_{\pi_0}(\pi;W).
\end{equation}
We then propose to choose the policy $\pi$ in our class $ \Pi$ to minimize this regret bound,
\textit{i.e.}, $\hat{\overline{\pi}}(\Pi,{\mathcal{W}}_n^\Gamma,\pi_0)$, where
\begin{equation}\label{pb-opt-pol} \hat{\overline{\pi}}(\Pi,{\mathcal{W}}_n^\Gamma,\pi_0) \in \argmin_{\pi \in \Pi}\;
	\hat{\overline{R}}_{\pi_0}(\pi;{\mathcal{W}}_n^\Gamma)
	\end{equation}

{We emphasize that different components of the framework such as weight normalization and estimation error change the population-optimal confounding-robust policy, in contrast to the policy learning setting with unconfoundedness, where these components only affect finite-sample considerations. In particular, for our worst-case regret objective $\hat{\overline{R}}_{\pi_0}(\pi;{\mathcal{W}}_n^\Gamma)$, weight normalization is crucial for only enforcing robustness against \emph{consequential} realizations of confounding that affect the \textit{relative} weighting of outcomes. Any mode of the confounding that affects all weights similarly should have no effect on policy choice. Even if we do not know 
$W_i^*$, we know that they must satisfy the population moment conditions
$\E[W^* \mathbb{I}[T = t]]=1, \forall t \in \mathcal T $, so any realization that violates that is impossible.
Moreover, different baseline policies $\pi_0$ structurally change the solution to the adversarial subproblem by shifting the contribution of the loss term $Y_i \mathbb{I}[T_i = t](\pi(T_i \mid X_i)-\pi_0(T_i \mid X_i))$ to emphasize improvement upon different baselines. In particular, if the baseline policy is in the policy class $\Pi$, it already achieves 0 regret; thus, minimizing regret necessitates learning a policy that \emph{must} offer some benefits in terms of decreased loss regardless of confounding.
}

\subsection{The population minimax-optimal policy}

In the above, we proposed to minimize an upper bound on an estimate for the policy regret. We can also similarly define a population-level bound and consider the population-level minimax-optimal policy. Specifically, we can translate the marginal sensitivity model, eq.~\eqref{oddsratio}, to an uncertainty set about the population random variable $W^*=1/e_T(X,Y)$:
\begin{align*}
\mathcal W^\Gamma&=\{W(t,x,y) \colon  a_t^\Gamma(x) \leq W(t,x,y)\leq b_t^\Gamma(x)~~\forall t\leq m-1, x \in \mathcal{X}, y\in \Rl \},~~\text{where}\\
\textstyle	a_{t}^\Gamma(x)&=
	1+\Gamma^{-1} (\nicefrac{1}{\tilde{e}_t(x)} - 1)
	,~
	b_{t}^\Gamma(x)=
	1+\Gamma(\nicefrac{1}{\tilde{e}_t(x)} - 1).
\end{align*}
Notice that $\mathcal W_n^\Gamma=\{(W(T_1,X_1,Y_1),\,\dots,\,W(T_n,X_n,Y_n):W\in \mathcal W^\Gamma\}$ can be understood as the restriction of the above to the data.
The corresponding bound on the population-level regret is $\overline{R}_{\pi_0}(\pi; \mathcal{W}^\Gamma)$, where
\begin{align*}
\overline{R}_{\pi_0}(\pi; \mathcal{W})&=\sup \left\{{R}_{\pi_0}(\pi; W) \colon W \in  \mathcal{W} \right\},\qquad\text{where}\\
{R}_{\pi_0}(\pi; W)&=\suma {R}^{(t)}_{\pi_0}(\pi; W),\quad
{R}^{(t)}_{\pi_0}(\pi; W)=
\frac{\E[ \indic{T=t}  (\piAX - \piAXbaseline)W(T,X,Y) Y] }{\E[\indic{T=t} W(T,X,Y)  ] }.
\end{align*}
Note that $R_{\pi_0}(\pi)=R_{\pi_0}(\pi;W^*)$.
In words, $\overline{R}_{\pi_0}(\pi; \mathcal{W}^\Gamma)$ is the largest-possible true regret of $\pi$ relative to $\pi_0$ over all possible distributions that agree with the observable data-generating distribution of $(X,T,Y)$ \emph{and} with the restrictions of the marginal sensitivity model. That is, every potentially-possible regret of $\pi$ is bounded by this quantity and this quantity is also tight in that there exist distributions agreeing with the data and the assumptions that are arbitrarily close to it. The denominator in ${R}_{\pi_0}(\pi; W)$ ensures that we adhere to the requirement that $\E[\indic{T=t} W^*]=1$.\footnote{{As an uncertainty set over the joint distribution $\Prb{T,X,Y(0),\dots,Y(m-1)}$ this would correspond to $\braces{\mathbb P\colon\psi_t/b_t^\Gamma(x)\leq\Prb{T=t\mid X=x,Y(t)=y}\leq\psi_t/a_t^\Gamma(x)~\forall t\leq m-1,\,\psi\in\R m_{+},\,\mathbb P\text{ is a probability distribution}}$.}}

In fact, the interval generated by the smallest-possible and largest-possible regret is sharp in that it is equal to the closure of all possible regrets under the marginal sensitivity model. We summarize this side observation as follows:
\begin{proposition}[Sharpness]\label{prop-sharpness}
$$\textstyle
\overline{\left\{{R}_{\pi_0}(\pi; W) \colon W \in  \mathcal{W}^\Gamma\right\}}=\left[\inf_{W\in\mathcal{W}^\Gamma}{R}_{\pi_0}(\pi; W),\,\sup_{W\in\mathcal{W}^\Gamma}{R}_{\pi_0}(\pi; W)\right].
$$
\end{proposition}

We can correspondingly conceive of what would be the minimax-optimal policy at the population level, \textit{i.e.}, ${\overline{\pi}}^*(\Pi,{\mathcal{W}}^\Gamma,\pi_0)$, where
\begin{equation}\label{eq:minimaxoptpop} {\overline{\pi}}^*(\Pi,{\mathcal{W}^\Gamma},\pi_0) \in \argmin_{\pi \in \Pi}\;
	{\overline{R}}_{\pi_0}(\pi;{\mathcal{W}^\Gamma})
\end{equation}
This minimax-optimal policy is the one that would obtain the best-possible uniform control over all possible regrets under any possible realization of the true distribution of outcomes that agrees with the observable data-generating distribution of $(X,T,Y)$ \emph{and} with the restrictions of the marginal sensitivity model. This feature makes it an attractive target to aim for in the absence of unconfoundedness.

\subsection{Extension: Budgeted uncertainty sets to address ``local'' confounding}
{ 
Our approach can flexibly accommodate additional modeling assumptions beyond the odds-ratio bounds, which was motivated by sensitivity analysis. We illustrate via an example of a total-variation bounded uncertainty set how to extend our framework to accommodate additional modeling assumptions. In the subsequent sections we show that this alternative uncertainty set enjoys similar minimax optimality and tractability guarantees as the approach above.
}

{
The pointwise interval odds-ratio uncertainty set, eq.~\eqref{uncertaintyset1eq}, might be pessimistic in ensuring robustness against every possible worst-case realization of unobserved confounding for each unit, which may be plausible under individual self-selection into treatment, whereas concerns about unobserved confounding might instead be limited to ``exceptions'', e.g. individuals with specific unobserved subgroup risk factors. For the Rosenbaum model in hypothesis testing, this has also been recognized by \citet{fogartyhasegawa17,hasegawa2017sensitivity} in the context of classic sensitivity analysis. 
}

Specifically, 
we construct the uncertainty set 
	\renewcommand*{\arraystretch}{1.4}
	$${\mathcal{W}}_n^{\Gamma,\Lambda} = \left\{ W \in \mathbb{R}^{\mathcal{I}_t} \colon
	\begin{array}{c}  \frac{1}{\vert \mathcal{I}_t \vert }\sum_{i \in \mathcal{I}_t }\fabs{ W_i - \tilde W_i} \leq {\Lambda_t}\;\forall t ,\\
	~ a_i^\Gamma  \leq W_i \leq   b_i^\Gamma\;\forall i
	\end{array}
	\right\}$$
with the population counterpart,
	\renewcommand*{\arraystretch}{1.4}
$${\mathcal{W}}^{\Gamma,\Lambda} = \left\{ W(t,x,y) \colon 
\begin{array}{c}
\E[ \fabs{ W(T,X,Y) - \tilde W(T,X)}\mid T=t ] \leq {\Lambda_t}\;\forall t,\\ a_t^\Gamma (x) \leq W(t,x,y) \leq   b_t^\Gamma (x)
\;\forall t\leq m-1, x \in \mathcal{X}, y\in \Rl\end{array}\right\}$$
When plugged into eq.~\eqref{pb-opt-pol}, this provides an alternative
policy choice criterion that is less conservative.
To make the choice of parameters easier,
we suggest to calibrate 
$\Lambda_t$ as a fraction, $\rho<1$, of the total deviation already allowed by $\mathcal W_n^\Gamma$. Specifically,
$\Lambda_t = \rho \frac{1}{\vert \indt \vert}
\sum_{i \in \indt } \max( \tilde W_i -a^\Gamma_i, b_i^\Gamma-\tilde W_i )$.\footnote{Enforcing the uncertainty budget separately within each treatment partition is crucial for computationally tractable policy learning and evaluation, as we discuss in Section \ref{sec-opt-pol}.}

\section{Analysis, improvement guarantees, and minimax optimality}\label{sec-analysis}
{Before discussing how we actually algorithmically compute $\overline \pi$, we next introduce finite-sample statistical guarantees on the performance of our approach. 
We first prove a finite-sample \textit{improvement} guarantee that provides that the policy we learn is assured to induce no harm, as long as the sensitivity model is well-specified.
We then prove a uniform convergence result simultaneously over \emph{both} the space of policies, $\Pi$, and the space of possible weights that agree with our sensitivity model, $\mathcal W^\Gamma$. As a consequence of this uniform convergence, we obtain a bound on the minimax regret that converges to the population optimum.}

In our analysis, in Sections 5.1 and 5.2, we assume the nominal propensities $\tilde{e}_t(x)$ are known so that the nominal inverse weights $\tilde W_i$ are known. In Section 5.3, we extend all of our results to the case of \emph{estimated} nominal propensities, where we instead plug in the estimate $\hat{e}_t(x)$ of $\tilde{e}_t(x)$. In particular, we analyze how our results change when we solve the optimization problem with some $\hat e_t(x)$ instead of $\tilde{e}_t(x)$, which provides a bound in terms of the estimation error, which generally vanishes as we collect more data. 

For both of these bounds we assume that both outcomes and true propensities are bounded.
\begin{assumption}[Bounded outcomes]
Outcomes are bounded, i.e. $\abs{Y} \leq  B$.\label{asn-bounded-outcomes}
\end{assumption}
\begin{assumption}[Overlap]\label{asn-overlap}
Strong overlap holds with respect to the true propensity: there exists $\nu> 0$ such that $e_t(x,y) \geq \nu~\forall t \in \{ 0, \dots, m-1 \} $
\end{assumption}

Moreover, both of these bounds depend on the flexibility of our policy class: it is critical that we search over a flexible but not completely unrestricted class in order to be assured improvement. 
We express the flexibility of $\Pi$ using the notion of the \emph{Vapnik-Chervonenkis (VC) major dimension}, which we define below \citep[see][p. 1309]{dudley1987universal}. 
\begin{definition}\label{vceq}
Given a ground set $\mathcal G$ and set of maps $\mathcal F\subseteq[\mathcal G\to\Rl]$,
the \textit{VC-major dimension} of $\mathcal{F}$ is the largest number $v\in\mathbb N$ such that there exists $g_1,\dots,g_v\in\mathcal G$ with
\begin{equation}\label{eq:vceq}
\braces{
	\prns{\indic{f(g_1)>\theta},\dots,\indic{f(g_v)>\theta}}
	:f\in\mathcal F,\theta\in\Rl
}=\fbraces{0,1}^v.
\end{equation}
\end{definition}
If eq.~\eqref{vceq} holds then we say that the superlevel sets of $\mathcal F$ shatter $x_1,\dots,x_v$, which means that {any} subset of the points belong \emph{exclusively} to some superlevel set of some $f\in\mathcal F$ and its complement to the corresponding sublevel set. The more complex a class is, the larger the point sets it can shatter. Thus, VC dimension is a natural expression of function class complexity or flexibility.

We will express the flexibility of $\Pi$ in terms of its VC-major dimension as a set of functions from $(t,x)\in\{0,\dots,m-1\}\times\mathcal X$ to $[0,1]$.
\begin{assumption}\label{assumption:vc1}
The policy class $\Pi$, as a class of functions $\{0,\dots,m-1\}\times\mathcal X\to[0,1]$, has a finite VC-major dimension no larger than $v$.
\end{assumption}
Assumption~\ref{assumption:vc1} holds for all multi-treatment policy classes we consider, including linear, logistic, and tree policies with bounded depth. Note that our treatment differs from multi-class classifiers as we treat $(t,x)$ as the ground set. It is nonetheless immediate to see that the VC-major dimension of both linear and logistic policies is at most $(m-1)(d+1)$. Moreover, for binary decision trees of depth no more than $D$, if each inner node 
can be a query $x_i\leq\theta$ for any $i=1,\dots,d$ and $\theta=\theta_{i1},\dots,\theta_{iK}$ and each leaf node is assigned its own 
probability vector in $\Delta^m$, then the VC dimension of this class is at most $2^D(m-1)\log_2(dK+2)$, as can been seen by following the arguments of \citet{golea1998generalization} and seeing this a direct sum of $2^D$ leaf functions, each consisting of $D-1$ conjunctions.

\subsection{Improvement guarantee}

We next prove that, if we appropriately bounded the potential hidden 
confounding, then the optimal value of our worst-case empirical regret objective  $\hat{\overline R}_{\pi_0}(\pi; \mathcal W_n)$ (defined in \Cref{rbareq-prod}) and achieved by $\overline \pi$, 
is asymptotically an upper bound on the true population regret of 
$\overline \pi$. The result is in fact a finite-sample result that gives precisely a bound on how much the latter might exceed the former due to finite-sample errors -- terms that vanish as $n$ grows, even if there is unobserved confounding.

Our guarantee relating the sample minimax regret (defined in \Cref{rbareq-prod}) is then as follows:
\begin{theorem}[Improvement bound]\label{improvethm}
Suppose Assumptions~\ref{asn-bounded-outcomes}, \ref{asn-overlap}, and \ref{assumption:vc1} hold. Suppose, moreover, that $W^*_{1:n}\in\mathcal W_n$.
Then, for a constant $K^\Pi$ which only depends on the VC-major dimension $v$ of $\Pi$, we have that with probability at least $1-\delta$:
\begin{equation}\label{complexitybound}
R_{\pi_0}(\hat{\overline{\pi}}(\Pi,{\mathcal{W}}_n,\pi_0))=V(\overline\pi)-V(\pi_0)\leq \hat{\overline R}_{\pi_0}(\pi; \mathcal W_n) +
\frac{1}{\nu}( {B}
K^\Pi
+ 3  ) 
\sqrt{\frac{ 2\log(\nicefrac{8m\vee 20 }{\delta}) }{{n}}}.
\end{equation}
\end{theorem}

Theorem~\ref{improvethm} says that the \emph{true} population regret of the policy we learn, when we implement it in fact, is bounded by the objective value that the policy minimizes, plus vanishing terms.
These vanishing terms, that is, the second term on the right hand side of  eq.~\eqref{complexitybound}, vanish 
at a rate of $O(n^{-1/2})$ and have sub-Gaussian tails, regardless of \emph{any} unobserved confounding.
Notice that, as long as $\pi_0\in\Pi$, which can be ensured by design, then we have that our objective is nonpositive, $\hat{\overline R}_{\pi_0}(\pi; \mathcal W_n)\leq 0$. Therefore, this means that we never do worse than $\pi_0$ (i.e., do harm), up to vanishing terms. Additionally, if our objective is sufficiently negative, which we can check by just evaluating it, then we are assured some strict improvement. 
Since we are able to guarantee this without being able to identify or estimate \emph{any} causal effect
due to the unobserved confounding,
Theorem~\ref{improvethm}
exactly captures the special appeal of our approach.

Our result above is generic for any uncertainty set $\mathcal W_n$; it only requires that it be well-specified. Note that for both of the uncertainty sets we propose in Section~\ref{sec-estimator}, the specification of the population sensitivity model ($W^*\in\mathcal W$) imply $W_{1:n}^*\in\mathcal W_n$, as the latter is simply the restriction of the former to the data. In the next section we further show that we can obtain the minimax-optimal regret in these sensitivity models. These results, however, \emph{will} depend on the uncertainty set and their complexity being manageable.

\subsection{Minimax optimality}

In the previous section we argued that our policy is assured (almost) no harm. A remaining question is whether it achieves the most improvement while doing no harm: whether or not, over all distributions that agree with our sensitivity model, it obtains the best possible uniform control on policy regret.
That is, since unconfoundedness does not hold, each policy may incur a range of possible regrets, depending on the true distribution of outcomes, which we cannot pin down even with infinite data. 
The \emph{best} safe policy uniformly minimizes all of these potential regrets simultaneously and is the minimax-optimal policy ${\overline{\pi}}^*(\Pi,{\mathcal{W}},\pi_0)$ defined in eq.~\ref{eq:minimaxoptpop}. We next show our policy is not only safe but also achieves the same uniform regret control asymptotically. In fact, we will give a finite-sample bound on our uniform regret control.

\paragraph{Controlling the complexity of the sensitivity model.}
Recall that our policy, $\hat{\overline{\pi}}(\Pi,{\mathcal{W}},\pi_0)$, is defined as the minimum over $\pi$ of the maximum over $W$ of $\hat R_{\pi_0}(\pi;W)$. Therefore, one approach may be to establish the uniform convergence of $\hat R_{\pi_0}(\pi;W)$ to $R_{\pi_0}(\pi;W)$ over all policies \emph{and} all weight functions in the sensitivity model. However, for the uncertainty sets we propose, this will fail. For example, the weight functions in $\mathcal W^\Gamma$ are far too many (isomorphic to all bounded functions) to expect such uniform convergence.

Instead, we first establish that we need only consider a special subclass of weight functions, which will in fact have bounded functional complexity. This highlights that in contrast to the sensitivity analysis setting in causal inference, where we consider all possible realizations which would impose no a priori structure on the set of weights, in the optimization setting where we are interested in the weights that are optimal for some policy $\pi$, the set of weights that achieve the sharp interval end-points (see \Cref{prop-sharpness}) has a lot of structure.
\begin{proposition}[Monotone weight solution for $\mathcal W^{\Gamma}$]\label{prop:monotoneweights}
		\renewcommand*{\arraystretch}{1.4}
Let \begin{align*}
\overline{\mathcal W}^\Gamma(\pi)
&=\braces{W(t,x,y)\colon 
\begin{array}{c}
W(t,x,y)=a_t^\Gamma(x)+u(y(\pi(t\mid x)-\pi_0(t\mid x))) \cdot (b_t^\Gamma(x)-a_t^\Gamma(x)),\\
u:\Rl\to[0,1] \text{ is monotonic nondecreasing}
\end{array}},\\
\overline{\mathcal W}_n^\Gamma(\pi)
&=\{(W(T_1,X_1,Y_1),\,\dots,\,W(T_n,X_n,Y_n):W\in \overline{\mathcal W}^\Gamma(\pi)\}.
\end{align*}
Then, for any $\pi:\mathcal X\to\Delta^m$,
\begin{align*}
{\overline R}_{\pi_0}(\pi;\mathcal W^\Gamma)
=\suma\sup_{W\in\overline{\mathcal W}^\Gamma(\pi)}{R}^{(t)}_{\pi_0}(\pi;W),\quad
\hat{\overline R}_{\pi_0}(\pi;\mathcal W_n^\Gamma)
=\suma\sup_{W\in\overline{\mathcal W}_n^\Gamma(\pi)}{\hat R}^{(t)}_{\pi_0}(\pi;W).
\end{align*}
\end{proposition}
This result is due to the special optimization characterization we present later in \Cref{thm-norm-wghts-soln}, which uses linear-fractional optimization to show that the solution takes a monotonic, thresholding form.
\begin{corollary}\label{cor:monotoneweights}
Let $\overline{\mathcal W}^\Gamma=\bigcup_{\pi\in\Pi}\overline{\mathcal W}^\Gamma(\pi),\,\overline{\mathcal W}_n^\Gamma=\bigcup_{\pi\in\Pi}\overline{\mathcal W}_n^\Gamma(\pi)$.
Then, for any $\pi\in\Pi$,
\begin{align*}
{\overline R}_{\pi_0}(\pi;\mathcal W^\Gamma)
=\suma\sup_{W\in\overline{\mathcal W}^\Gamma}{R}^{(t)}_{\pi_0}(\pi;W),\quad
\hat{\overline R}_{\pi_0}(\pi;\mathcal W_n^\Gamma)
=\suma\sup_{W\in\overline{\mathcal W}_n^\Gamma}{\hat R}^{(t)}_{\pi_0}(\pi;W).
\end{align*}
\end{corollary}
Corollary~\ref{cor:monotoneweights} shows that, when searching for policies in $\Pi$ to obtain uniform control on regret, it suffices to consider weight functions in $\overline{\mathcal W}^\Gamma$, which is a subset of ${\mathcal W}^\Gamma$. Again, this result crucially relies on the optimization structure of our problem.

Importantly, this subset, $\overline{\mathcal W}^\Gamma$, has much more structure and, in contrast to $\mathcal W^\Gamma$,
has bounded complexity.
\begin{proposition}\label{prop:vcmonotone}
Suppose Assumption~\ref{assumption:vc1} holds. Then $\overline{\mathcal W}^\Gamma$ has a finite VC-major dimension $v_w$.
\end{proposition}
\Cref{prop:vcmonotone} leverages the stability of VC-major classes (see \citealp[Lemma 2.6.19]{van1996weak} and \citealp[Proposition 4.2]{dudley1987universal}). Note that monotone functions are \emph{not} a VC class in the usual sense of having VC subgraphs, but they are VC-hull \citep[Example 3.6.14]{gine2016mathematical}.

Using Corollary~\ref{cor:monotoneweights} and Proposition~\ref{prop:vcmonotone}, we can obtain the following uniform convergence:
\begin{theorem}\label{unifconv}
Suppose Assumptions~\ref{asn-bounded-outcomes}, \ref{asn-overlap}, and \ref{assumption:vc1} hold. 
Then, for a constant $K^\Pi$ that depends only on the VC-major dimension $v$ of $\Pi$, we have that, with probability at least $1-\delta$: 
\begin{align}\notag
&\sup_{\pi\in\Pi}\abs{
\hat{\overline R}_{\pi_0}(\pi;\mathcal W_n^\Gamma)
-
\overline R_{\pi_0}(\pi;\mathcal W^\Gamma)
}
\leq 
	  36(12+\nu^{-1})  (B K^\Pi +\nu^{-1} (\Gamma  -  \Gamma^{-1} ) ({K^{  \Pi }} + B + m ) ) \sqrt{ \frac{\log(\nicefrac{15m}{p})}{ {n} }} 
\end{align}
\end{theorem}
Relative to \Cref{improvethm}, the additional dependence on $m$ arises due to the flexibility of $\overline{\mathcal W^\Gamma}$ where, per \Cref{prop:monotoneweights}, we may effectively choose a \emph{different} monotone function $u$ for \emph{each} treatment level $t=0,\dots,m-1$.

As a corollary to \Cref{unifconv} we obtain a finite-sample bound on our minimax suboptimality, which ensures asymptotic minimax optimality:
\begin{corollary}[Minimax regret bounds for $\mathcal W^\Gamma$] Suppose Assumptions~\ref{asn-bounded-outcomes}, \ref{asn-overlap}, and \ref{assumption:vc1} hold. 
Then, with probability at least $1-\delta$, we have that
\begin{align*}
&\overline R_{\pi_0}(\hat{\overline{\pi}}(\Pi,{\mathcal{W}}^{\Gamma}_n,\pi_0);\mathcal W^\Gamma)\\
&\leq \inf_{\pi\in\Pi}\overline R_{\pi_0}(\pi;\mathcal W^\Gamma)+ 
36(12+\nu^{-1})  (B K^\Pi +\nu^{-1} (\Gamma  -  \Gamma^{-1} ) ({K^{\Pi }} + B + m ) ) \sqrt{ \frac{\log(\nicefrac{15m}{\delta})}{ {n} }} 
\end{align*}
\end{corollary}
It is important to note that, in contrast to Theorem~\ref{improvethm}, this result depends crucially on the structure of $\mathcal W^\Gamma$. The key question is how flexible is the set of worst-case weight functions for any policy.

While our budgeted uncertainty set, $\mathcal W^{\Gamma,\Lambda}$, is also too flexible to expect uniform convergence over it, we can make similar arguments, focusing only on the set of worst-case weights: they satisfy a nondecreasing property similar to that of \Cref{prop:monotoneweights}, despite the additional constraint. 
	\renewcommand*{\arraystretch}{1.4}
\begin{proposition}[Monotone weight solution for $\mathcal W^{\Gamma,\Lambda}$]\label{prop:monotoneweightsLambda}
	Let \begin{align*}
	\overline{\mathcal W}^{\Gamma, \Lambda}(\pi;\mathbb P)
	&=\braces{W(t,x,y)\colon 
		\begin{array}{c}
		W(t,x,y)=a_t^\Gamma(x)+u(y(\pi(t\mid x)-\pi_0(t\mid x))) \cdot (b_t^\Gamma(x)-a_t^\Gamma(x)),\\
		\text{$u(y(\pi(t\mid x) - \pi_0(t\mid x ) )):\Rl\to[0,1]$ is monotonic nondecreasing}, \\
		\E_{\mathbb P}[ \fabs{ W(T,X,Y) - \tilde W(T,X)}\mid T=t ] \leq {\Lambda_t}\;\forall t
		\end{array}},\\
	\overline{\mathcal W}_n^{\Gamma,\Lambda}(\pi;{\mathbb P})
	&=\{(W(T_1,X_1,Y_1),\,\dots,\,W(T_n,X_n,Y_n):W\in \overline{\mathcal W}^{\Gamma,\Lambda}(\pi;\mathbb P)\}.
	\end{align*}
	Then, for any $\pi:\mathcal X\to\Delta^m$,
	\begin{align*}
	{\overline R}_{\pi_0}(\pi;\mathcal W^{\Gamma, \Lambda})
	=\suma\sup_{W\in\overline{\mathcal W}^{\Gamma, \Lambda}(\pi;\mathbb P)}{R}^{(t)}_{\pi_0}(\pi;W),\quad
	\hat{\overline R}_{\pi_0}(\pi;\mathcal W_n^{\Gamma, \Lambda})
	=\suma\sup_{W\in\overline{\mathcal W}_n^{\Gamma, \Lambda}(\pi;\mathbb P_n)}{\hat R}^{(t)}_{\pi_0}(\pi;W),
	\end{align*}
	where $\mathbb P$ denotes the population distribution of $T,X,Y$ and $\mathbb P_n$ the corresponding empirical distribution.
\end{proposition}
These arguments require proving structural properties of the optimal solution under this budgeted uncertainty set, which allow us to use the same stability arguments for various compositions of VC-major classes. We remark that the structural results for the budgeted uncertainty set are weaker than that of the unbudgeted one (\Cref{thm-norm-wghts-soln}), where we also obtain efficient algorithms. Since these are somewhat more involved, we only quote the final regret bound, and refer the reader to the supplement for details.
\begin{proposition}[Minimax regret bounds for $\mathcal W^{\Gamma,\Lambda}$]\label{cor-budgeted-minimax}
	Suppose Assumptions~\ref{asn-bounded-outcomes}, \ref{asn-overlap}, and \ref{assumption:vc1} hold. Then, with probability at least $1-\delta$, we have that
\begin{align*}
&\overline R_{\pi_0}(\hat{\overline{\pi}}(\Pi,{\mathcal{W}}^{\Gamma,\Lambda}_n,\pi_0);\mathcal W^{\Gamma,\Lambda})\\
&\leq \inf_{\pi\in\Pi}\overline R_{\pi_0}(\pi;\mathcal W^{\Gamma,\Lambda})+
36(12+\nu^{-1})  (B K^\Pi +\nu^{-1} (\Gamma  -  \Gamma^{-1} ) ((m \frac{2 B \Gamma\nu^{-1} }{ \min_t {\Lambda}_t \wedge 1 } +1){K^{  \Pi }}  + B + m ) ) \sqrt{ \frac{\log(\nicefrac{30m}{\delta})}{ {n} }} 
\\
& 
\phantom{\leq}+ \frac{2 mB \Gamma\nu^{-1} \max_t  {\Lambda_t}}{ \min_t {\Lambda}_t \wedge 1 }\frac{\max_t \nicefrac{1}{p_t}}{n}  
	\end{align*} 
\end{proposition}

\subsection{Estimated propensities}\label{sec:estprop}
All of the above results are presented for the case of known nominal propensities, $\tilde e_t(x)$, that is, when $\mathcal W^\Gamma$, which is centered at the nominal inverse propensity weights $\tilde W_i$, is known. 
If, as is the case for an observational study, the nominal propensities need to be estimated from data, we may not have $\mathcal W^\Gamma$ and may not compute $\hat{\overline{\pi}}(\Pi,{\mathcal{W}}_n,\pi_0)$. We next show that in fact if we use estimated nominal propensities, $\hat e_t(x)$, instead of the true nominal propensities then the resulting error is ultimately additive and controlled by the accuracy of propensity estimation, which vanishes as we collect more data.
	\begin{proposition}[Bounded perturbations]\label{lemma-lipschitzness}
Let $\hat W_i=1/\hat e_{T_i}(X_i)$ and
\begin{align*}
\hat{\mathcal{W}}_n^{\Gamma} &= \left\{ W \in \R n \colon  \hat a_{i}^\Gamma  \leq W_i \leq   \hat b_{i}^\Gamma,\;\forall i=1,\dots,n\right\}
,~
\text{where}~
\hat a_{i}^\Gamma=
1+\Gamma^{-1}(\hat W_i - 1)
,~
\hat b_{i}^\Gamma=
1+\Gamma(\hat W_i - 1).
\end{align*}
Then, under Assumption \ref{asn-bounded-outcomes}, for any $\pi:\mathcal X\to\Delta^m$,
		\begin{equation}\label{eq:boundedperturb} \abs{ \hat{\overline{R}}_{\pi_0}(\pi, \hat{\mathcal{W}}^\Gamma_n)  -\hat{\overline{R}}_{\pi_0}(\pi, {\mathcal{W}}^\Gamma_n) }	\leq 2B (\Gamma + \Gamma^{-1}) 
	\frac1n\sum_{i=1}^n
	\abs{ 	\frac{1}{\hat{e}_{T_i}(X_i)}- \frac{1}{\tilde{e}_{T_i}(X_i)} }
	\end{equation}
	\end{proposition}

Proposition \ref{lemma-lipschitzness} is a consequence of the unique optimization structure of the worst-case regret over weights in $\mathcal W_n^\Gamma$. The proof leverages a partial Lagrangian dual of the optimization problem and studies sensitivity to plugged-in nominal propensities in the dual. Note that by additionally assuming strong overlap in the nominal propensities, we can bound errors in the inverse propensities in terms of errors in the propensity function itself, which we can in turn bound using standard finite-sample guarantees for learning conditional expectations \citep{bartlett2005local}.
Note that the bound in \cref{eq:boundedperturb} would scale as these bounds.
It remains an important direction for future research to obtain bounds of the form of \cref{eq:boundedperturb} that have a multiplicative-bias property, allowing for slower-than-$n^{-1/2}$ estimation of propensities without deteriorating the overall $n^{-1/2}$ rate, as in \citet{wager17}.

Since \Cref{lemma-lipschitzness} holds deterministically and for all policies, including the sample-optimal policy, it immediately shows that the policy we get by optimizing our worst-case empirical regret with estimated nominal propensities, $\hat{\overline{\pi}}(\Pi,\hat{\mathcal{W}}_n,\pi_0)$, is actually near-optimal in objective relative to the worst-case empirical regret we would obtain with true nominal propensities, that is, $\hat{\overline{R}}_{\pi_0}(\pi, {\mathcal{W}}^\Gamma)$. Therefore, all previous results for our method similarly hold for $\hat{\overline{\pi}}(\Pi,\hat{\mathcal{W}}_n,\pi_0)$ with the addition of two times the right-hand side of \cref{eq:boundedperturb} to any previous bound. In particular, for the improvement guarantee for the case of $\mathcal W^\Gamma$, we need only ensure that $W^*\in\mathcal W^\Gamma$, which is implied by the validity of the marginal sensitivity model; we do \emph{not} need to ensure that $W^*_n\in\hat {\mathcal W}^\Gamma_n$, which may be a random event depending on our estimation.
}

\section{Algorithms for Optimizing Robust Policies}\label{sec-opt-pol} 

We next discuss how to algorithmically solve the policy optimization problem in eq.~\eqref{pb-opt-pol} and actually find the confounding-robust policy, $\hat{\overline \pi}$.
{ In the main text, we focus on differentiable parametrized policy classes, $\mathcal{F}=\{ \pi_\theta(\;\cdot \; ): \theta \in \Theta \}$ such that $\pi_\theta(t\mid x)$ is differentiable with respect to $\theta$, 
such as logistic policies. We will use a subgradient method to find the robust policy. In the appendix, we also discuss optimization over decision-tree based policies, using a mixed-integer optimization formulation. In both cases, our solution will depend on a characterization of the inner worst-case regret subproblem. }

We first discuss how to solve the worst-case regret subproblem \textit{for a fixed }policy, which we will then use to develop our algorithms.

\subsection{Dual Formulation of Worst-Case Regret}

The minimization in eq.~\eqref{pb-opt-pol}  
involves an inner supremum, 
namely {$\hat{\overline{R}}_{\pi_0}(\pi;\mathcal{W}_n^\Gamma)$. }
Moreover, this supremum
over weights $W$ does not on the face of it appear to be a convex
problem. However, a standard transformation will reveal its convexity.
We next proceed to characterize this supremum, formulate it as
a linear program, and, by dualizing it, provide an 
efficient line-search procedure for finding the pessimal weights. 

For compactness and generality, we address the  
optimization problem 
{ $\hat{\overline Q}_t(r;\mathcal{\weightuncertainty})$ }
parameterized by 
an arbitrary reward vector $r \in \mathbb{R}^n$, where
{
\begin{equation}\label{opt-FLPP} 
\hat{\overline Q}_t(r;\mathcal{\weightuncertainty})=\max_{W\in\mathcal{\weightuncertainty}}
\frac{\sum_{i=1}^n r_i W(T_i, X_i, Y_i) \mathbb{I}[T_i=t] }{  {\sum_{i=1}^n W(T_i, X_i, Y_i) \mathbb{I}[T_i=t]} }.
\end{equation}
}
{ 
	To recover $\hat{\overline{R}}_{\pi_0}(\pi;\mathcal{\weightuncertainty}_n^\Gamma)$, we would simply compute, with $r_i =( \pi(T_i \mid X_i) - \pi_0(T_i \mid X_i) )Y_i $, $$\hat{\overline{R}}_{\pi_0}(\pi;\mathcal{\weightuncertainty}_n^\Gamma) = \suma  \hat{\overline Q}_t (r\; ;\mathcal{\weightuncertainty}^{\Gamma}_n) .$$
}
{
For the remainder of this subsection, we discuss solving the program generically for the $r$-weighted linear fractional objective $Q(r;\mathcal{\weightuncertainty})$, without discussion of multiple treatment partitions. In doing so, we reindex $n$. First we consider $\mathcal{\weightuncertainty}_n^\Gamma$.
Since $\mathcal{\weightuncertainty}_n^\Gamma$ involves only linear constraints on $W$,
eq.~\eqref{opt-FLPP} for $\mathcal{\weightuncertainty}=\mathcal{\weightuncertainty}_n^\Gamma$ is a \emph{linear fractional program}. We can reformulate it as a linear program by applying the Charnes-Cooper transformation \citep{cc62}, requiring weights to sum to 1, and rescaling the pointwise bounds by a nonnegative scale factor $\ccpscalar$. 
We obtain the following equivalent linear program in a scaling factor and normalized weight variables, $\ccpscalar = \frac{1}{ \sum_i^n W_i}; w = {W}{\ccpscalar}$:
\begin{equation}\label{opt-LPP}  
\begin{aligned}
\hat{ \overline Q}(r;\mathcal{\weightuncertainty}_n^\Gamma) 
= {\max}_{\ccpscalar\geq0,w\geq 0}  ~&~
\sum_{i=1}^n r_i w_i   \\
\text{s.t.}~&~ \sum_{i=1}^n w_i = 1\\
&~\ccpscalar a_i^\Gamma \leq w_i \leq  \ccpscalar b_i^\Gamma~~\forall\; i = 1,\dots,n 
\end{aligned}
\end{equation}
}
The dual problem to eq.~\eqref{opt-LPP} has dual variables $\lambda \in \mathbb{R}$ for the weight normalization constraint and $u, v \in \mathbb{R}^n_+$ for the lower bound and upper bound constraints on weights, respectively. By linear programming duality, we then have that
{\begin{equation}\label{pbm-dual} 
\begin{aligned}
\hat{ \overline Q}(r;\mathcal{\weightuncertainty}_n^\Gamma)={\min}_{u\geq0, v\geq0, \lambda} ~&~   \lambda  
\\
\text{s.t.}~&~ 
-v^\intercal b^\Gamma + u^\intercal a^\Gamma \geq 0\\ 
&~v_i -u_i + \lambda  \geq r_i~~\forall\; i = 1,\dots,n 
\end{aligned}
\end{equation}}
We use this to show that solving the inner subproblem requires only sorting the data and a ternary search to optimize a unimodal function. This generalizes the result of \citet{aronowlee12} for arbitrary pointwise bounds on the weights. Crucially, the algorithmically efficient solution will allow for faster subproblem solutions when optimizing our regret bound over policies in a given policy class. 
\begin{theorem}[Normalized optimization solution]\label{thm-norm-wghts-soln}

Let $(i)$ denote the ordering such that $r_{(1)} \leq r_{(2)} \leq \cdots \leq r_{(n)}$. 
Then, {$\hat{\overline Q}(r;\mathcal{\weightuncertainty}_n^\Gamma)=\lambda(k^*)$}, where 
$k^* = \inf \{k=1,\dots,n+1: \lambda({k}) < \lambda({k-1}) \}$ and
\begin{equation} 	\lambda(k) = \frac{   {\sum_{i < k }}a^\Gamma_{(i)}r_{(i)} + {\sum_{i \geq k}} b^\Gamma_{(i)} r_{(i)}}{ {\sum_{i < k}} a^\Gamma_{(i)} + {\sum_{i \geq k}} b^\Gamma_{(i) }}. \end{equation}
{\blockedit
Specifically, we have that $\hat{\overline{Q}}(r;\mathcal W_n^\Gamma)=\frac{\sum_{i=1}^nr_iW^\dagger_i}{\sum_{i=1}^nW^\dagger_i}$ where $W^\dagger_{(i)}=a_{(i)}^\Gamma$ if $i<k^*$ and $W^\dagger_{(i)}=b_{(i)}^\Gamma$ if $i\geq k^*$.}

Moreover, $ \lambda({k}) $ is a discrete concave unimodal function.
\end{theorem}

Next we consider a budgeted uncertainty set, {$\hat{\overline{Q}}(r;\mathcal{\weightuncertainty}_n^{\Gamma,\Lambda})$}.
Write an extended formulation for $\mathcal{\weightuncertainty}_n^{\Gamma,\Lambda}$ using only linear constraints:
{ 
$$
\mathcal{\weightuncertainty}_n^{\Gamma,\Lambda}=\left\{ W \in \mathbb{R}^n_+ \colon  \exists d~\text{s.t.}~\sum_{i=1}^nd_i \leq \Lambda,~d_i\geq W_i - \tilde W_i,~d_i\geq \tilde W_i-W_i,~ a_i^\Gamma  \leq W_i \leq   b_i^\Gamma~\forall i  \right\}
$$
}
This immediately shows that {$\hat{\overline{Q}}(r;\mathcal{\weightuncertainty}_n^{\Gamma,\Lambda})$} remains a fractional linear program.
Indeed, 
a similar Charnes-Cooper transformation as used above 
yields a 
non-fractional linear programming formulation:
{
\begin{equation*}\begin{aligned}\label{pbm-dual-budgeted} 
\hat{\overline{Q}}(r;\mathcal{\weightuncertainty}_n^{\Gamma,\Lambda})=
{\max}_{\ccpscalar > 0,w\geq0,d}~&~  \sum_{i=1}^n w_i r_i
\\\text{s.t.}~&~ \sum_{i=1}^n d_i \leq \Lambda \ccpscalar,~~\sum_{i=1}^n w_i = 1
\\&~{a^\Gamma_i \ccpscalar} \leq w_i \leq {b^\Gamma_i } \ccpscalar~~\forall\;i=1,\dots,n
\\&~d_i \geq w_i - \tilde W_i \ccpscalar~~\forall\;i=1,\dots,n
\\&~d_i \geq \tilde W_i \ccpscalar - w_i~~\forall\;i=1,\dots,n
\end{aligned}\end{equation*}
}
The corresponding dual problem is: 
\begin{equation*}\begin{aligned}
\hat{\overline{Q}}(r;\mathcal{\weightuncertainty}_n^{\Gamma,\Lambda})=
{\min}_{g\geq 0,h\geq 0,u\geq 0,v\geq 0, \nu \geq 0, \lambda } %
~&~\lambda
\\\text{ s.t. } ~&~
v_i - u_i + g_i- h_i + \lambda \geq r_i~~\forall\;i=1,\dots,n
\\&~v_i \geq g_i + h_i ~~\forall\;i=1,\dots,n
\\&~-b^\intercal v + a^\intercal u - \Lambda \nu + g^\intercal \tilde W+ h^\intercal \tilde W  \geq 0   
\end{aligned}\end{equation*}
As { $\hat{\overline{Q}}(r;\mathcal{\weightuncertainty}_n^{\Gamma,\Lambda})$} remains a linear program, we can easily solve it using off-the-shelf solvers, even if it does not admit as simple of a solution as {$\hat{\overline{Q}}(r;\mathcal{\weightuncertainty}_n^{\Gamma})$} does. 
\subsection{Optimizing Parametric and Differentiable Policies}\label{sec-subgrad}

In the main text, we consider iterative optimization to optimize over a \textit{parametrized} policy class $\Pi=\fbraces{\pi_\theta(\cdot, \cdot):\theta\in\Theta}$, where the parameter space $\Theta$ is convex (usually $\Theta=\R m$), and {$\pi_\theta(t\mid x)$} is differentiable with respect to the parameter $\theta$. In the appendix, we discuss global optimization approaches for policy learning, for example over the interpretable policy class of optimal trees.
We suppose that 
{$\nabla_\theta \pi_\theta(t\mid x)$} is given as 
an evaluation oracle.
An example is logistic policies for binary treatments where it is sufficient to only parametrize for assigning $T=1$,
{
$\pi_{\alpha,\beta}(1\mid X) = \sigma(\alpha+\beta^\intercal X)$ and}
$\Theta=\R{d+1}$. Since $\sigma'(z)=\sigma(z)(1-\sigma(z))$, evaluating derivatives is simple.

Our gradient-based procedure leverages that we can solve the inner subproblem to full optimality in the sample. 
Note that {$\hat{\overline Q}(r;\mathcal{W})$} is convex in $r$ since it is a maximum over linear functions in $r$. Correspondingly, its subdifferential at $r$ is given by 
the argmax set, 
{where $\sum_{i \in \mathcal{I}_T} W$ denotes the vector of normalizing weights corresponding to the observed treatment pattern $T$:
$$
\partial_r\hat{\overline Q}(r;\mathcal W)=
\braces{
	\frac{W}{  \sum_{i \in \mathcal{I}_T} W} 
:
W\in\mathcal W~,
\sum_{i=1}^n \frac{W_i}{\sum_{i\in {\mathcal{I}}_{T_i}^n} W_i} r_i \geq\hat{\overline Q}(r;\mathcal W).
}
$$
	If we set $r_i(\theta) =( \pi_\theta(T_i\mid X_i) - \pi_0(T_i \mid X_i) ) Y_i$,
so that 
$\hat{\overline Q}(r(\theta);\mathcal W)=\hat{\overline R}_{\pi_0}(\pi_{\theta}(\cdot);\mathcal W)$, then
$\frac{\partial r_i(\theta)}{\partial \theta_j}=Y_i \frac{\partial{\pi_\theta(T_i\mid X_i)}}{\partial \theta_{j}}$.
Although $F(\theta):=\hat{\overline R}_{\pi_0}(\pi_\theta;\mathcal W)$
may not be convex in $\theta$, this suggests a subgradient descent approach. 
Let 
}
{
\begin{align*}g(\theta;W)&=
\nabla_\theta
\sum_{i=1}^n \frac{W_i}{\sum_{j\in \mathcal{I}_{T_i} }^n W_j} (\pi_\theta(T_i\mid X_i) - \pi_0(T_i \mid X_i) ) Y_i =\sum_{i=1}^n \frac{W_i}{\sum_{j\in \mathcal{I}_{T_i}}^n W_j}Y_i
\nabla_\theta \pi_\theta(T_i\mid X_i)
\end{align*}
}
Note that whenever 
{$\partial_r\hat{\overline Q}(r(\theta);\mathcal W)=\fbraces{\frac{W}{  \sum_{i \in \mathcal{I}_T} W} }$}
is a singleton then $g(\theta;W)$ is in fact a gradient of $F(\theta)$.

{
At each step, our algorithm starts with a current value of $\theta$, then proceeds by finding the weights $W$ that realize $
\hat{\overline R}_{\pi_0}(\pi(\cdot\; ;\theta)$ by using an efficient method as in the previous section, and then takes a step in the direction of $-g(\theta;W)$. 
Using this method, we can optimize policies, over both the unbudgeted uncertainty set $\mathcal{W}^\Gamma_n$ and the budgeted uncertainty set $\mathcal{W}^{\Gamma, \Lambda}_n$. 
We return the averaged $\theta$ parameter for each initialization; and we ultimately average the parameter achieving the best over multiple restarts. 
Our method is summarized in Alg.~\ref{subgradalgo}. 
}
\begin{algorithm}[t!]
\caption{Subgradient Method}\label{subgradalgo}
\begin{algorithmic}[1]
\State Input: step size $\eta_0$, step-schedule exponent $\kappa\in(0,1]$, initial iterate $\theta_0$,  number of iterations $N$
\For{$k=0,\dots,N-1$}: 
\State $\eta_k \gets \eta_0 t^{-\kappa}$ \Comment{Update step size}
\State {$\ell_k, W \gets \underset{W \in \mathcal{W}}{\max},~\in\underset{W \in \mathcal{W}}{\arg\max}~{\sum_{i=1}^n \frac{W_i}{\sum_{i \in \mathcal{I}_{T_i}} W_i}(\pi_{\theta_k}(T_i \mid X_i)- \pi_0(T_i\mid X_i))  Y_i }$} \Comment{Solve inner subproblem for $\theta_t$}
\State $\theta_{k+1} \gets \op{Projection}_\Theta(\theta_k - \eta_k \cdot g(\theta_k;W))$ \Comment{Move in subgradient direction}
\EndFor
\State
\Return { $\frac 1n\sum_{t = 1}^{N}\theta_{t}$ }
\end{algorithmic}
\end{algorithm}
{ 
In \Cref{apx-opt-refinements} of the Appendix, we include further algorithmic refinements to this subgradient procedure that leverage the special nested structure of the uncertainty sets. We find that these refinements help empirically in stabilizing the optimizing when we compute confounding-robust policies for multiple values of $\Gamma$, as we anticipate a decision-maker would, over reasonable plausible ranges of $\Gamma$. 
}

{\blockedit
\subsection{Optimizing Over Other Policy Classes}
We next discuss how our approach can be extended to other, more general policy classes, if we have a representation of the constraint $\pi\in\Pi$ that is compatible with conic or integer programming solvers.
\begin{proposition}\label{thm:reformgenpolicy}
Suppose $\mathcal W_n=\{W_{1:n}:(W_i)_{i:T_i=t}\in\mathcal W_{n,t}\;\forall t=0,\dots,m-1\}$ takes a product form and that $\mathcal W_{n,t}$ is convex with a non-empty relative interior.
Let $\Pi_n=\{(\pi(T_1\mid X_1),\dots,\pi(T_n\mid X_n)):\pi\in\Pi\}$.
Then,
$$
\min_{\pi\in\Pi}\hat{\overline{R}}_{\pi_0}(\pi;\mathcal W_n)=
\inf\braces{\suma\lambda_t\colon
p\in\Pi_n,\,
(\lambda-Y_i(p_i-\pi_0(T_i\mid X_i)))_{i:T_i=t}\in\mathcal W_{n,t}^*\;\forall t,
}
$$
where $S^*=\{p:u^Tp\leq0\;\forall u\in S\}$ denotes the dual cone of a set $S$.
\end{proposition}
Aside from the constraint $p\in\Pi_n$,
\Cref{thm:reformgenpolicy} provides a convex conic formulation of our optimization problem. If $\mathcal W_n$ is polyhedral as for our two proposed uncertainty sets then this formulation is in fact linear. What remains is to formulate the constraint $p\in\Pi_n$. For the case of sparse linear policies \citep{ur15} and fixed-depth decision trees \citep{bd17,k16}, existing such formulations based on integer optimization exist and can be used to adapt our approach to such policy classes. In the appendix, we provide a more detailed treatment for the case of decision trees.
}

\section{Empirical Results} 

In this section we present empirical results on two experiments to investigate the benefit of confounding robustness. Our first experiment is a simple synthetic example that we use to illustrate the different methods in a controlled setting. {Our second experiment develops a case-study, drawing on the data from the parallel WHI observational study and clinical trial. There, harm would be done by unwarranted aggressive intervention by personalized policy learning led astray by confounding. Our more careful approach to learning to personalized policies is able to avoid such harm, and still offer sizable improvements over baseline by personalizing care, for a variety of possible reward scalarizations of reductions in high blood pressure against known clinical benefits.}

\begin{figure}[t!]
    \includegraphics[width=\textwidth]{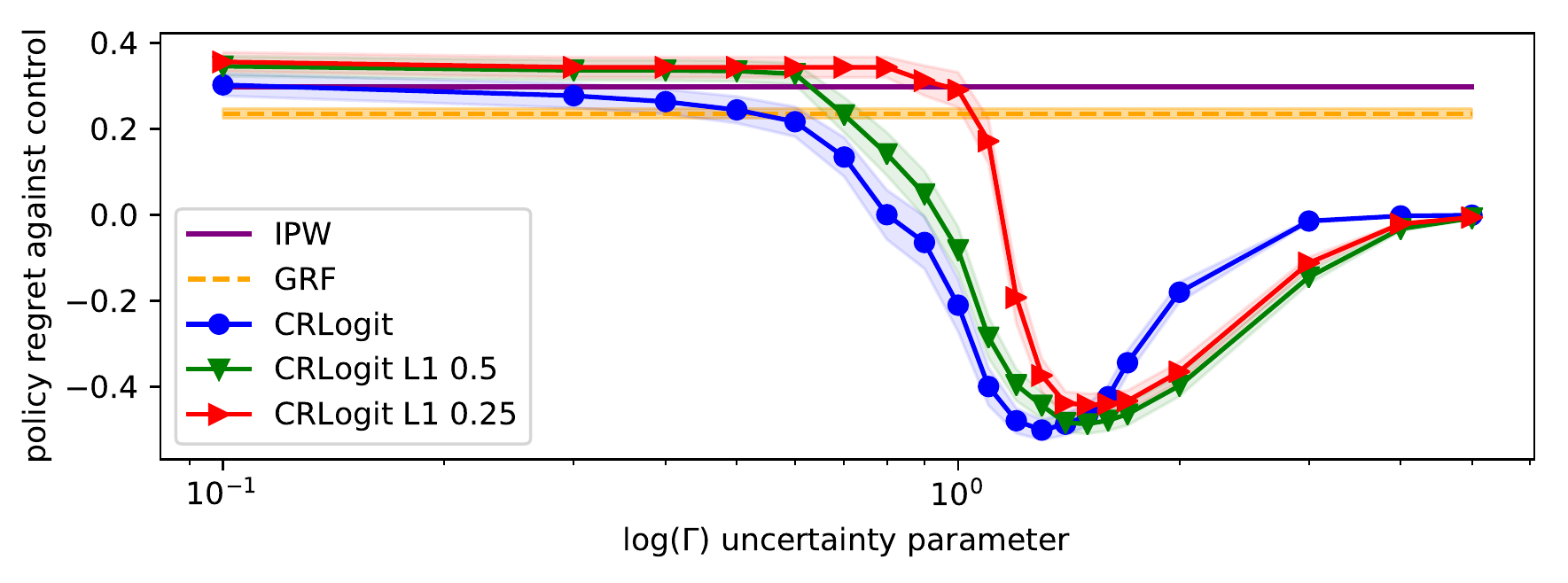}
    \caption{Out of sample policy regret on simulated data in Sec.~\ref{exp-sim}}
    \label{fig-syn}          
\end{figure}
\subsection{Simulated data}\label{exp-sim} 
\subsubsection{Binary Treatments} 
We first consider a simple linear model specification demonstrating the possible effects of significant confounding on inverse-propensity weighted estimators. We generate our data as follows, from a true propensity model based on an unobserved confounder $U$ which is a function of the potential outcomes:
{ 
\[ \xi \sim \Bern(\nicefrac{1}{2}), \quad X \sim N((2T-1)\mu_x, I_5 ), \quad g(Y(0), Y(1))= \mathbb{I}[Y_i(1) < Y_i(0) ] \]
\[  Y(t) = \tilde{\beta}^\intercal x + \mathbb{I}[T=1] \beta_{treat}^\intercal x  + \alpha  \mathbb{I}[T=1]  + \tilde{\eta} + \tilde{\eta}\xi + \epsilon \]
}
The constant treatment effect is $\alpha = 2.5$ with the linear interaction $\beta_{treat} = [-1.5,1,-1.5,1,0.5]$. The covariate mean is $\mu_x =[-1,.5,-1,0,-1]$. The noise term $\xi$ affects outcomes with coefficients $\eta = -2, \omega = 1$, in addition to a uniform noise term $\epsilon\sim N(0,1)$ which is the same for both treatments. 
We let the nominal propensities be logistic in $X$, $\tilde e(X_i)=\sigma(\beta^\intercal x)$ with $\beta =[0, .75, -.5,0,-1,0]$, and we generate $T_i$ according to the true propensities, which we set to
{
$$
e(X_i,g(Y(0), Y(1)))=\frac{4+5g+\tilde e(X_i)(2-5g)}{6\tilde e(X_i)}.
$$
In particular, this makes $e(X_i,g_i(Y_i(0), Y_i(1)))$ realize the upper bound
in eq.~\eqref{oddsratio} for $\Gamma=1.5$ when $g_i(Y_i(0), Y_i(1))=1$ and the lower
bound otherwise.}

We compare the policies learned by a variety of methods.
We consider two commonplace standard methods 
that assume unconfoundedness:
the logistic policy minimizing the IPW estimate with nominal propensities
and the direct comparison policy gotten by estimating CATE using
causal forests and comparing it to zero \citep[GRF;][]{wager2017estimation}. 
We compare these to two variants of our methods using the never-treat
baseline policy, {$\pi_0(0\mid x)=1$}:
our robust logistic policy using the unbudgeted uncertainty set and
our robust logistic policy using the budgeted uncertainty set
{ and multipliers $\rho = 0.5, 0.25$.}
For each of these we vary the parameter $\Gamma$ in
{$\fbraces{0.1,0.2,0.3,\dots,1.8,1.9,2,3,4,5}$.}
For logistic policies, we run 15 random restarts of Alg.~\ref{subgradalgo} with a step-schedule of $\kappa=0.5$ and return the one with the best robust objective value, unless the best robust objective value is positive, in which case we just return $\pi_0$, which is feasible.

For each of $50$ replications, we generate an observational dataset of $n=200$ according to the above model, run each of the above mentioned methods to learn a policy, and compute the \emph{true} value of each learned policy (by using the known counterfactuals, which we generated). We report the value as the regret relative to the value of $\pi_0$. We plot the results in Fig.~\ref{fig-syn}, showing the mean regret over replications along with the standard error (shaded regions).
{We highlight that the worse performance of IPW and GRF does not imply an issue with the algorithms themselves but rather with relying on the assumption of unconfoundedness when it in fact fails to hold.}
{
\subsubsection{Multiple Treatments} 
For comparison, we include an example with multiple (three) treatments. We parametrize the policy class with a multinomial logistic probability model (a direct extension of the binary treatment case), e.g. $\pi(t\mid X)  = \frac{\exp(\beta_t^\top X )}{\suma \exp(\beta_t^\top X )} $. Our simulation setup is similar to the case for binary treatments. We define the outcome models. In the simulation setup, one treatment arm is highly variable with heterogeneous effects, with also greater confounding (variability in underlying propensity scores). The unobserved confounding affects the confounding generation for the variable treatment, $T=1$, while now $X$ is generated uniformly on $[-3,3]$ for all covariates, to reduce variability.
\[ \xi \sim \Bern(\nicefrac{1}{2}), \quad X \sim \text{Unif}(-3, 3)^5, \quad U= \mathbb{I}[Y_i(1) < Y_i(0) ] \]
\[  Y(t) = \tilde{\beta}^\intercal x + \tilde{\eta} \xi + \epsilon +\sum_{t'=1}^{m-1}  \mathbb{I}[t=t']  (\beta_{t'}^\intercal x  + \alpha_{t'} + \eta_{t'} \xi  ) \]
We parametrize the simulation by vectors of confounding effect and average treatment effect, $\eta = (0,-2,0), \alpha = (0,2,0.5)$, linear effects $ \tilde{\beta} = (0, .5, -0.5, 0, 0), \beta_0 = \vec{0}, \beta_1 = 0.75(-1.,0.5,-1.,1.,0.5), \beta_2 = \vec{0}$, and confounding effects $\beta_{0,\text{treat}}  = \vec{0}, \beta_{1,\text{treat}} = (0,1.5, -1, 0, -2), \beta_{2,\text{treat}} = (0, 0, 0.5,0,0.5)$.
\begin{figure}[t!]
	\includegraphics[width=\textwidth]{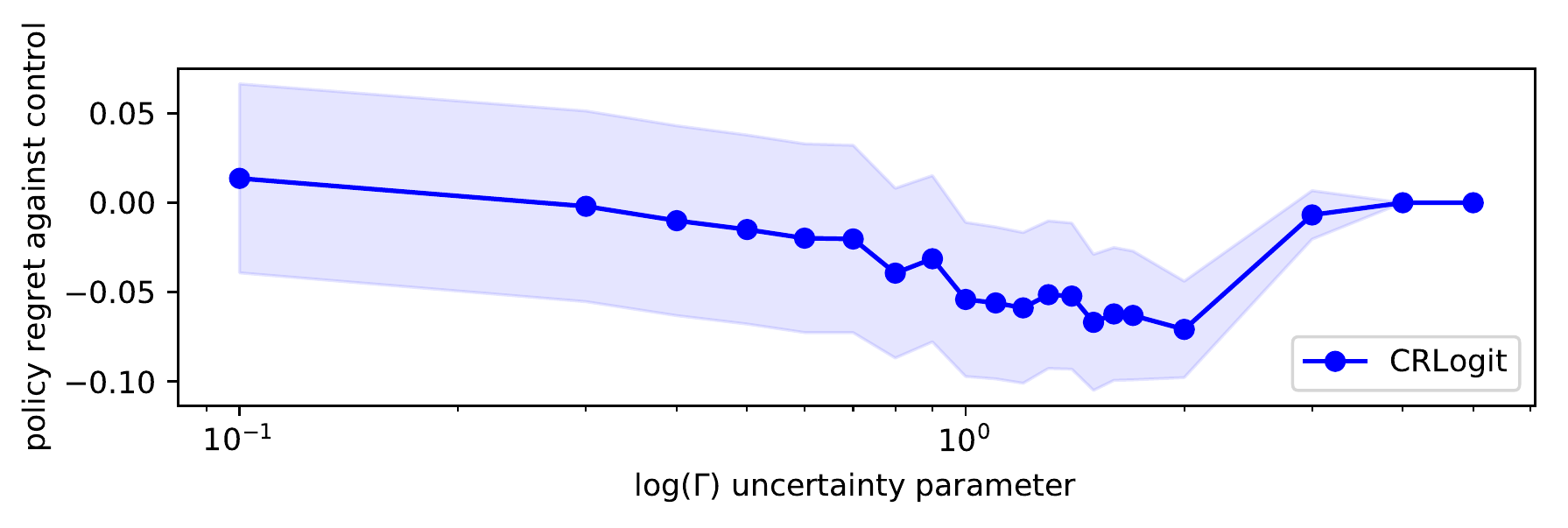}
	\caption{Out of sample policy regret on simulated data, three treatments in Sec.~\ref{exp-sim}}
	\label{fig-syn-mt}          
\end{figure}
We include the results in \Cref{fig-syn-mt}. We generate 50 replications from this data generating process with $n=200$, and evaluate on a large generated test set with known counterfactuals. However, the additional parametrization (scaling with the number of treatments) leads to a noisier optimization process by the method of Alg.~\ref{subgradalgo}; we leave further refining the optimization for future investigation.
}
}
{ 
\subsection{Assessment with Clinical Data: Women's Health Initiative Trial}\label{sec-whi}
	We next develop a case study on the parallel Women's Health Initiative (WHI) clinical trial and observational study, using semi-synthetic outcomes which scalarize actual clinical outcomes with a treatment effect ``bonus'' reflecting known ancillary benefits. The historical case of HRT is a clear example of an intervention policy
	learned from observational data that led to unwarranted harm due to
	unaccounted-for confounding. We now revisit the real data, under a hypothetical scenario where treatment provides some benefit (in both the observational study and clinical trial), to illustrate that under a variety of possible scalarizations, our approach provides guaranteed benefit above a known baseline whereas naive policies introduce harm. 
	Since we vary over a range of possible scalarizations, our focus here is not on drawing specific clinical prescriptions, but rather illustrating the behavior of the method in a variety of treatment effect profile environments, and illustrating that for a variety of parameters, our approach will lead to \textit{some} degree of improvement while confounded methods would introduce harm.
\subsubsection{WHI Case Study Background} 
\paragraph{Unobserved confounding and WHI.} 
The range of possible unobserved confounders which were posited to reconcile the different findings from the clinical trial and observational study illustrate broader confounding patterns that might plague observational studies in general.
The observational study may have been confounded by plausible, well-recognized confounding phenomena, \textit{healthy user bias} and \textit{confounding by indication}, which challenge the validity of all epidemiological research from observational health databases and which may induce correlation in either direction between treatment selection and outcomes \citep{bsgrs10}. Such possible confounding factors arise naturally from the structure of healthcare data in which physicians determined treatment assignment to manage health outcomes in the first place. Much discussion attributes confounding in the study to \textit{healthy user bias} arising from differing lifestyle factors in the population of women self-selecting into HRT, where lifestyle factors such as general health-seeking behaviors, such as exercise or maintaining heart-healthy diets, are correlated with better expected outcomes related to CHD on average. These general health-seeking and health-promoting behaviors tend to reduce artherosclerotic risk and risk of CHDs, but are unobserved lifestyle factors that are confounders for self-enrollment into HRT. Conversely, the study may have also been \textit{confounded by indication} or severity, where the presence of clinical activities such as prescription of HRT is correlated with, or indicates, greater initial symptom severity, which may lead to attenuation in the perceived reduction in vasomotor symptoms. 
\paragraph{Policy learning and Hormone Replacement Therapy (HRT).} 
We motivate our policy learning setting noting that modern clinical guidelines, included in \citet{bakour2015latest}, recognize that ``when HRT is individually tailored, women gain maximum advantages and the risks are minimised.'' For all women, the improvement of vasomotor symptoms was significant, but ultimately the greater risks of adverse events outweigh the clinical benefits for older women. Since the clinical trial itself did not include many younger women for whom treatment could be beneficial, the clinically relevant policy learning question is determining the optimal 
tailoring of targeted treatments
such that the clinical benefits of HRT do not also incur substantial increase in risk of CHD and other adverse events. Further heterogeneity in age arises due to clinical explanations that estrogen may slow down \textit{early} artherosclerosis, the formation of plaques in arteries, and have favorable endothelial effects in women with recent onset in menopause. However, unlike other options such as statin therapies which help prevent CHD at any age and stage of disease, HRT may actually worsen already-established plaques and thus increase the frequency of coronary events in older women \citep[see][]{roussouw-whi-13,Manson2013}.  
}
{
\subsubsection{WHI Case Study Evaluation Setup and Outcome Measures} 
\paragraph{Dataset details.} 
We restrict attention to a complete-case subset of the WHI clinical-trial data ($n=13594$) and a complete-case subset of the observational study ($n=48458$), obtained after dropping the cardiac arrest covariate (which is mostly missing). 
$T=1$ denotes treatment with combined estrogen-plus-progestin hormone replacement therapy.
\paragraph{Outcome variable.} 
We define our outcome variable to account for cardiovascular health as well as the clinical benefits of HRT for menopause symptoms, and we range over the potential combinations of these to study the changing behavior of our method. Specifically, letting $S$ denote systolic blood pressure and given $\lambda<0$, we define our outcome as 
$$Y = S + T\lambda.$$
We vary $\lambda$ in a grid on $[-0.5,-1.5]$. Every $\lambda$ generates a new dataset, on which
we learn policies from the observational study using our framework, varying the sensitivity parameter $\Gamma$, and estimate the outcomes of these learned policies on the actual randomized WHI clinical trial data. 
In training on the observational dataset, nominal propensities are estimated using logistic regression.
We assess our methods and appropriate baselines, which learn from the confounded observational data, and evaluate their performance on the clinical trial dataset, with constant treatment arm randomization probabilities. 
This demonstrates the range of possible behaviors as treatment becomes overall more or less beneficial and offers a sensitivity analysis of our method to different scalarizations of clinical benefits with the blood pressure outcome. Our case study illustrates the improvements of robustness compared to naively learning from confounded data. 
}
{ 
	\paragraph{Clinical trial evaluation.} Without access to the true counterfactual outcomes for patients, 
	we evaluate the performance of policies out of sample by using 
	an unnormalized Horvitz-Thompson estimator on the held-out
	truly-randomized data from the WHI clinical trial. 
	As reported above, treatment was randomized at $1/2$ probability.
	Correspondingly, our out-of-sample estimate of policy regret relative to a \textit{control} baseline, $\pi_0(0 \mid x)=1$, is given by\footnote{{Note that the actual realized fraction treated in the dataset are 0.502 so the estimate is also nearly equal to the corresponding H\'ajek estimator.}}
	\[\hat R_{\pi_0}^\text{test}(\pi)= \frac{1}{{n}} \sum_{i=1}^{n}  \frac{Y_i}{p_{T_i}} (\mathbb{I}[T_i = 0]  (\pi(0 \mid X_i) -1)  + \mathbb{I}[T_i = 1]  \pi(1 \mid X_i)   )  \]
}
\subsubsection{WHI Case Study Policy Learning Results}

\begin{figure}
	\includegraphics{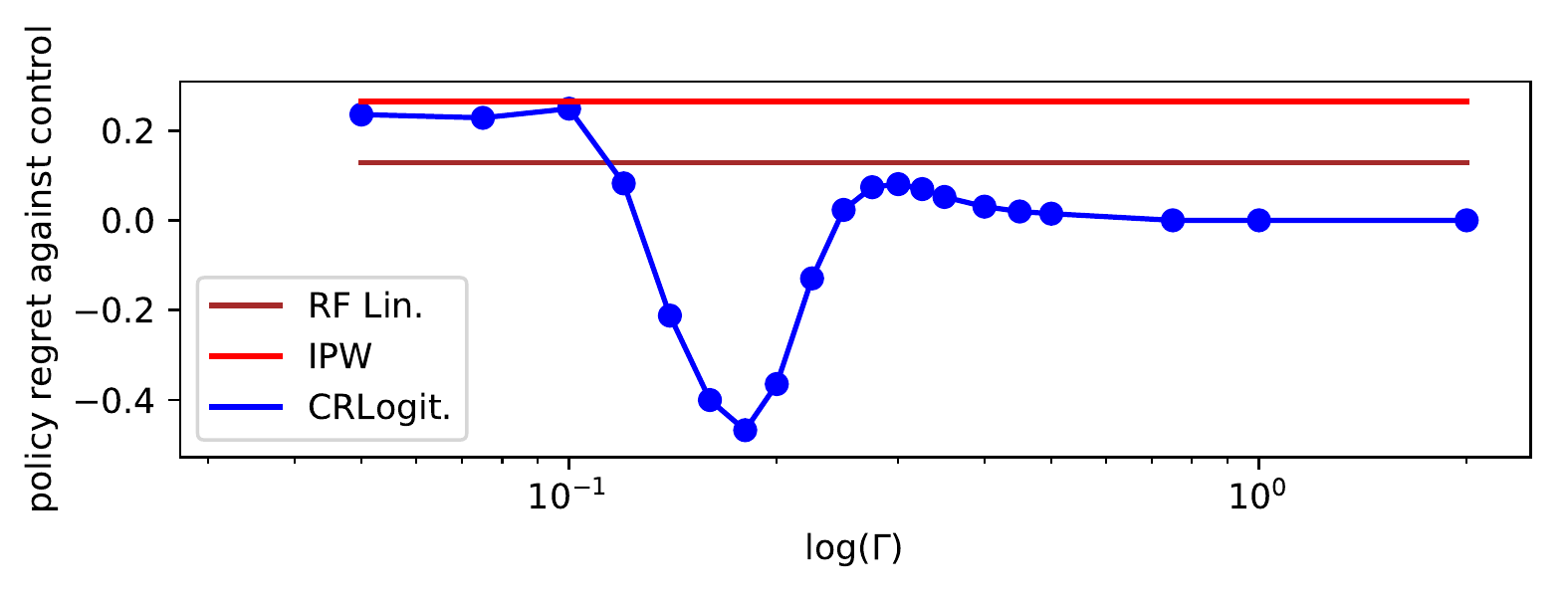}
	\caption{Plots of out of sample regret on WHI case study data for a single treatment effect scalarization, $\lambda=-0.93$  }
	\label{fig:whi-lambdachoice} 
\end{figure}
\begin{figure}\centering
		\includegraphics[width=0.6\textwidth]{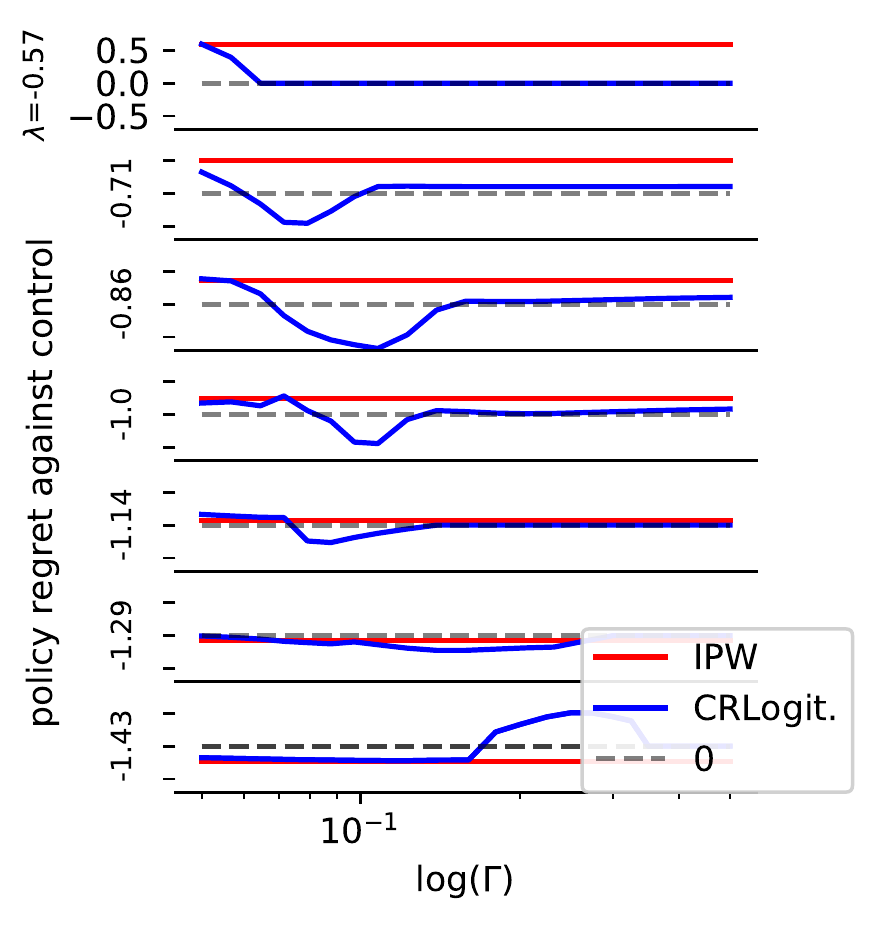}
		\caption{Plots of out of sample regret, sensitivity analysis on scalarizations of WHI data, varying $\lambda$}
		\label{fig:whi-sparklines}
\end{figure}
\begin{figure}%
	\begin{subfigure}{0.5\textwidth}
		\includegraphics[width=\textwidth]{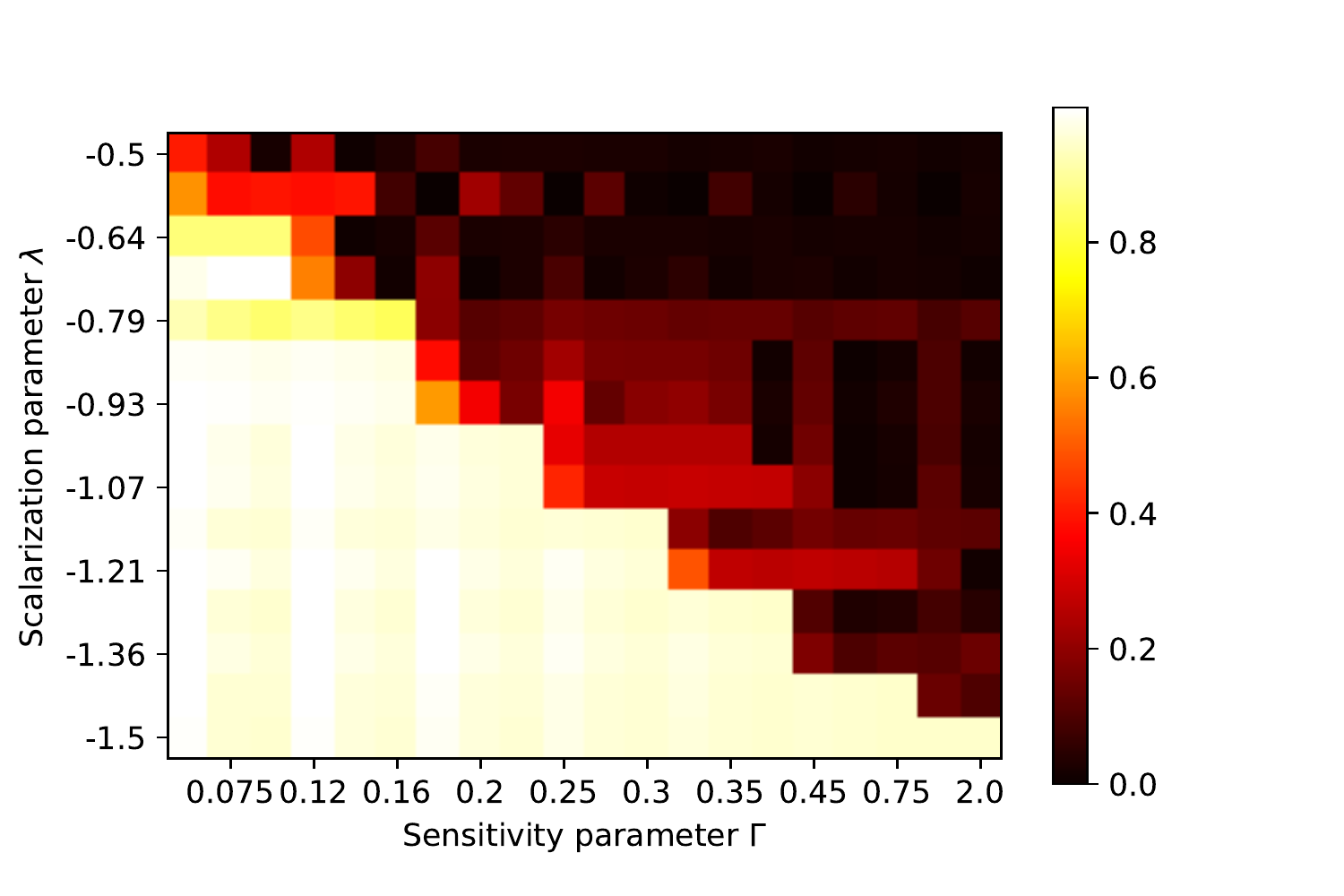}
		\caption{Heatmap of percentage of patients \\ treated with probability $\pi(1\mid X) > 0.5$ (Sec.~\ref{sec-whi})}
		\label{fig-whi-treated}          
	\end{subfigure}
	\begin{subfigure}{0.5\textwidth}
		\includegraphics[width=\textwidth]{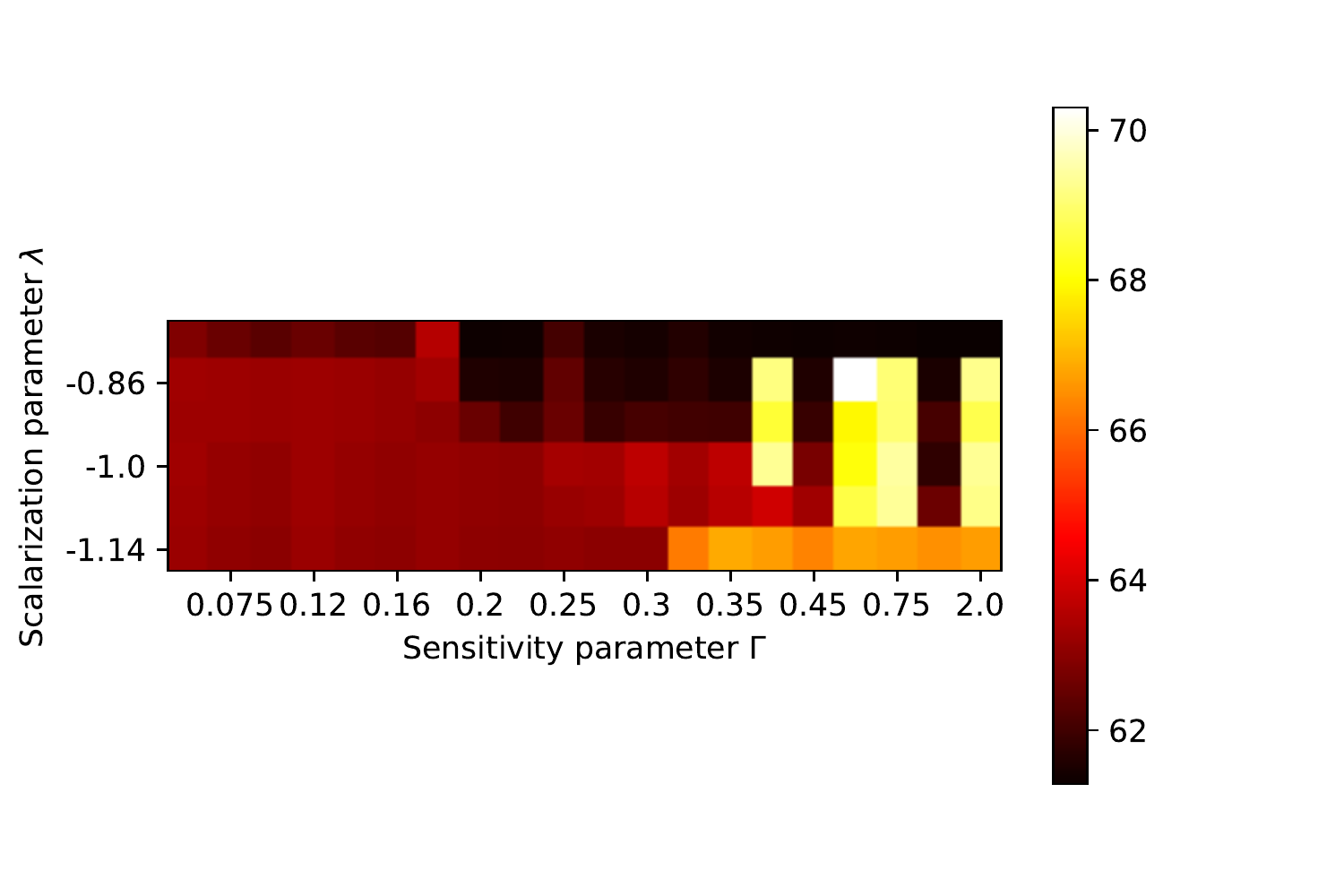}
		\caption{Heatmap of $\E[ \text{age} \mid \pi(1\mid X)>0.5]$, robust policies with scalarized WHI data (Sec.~\ref{sec-whi})}
		\label{fig-whi-age}          
	\end{subfigure} %
\end{figure}

{
We compare to two baselines that assume unconfoundedness:
the logistic policy minimizing the IPW estimate with nominal propensities (IPW) and the same with policy value estimates gotten by estimating $\op{CATE}$ using
causal forests \citep[GRF Lin;][]{wager2017estimation}. 
In \Cref{fig:whi-lambdachoice} we display a (favorable) treatment effect scalarization, $\lambda=-0.93$, where our policy, for certain values of the sensitivity parameter, indeed finds benefit, while linear policies using only estimated propensity scores or confounded outcome regression (IPW and RF lin.) still incur relative harm relative to the all-control baseline.}

{Of course, whether or not our approach finds relative improvement (or if the robust approach is overly conservative), depends on the exact treatment effect scalarization parameter $\lambda$. In \Cref{fig:whi-sparklines}, we include a comprehensive comparison of the relative performance of our approach, IPW, and the control baseline, for a various levels of $\lambda$. (We report full numerics in \Cref{tbl-whi-lambdas-crlogit} in the appendix.) This shows that when treatment is almost uniformly bad (less negative $\lambda$), our approach avoids the harm done by policy learning assuming unconfoundedness but it cannot offer improvement over control as it is likely indeed best to treat no one. When the treatment is almost uniformly good (more negative $\lambda$), as our approach defaults to less treatment it achieves worse performance than policies that treat often such as IPW. In between, for moderate values of $\lambda$, treatment effect is heterogeneous and while approaches that assume unconfoundedness introduce harm, our approach achieves both safety and, for moderate values of $\Gamma$, improvement.
}

{
 In any particular setting, we do not have \textit{a priori} knowledge on quantitatively which improvement regime is possible; our analysis for changing $\lambda$ is intended to partly illustrate the fuller range of qualitative performance under our estimator. When there is unobserved confounding but little room for improvement (e.g. where $\lambda=0$, with only high blood pressure as the outcome), our method may simply default to baseline for smaller levels of $\Gamma$. For moderate regions of benefit ($\lambda \in [-0.7,-1]$), confounded linear policies perform poorly while our approach recovers regions of improvement (where the blue line is below the dotted line of $0$ regret). Where our robust approach is poorly specified (very negative $\lambda \in [-1,-1.5]$), even confounded IPW with a linear policy achieves benefit, and robustly defaulting to baseline is a worse strategy than using confounded IPW. 
}
{In \Cref{fig-whi-treated}, we interpret the range of policies by plotting the percentage of individuals treated under each $\log(\Gamma)$ value on the x-axis, ranging over scalarization parameters $\lambda$ (on the y-axis). In \Cref{fig-whi-age}, we plot the average age conditional on being treated with probability greater than 0.5 under a confounding-robust policy. For regions of moderate improvement, the confounding-robust policies tend to treat younger patients on average. (Artifacts arise when assessing age conditional on treating very few people). 
}

{\blockedit
\section{Practical Considerations in Calibrating Uncertainty Sets}\label{apx:rbars}

In the above we demonstrated the performance of our method as we vary the parameter $\Gamma$ controlling the amount of allowed confounding. A remaining important question is how should $\Gamma$ be chosen in practice. Below we review recommendations from traditional sensitivity analysis and then we propose an approach specifically designed for the policy learning problem.

\subsection{Comparison to effect of observed covariates on treatment selection}
As mentioned in \Cref{sec:optminimaxregret,sec:relatedwork},
one broad strategy for calibrating a sensitivity model benchmarks the level of unobserved confounding relative to the informativeness of observed covariates for selection into treatment \citep{hsu2013calibrating}.
For example, 
 we can compute the effect of omitting each observed covariate on the odds ratio of the propensity score. A decision-maker could use domain knowledge to assess whether there are plausible unobserved confounders that could have as large an effect as the observed one, suggesting a plausible upper bound on $\Gamma$. While in traditional sensitivity analysis this suggests how large $\Gamma$ one should consider in testing the robustness of one's inferences, in the context of policy learning, this suggests what amount of confounding should one be concerned with protecting against to ensure no harm. In the next section, we discuss how to combine this strategy with calibration plots, which we develop, to make an informed choice about which $\Gamma$ to choose for training a policy.

 To illustrate this benchmarking in the WHI case study, we plot the odds ratios induced by dropping different variables in \Cref{fig:whi-joyplot} in the Appendix.
 This shows that, aside from variables such as age that are highly predictive of treatment selection, the induced odds ratios are safely bounded by $\Gamma$ somewhere in range of $0.8$ to $1.2$.
 Therefore, if we believe our omitted confounders cannot be as informative as age, we should consider the safety of our policy for confounding levels as large as $\Gamma$ in the range of $0.8$ to $1.2$.

\subsection{Calibration plots}

Next, we propose a tool to visualize the trade-offs between choosing too-high or too-low a value for $\Gamma$. 
Choosing too-high $\Gamma$ leads to better uniform control on regret on a larger range of potential confounding, but may be conservative if the actual confounding was in fact controlled by a smaller $\Gamma$, while too-low a value of $\Gamma$ achieves worse uniform control over a larger range of potential confounding.
We propose to analyze this by considering the estimated worst-case regret over confounding bounded by $\Gamma'$ incurred by a policy learned using a parameter $\Gamma$, that is, $\hat{\overline{R}}_{\pi_0}(\hat{\overline\pi}(\Pi,\mathcal W_n^{\Gamma},\pi_0); \mathcal W_n^{{\Gamma}'})$.
\begin{figure}
	\includegraphics[width=\textwidth]{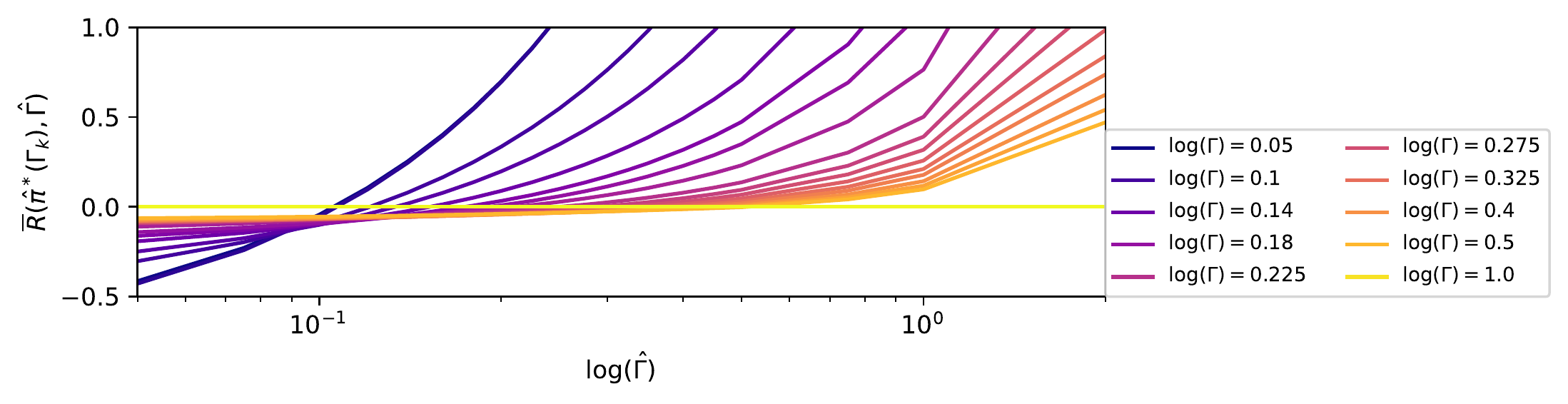}
	\caption{Calibration plot for WHI case study: 
	for each parameter choice $\Gamma$,
	the curve shows the possible estimated worst-case regret when confounding may be as large as $\Gamma'$, as we vary $\Gamma'$.
	}
	\label{fig:calibration}
\end{figure}

{
Specifically, we propose to visualize this in a callibration plot produced thus:
}
{
\begin{itemize}
\item Fix a sequence of $\Gamma$ values, $\Gamma_1, \dots , \Gamma_K$. 
\item For every $k \in \{1, \dots, K\}$, train a confounding-robust policy under parameter $\Gamma_k$, $\hat{\overline\pi}_k=\hat{\overline\pi}(\Pi,\mathcal W_n^{\Gamma},\pi_0)$. 
\item For every $k,k' \in  \{1, \dots, K\}$, evaluate the minimax regret estimate under parameter $\Gamma_{k'}$, $\hat{\overline R}_{k,k'}=\hat{\overline{R}}_{\pi_0}(\hat{\overline\pi}_{k}; \mathcal W_n^{{\Gamma}_{k'}})$. 
\item For each $k$, plot $\hat{\overline{R}}_{k,k'}$ against $\Gamma_k$.
\end{itemize}
}
An example of such a calibration plot for our WHI case study is given in \Cref{fig:calibration}.

First, this plot shows how the regret of a policy trained with one $\Gamma$ may grow and possibly become positive if the true confounding may correspond to a larger $\Gamma'>\Gamma$. In the example of \Cref{fig:calibration}, for very small $\Gamma=1.05$, we see that the policy (which essentially assumes unconfoundedness) may incur large regret for even small values of $\Gamma$ in the range of 1.1 to 1.2. Since these values are smaller than the ranges of $\Gamma$ we found by considering the informativeness of observed variables, if we may have omitted a variable as important as these, we may be concerned about the safety of policies learned using such small values of $\Gamma$. Second, as we increase $\Gamma$ we find that we obtain uniform control on regret even for confounding corresponding to larger $\Gamma'$. We may, however, pay in terms of performance if confounding were in fact smaller. We can assess this using the plot, which shows us the deterioration in performance for smaller levels of confounding, $\Gamma'<\Gamma$, relative to policies that are trained with lower $\Gamma$, potentially even policies that are trained assuming unconfoundedness ($\Gamma=1$). In the example of \Cref{fig:calibration}, we find that using $\Gamma=1.14$ may offer safe control on regret for $\Gamma'$ up to 1.2, ensuring no harm in the ranges deemed of potential concern, while it would cause only minimal inefficiencies if confounding were really smaller relative to policies that would somehow exploit this fact.
Thus, calibration plots allow one to assess the trade-offs of safety and performance and choose a policy that best fits the requirements of the application domain. 
}

\section{Conclusion} 

In this paper, we addressed the problem of learning personalized intervention policies from observational data with unobserved confounding. Standard methods can be corrupted by this confounding and lead to harm compared to current standards of care, a concern of utmost importance in sensitive applications such as medicine, public policy, and civics. We therefore develop a framework for confounding-robust policy improvement, which optimizes personalization policies in view of possible unobserved confounding in observational data, allowing for more reliable and credible policy evaluation and learning. Our approach optimizes the minimax regret achieved by a candidate personalized decision policy against a baseline policy. We generalize the class of IPW-based estimators and construct uncertainty sets centered at the nominal IPW weights that can be calibrated by approaches for sensitivity analysis in causal inference. A future line of investigation is a confounding-robust variant based on the doubly robust estimator of policy value.

We prove a strong statistical guarantee that, if the uncertainty set is well-specified, our approach is guaranteed to do no worse than the standard of care so that it can be safely implemented, and possibly offer improvement if the data can support it. Specifically, the result proved a finite-sample guarantee that can be checked. We leverage the optimization structure of weight-normalized estimators of the policy value to perform policy optimization efficiently by subgradient descent on the robust risk and
{ 
we provide uniform convergence bounds showing that our approach achieves the population level minimax-optimal regret. Assessments on synthetic and clinical data demonstrate the benefits of confounding-robust policy improvement, which can recommend personalized treatment while maintaining strong guarantees on performance relative to baseline preferences. 
}
These tools allow an analyst to 
find reliable and personalized policies that can safely offer improvements even if there is unobserved confounding and to
assess the different plausible levels of confounding on the performance of a robust personalized decision policies.
We believe this development is absolutely crucial for the practical adoption of algorithms for personalization that work on the ever growing repositories of observational data, which are the future of algorithmic decision making due to their size and richness.

\ACKNOWLEDGMENT{This material is based upon work supported by the National Science Foundation under Grant No. 1656996.
Angela Zhou was supported through the National Defense Science \& Engineering Graduate Fellowship Program.}

{
\bibliography{sensitivity}
\bibliographystyle{abbrvnat}
}

 \clearpage

\ECSwitch

\ECHead{Supplemental Material}

\section{Proofs for optimization structure} 
\proof{Proof of the equivalence of programs \eqref{opt-FLPP} and \eqref{opt-LPP}.}\label{pf-equiv}
{We can easily verify that a feasible solution for one problem is feasible for the other: for a feasible solution $W$ to (FP), we can generate a feasible solution to (LP) as $w_i = \frac{W_i}{\sum_i W_i}, \ccpscalar = \frac{1}{\sum_i W_i }$ with the same objective value. In the other direction, we can generate a feasible solution to (\ref{opt-LPP}) from a feasible fractional program (\ref{opt-FLPP}) solution $W,\ccpscalar$ if we take $W_i = \frac{w_i}{ \ccpscalar}$. This solution has the same objective value since $\sum_i w_i = 1$.}
\endproof
\proof{Proof of Thm.~\ref{thm-norm-wghts-soln}.}\label{pf-norm-wghts-soln}
We analyze the program using complementary slackness, which will yield an algorithm for finding a solution that generalizes that of \citet{aronowlee12}.
{
At optimality only one of the primal weight bound constraints, (for nontrivial bounds $a^\Gamma <b^\Gamma$), $w_i \leq \ccpscalar b_i^\Gamma$ or $\ccpscalar a_i^\Gamma \leq w_i$ will be tight.}
For the nonbinding primal constraints, at the optimal solution, by complementary slackness the corresponding dual variable $u_i$ or $ v_i$ will be 0. Since at least $n +1$ constraints are active in the dual, the constraint $\sum_i  -b_i v_i +a_i u_i \geq 0$ is also active. So the optimal solution to the dual will satisfy: 
\begin{align*}
&\min \lambda\\
&\text{ s.t. } \lambda \geq r_{ i} + u_i - v_i,  \;\forall i \in 1,\dots,n\\
&\sum_i  -b^{\Gamma}_i v_i +a^{\Gamma}_i u_i = 0 
\end{align*}
By non-negativity of $u_i, v_i$, note that $u_i > 0$ if $r_i < \lambda$ and $v_i > 0$ if $r_i > \lambda$ such that $u_i = \max( 0, \lambda - r_i)$ and $ v_i = \max(0, r_i -\lambda)$. Additionally, feasible objective values satisfy $\lambda \leq  \max_i{Y_i}$ and $\lambda \geq \min_i{Y_i}$.  Let $(k)$ denote the $k$th index of the increasing order statistics, an ordering where $r_{(1)} \leq r_{(2}) \leq \cdots \leq r_{(n)}$. Then at optimality, there exists some index $(k)$ where $Y_{(k)} < \lambda \leq Y_{(k+1)}$. We can subsitute in the solution from the binding constraints $\lambda = r_i + u_i - v_i$ and obtain the following equality which holds at optimality: 
{
\begin{align*}& \ccpscalar \underset{i: (i) < (k) }{\sum} a_{(i)}^\Gamma  (\lambda - r_{(i)}) -  \ccpscalar\underset{{i: (i) \geq(k)}}{\sum}  b_{(i)}^\Gamma (r_{(i)} - \lambda ) = 0 
\end{align*}
Rearranging, we have that 
}
\begin{align*}
& 	\lambda_{(k)} = \frac{   \underset{i: (i) < (k) }{\sum}a_{(i)}^\Gamma r_{(i)} + \underset{i: (i) \geq (k)}{\sum} b_{(i)}^\Gamma r_{(i)}}{ \underset{{i: (i) < (k)}}{\sum} a_{(i)}^\Gamma + \underset{{i: (i) \geq(k)}}{\sum} b_{(i) }}
\end{align*} 

Therefore, we only need to check the possible objective values $\lambda_{(k)}$ for $k=1,\dots,n$.	The primal solution is easily recovered from the dual solution: for $r_{(i)}$, take $w_{(i)} =\frac{ a_{(i)}^\Gamma \mathbb{I}\{ (i) \leq k \}+b_{(i)}^\Gamma \mathbb{I}\{ (i) > k \}}{  \underset{i: (i) < (k) }{\sum} a_{(i)}^\Gamma +\underset{{i: (i) \geq(k)}}{\sum} b_{(i)}^\Gamma }$ and $t={  \underset{i: (i) < (k) }{\sum} a_{(i)}^\Gamma +\underset{{i: (i) \geq(k)}}{\sum} b_{(i)}^\Gamma }$.
{
Consider the parametric restriction of the primal program, where it is parametrized by the sum of weights $\ccpscalar$: the value function is concave in $\ccpscalar$ and concave in the discrete restriction of $\ccpscalar$ to the values it takes at the solutions of $\lambda_{(k)}$, $\ccpscalar_{(k)}$, and $\ccpscalar_{(k)}$ is increasing in $k$. So the optimal such $\lambda$ occurs with the order statistic threshold at $(k)$ for  $k^* = \inf \{k=1,\dots,n+1: \lambda({k+1}) < \lambda({k}) \}$. }

{
	Lastly, we discuss the case where $Y$ may be discrete, or if it is distributed as a mixture of a continuous density and atoms. Our characterization of the optimization solution as monotonic and also a function of the sort order on $Y$ implicitly assumes that outcomes $Y$ are generated from a continuous density, so that $Y_i =Y_j$ with probability 0. Our analysis, too, requires this. We show that in the case where tiebreaking among $Y$ is required, there is a natural lexicographic order. 
	Let $(1), ... (i) ... (n)$ denote the ordering that is lexicographically increasing in $(Y_{(i)}, b_{(i)}  - a_{(i)})$: when outcomes $Y$ are discrete, the appropriate sort order includes the weights $b_{(i)}  - a_{(i)}$. Denote the coefficients as $r_i$ and $b_{(i)}  - a_{(i)}$. 
	Suppose that for a given sort order, the optimum is achieved at $\lambda(k)$. We show that the lexicographic sort order, sorting first in $y$ and then increasing in $r \Delta$, preserves the unimodality property. Suppose $y_k$ is the same for some interval $[k, k+j]$: we want to show that the discrepancy $\lambda(k) - \lambda(j)$ is increasing in $i, i \leq j$. 	Denote $n(k) = \sum_{i \le k} r_i a_t(X_i)Y_i +  \sum_{i \ge k+1}r_i b_t(X_i)Y_i$ and $d(k) = \sum_{i \le k} r_ia_t(X_i) +  \sum_{i \ge k+1} r_i b_t(X_i)$.
	Then, 	\begin{align*}
	&	\lambda(k) - \lambda(k+1)  = \frac{n(k)}{d(k) } - \frac{n(k) - \Delta y_{k+1}  r_{k+1} }{ d(k) - \Delta_{k+1} r_{k+1}} \\
	&= \frac{n(k) (d(k) - \Delta_{k+1}r_{k+1}) -(n(k) - \Delta_{k+1}y_{k+1} r_{k+1})d(k)  }{d(k) ( d(k) - \Delta_{k+1}r_{k+1} )} \\
	&= \lambda(k)  \frac{\Delta_{k+1} r_{k+1} }{ d(k) - r_{k+1} \Delta_{k+1} } - \frac{ \Delta_{k+1} y_{k+1} r_{k+1} }{ d(k) - \Delta_{k+1}r_{k+1} }\\
	& = \frac{\lambda(k) - y_{k+1}}{\frac{d(k)}{r_{k+1}\Delta_{k+1} } - 1 }
	\end{align*}
	We show that if this difference changes sign, it continues to decrease: if $\lambda(k)\leq \lambda(k-1)$, and if $y_{k} = y_{k+1}$, then $\lambda(k+1) < \lambda(k)$. By the above analysis, telescoping the finite difference $\lambda(k) - \lambda(k+1)$, 
	$$ \lambda(k) - \lambda(j) = \sum_{i=1}^j   \frac{\lambda(k+i) - y_{(k+i+1)}}{\frac{d(k+i)}{r_{(k+i+1)}\Delta_{(k+i+1)} } - 1 } =  \frac{\lambda(k) - y_{k+j}}{\frac{d(k)}{r_{k+j}\Delta_{k+j} } - 1 } $$
	so that where $y_{(k)} = y_{(k+j)}$, $\lambda(k) - \lambda(j)$ is decreasing as $r_{k+j} \Delta_{k+j}$ increases.
}
\endproof
\vspace{20pt} 
\blockedit{
\proof{Proof of \Cref{prop:monotoneweights}}\label{proof:monotoneweightsU}
We show via a similar argument to \Cref{thm-norm-wghts-soln} that the linear program under the one-to-one change of variable $W_i = a(X_i) + (b(X_i) - a(X_i))u_i$, where $u_i \in [0, 1]$, has a similar solution structure in the variable $u_i$: that the optimal weights $u_i^*$ satisfy that $u_i^* = u(Y_i(\pi(T_i\mid X_i) - \pi_0(T_i \mid X_i) ))$ for some function $u: \mathcal{Y} \to [0, 1]$ such that $u(u(\pi(t\mid x) - \pi_0(t\mid x)))$ is nondecreasing in $y(\pi(t\mid x) - \pi_0(t\mid x))$. 
Define vectors $\alpha, \beta$ such that $\alpha_i = Y_i (\pi(T_i \mid X_i ) - \pi_0(T_i \mid X_i) )(b(X_i) - a(X_i))\mathbb{I}[T_i = t]$ and $\beta_i  = (b (X_i) - a(X_i)) \mathbb{I}[T_i = t]$, and constants $c = \sum_{i: T_i = t}^n a (X_i) Y_i(\pi(T_i \mid X_i ) - \pi_0(T_i \mid X_i) ), d = \sum_{i: T_i = t}^n a (X_i) $.
\begin{align*} 
\max\limits_{U} & \frac{\alpha^\top u + c}{\beta^\top u + d} \\
\text{s.t. } &
0 \leq u \leq 1
\end{align*}
By applying the Charnes-Cooper transformation with $\tilde u = \frac{u}{\beta^\top u + d}$ and $\tilde v = \frac{1}{\beta^\top u + d}$, the linear-fractional program above is equivalent to the following linear program: 
\begin{align*}
\max\limits_{\tilde u, v} & \; \; \alpha^\top \tilde u + c \tilde{v}\\
\text{s.t.} & \; \; 0 \leq u \leq \tilde v\\
&\beta^\top \tilde u + \tilde v d = 1, \tilde v \geq 0 
\end{align*}
where the solution for $\tilde u, \tilde v$ yields a solution for the original program: $\tilde{u}_i$ is such that $u_i = \frac{\tilde u_i}{\tilde v}$. 

Let the dual variables $p_i \geq 0$ be associated with the primal constraints $\tilde u_i \le \tilde v$ 
(corresponding to $u_i \leq 1$), $q_i \geq 0$ associated with $\tilde u_i \ge 0$ (corresponding to $u_i \geq 0$), and $\lambda$ associated with the constraint $\beta^\top \tilde u + d\tilde v  = 1$. 

The dual problem is: 
\begin{align*}
\min_{\lambda, p, q} \{  \lambda \colon p - q + \lambda \beta = \alpha , -1^\top p + \lambda d \geq c ,p_i \geq 0, q_i \geq 0 \}
\end{align*}
By complementary slackness, at most one of $p_i$ or $q_i$ is nonzero. For brevity, let $r_i = Y_i(\pi(T_i\mid X_i) - \pi_0(T_i \mid X_i))\mathbb{I}[T_i = t]$. Rearranging the first set of equality constraints gives $p_i - q_i = \mathbb{I}[T_i = t](b(X_i) - a(X_i))( r_i ) - \lambda)$, which implies that 
\begin{align*}
p_i &= \mathbb{I}[T_i = t] (b(X_i) - a(X_i))\max(r_i - \lambda, 0) , \qquad q_i =\mathbb{I}[T_i = t] (b(X_i) - a(X_i))\max(\lambda - r_i, 0)
\end{align*}
Since the constraint $-\vec{1}^\top p + \lambda d \geq c$ is tight at optimality (otherwise there exists smaller yet feasible $\lambda$ that achives lower objective of the dual program), 
\begin{align*}
\sum_i \mathbb{I}[T_i = t] (b(X_i) - a(X_i))\max(r_i - \lambda, 0) = \sum_i \mathbb{I}[T_i = t]  a (X_i) (r_i-\lambda)
\end{align*}

This rules out both $\lambda > \max_i{r_i}$ and $\lambda < \min_i{r_i}$, thus $r_{(k)} < \lambda \leq r_{(k+1)}$ for some $k$ where $r_{(1)}, r_{(2)}, \dots, r_{(n)}$ are the order statistics of the sample outcomes. This means that $q_i > 0$  can happen only when $r_i \le r_{(k)}$, i.e., $u_i = 0$; and $p_i > 0$ can happen only when $i > k + 1$, i.e., $u_i = 1$. Applying this, we may rewrite the above expression to recover that the optimal $\lambda$ must be one of $\lambda_{(k)}$. This proves that the structure of the optimal solution is such that there exists a nondecreasing function $u: \mathcal{R} \to [0, 1]$ such that $u_i = u(Y_i(\pi(T_i\mid X_i) - \pi_0(T_i \mid X_i)) \mathbb{I}[T_i = t])$ attains the upper bound. 
\qed
\endproof
}

\clearpage

\section{Proofs of Uniform Convergence Guarantees}
\subsection{Uniform convergence: tail inequalities} 
{
	In this subsection, we introduce definitions and stability results from empirical process theory in order to keep the argument self-contained, and provide maximal inequalities for the function classes of interest: $\Pi$, and reparametrizations of the optimal weight functions, $\Wgamma, \overline{\mathcal W }^\Gamma$. }
{ We will work with the \textit{packing and covering numbers} of $\Pi$ and the spaces of weight functions, and then relate these to bounds on the VC-major dimension of the policy class. For a subset $S$ of some metric space, the packing number $D(\epsilon, S)$ is the largest number of points we can take in $S$ that are not within $\epsilon$ distance of one another, and the covering number $N(\epsilon, S)$ is the smallest number of points we need to take in $S$ so that every other point is within $\epsilon$ of one of these \citep{pollard1990empirical}.\footnote{{The packing and covering numbers are closely related by the inequality $N(\epsilon, t_0) \leq D(\epsilon, T_0) \leq N(\epsilon/2, t_0)$.}}}
{ 
First we introduce the stability results from empirical process theory which will yield bounds on the covering numbers for the function classes of interest. We define the class of
\textit{VC-hull} functions, 
broader than VC-subgraph and related to VC-major, but which result in bounded Dudley entropy integrals. 
\begin{definition}[VC-hull class]\label{def-vchull}
Define ${\op{conv}}(\mathcal{F}) $, the convex hull of $\mathcal{F}$:
\begin{align*}
\op{conv}(\mathcal{F}) = \{ \sum_{f \in \mathcal F} \lambda_f  f \colon  f\in \mathcal F, \sum_f \lambda_f =1, \lambda_f \geq 0, \lambda \neq 0 \text{ for finitely many } f \} 
\end{align*}
$\overline{\op{conv}}(\mathcal{F}) $ is the pointwise sequential closure of the convex hull of $\mathcal{F}$: $f \in \overline{\op{conv}}(\mathcal{F}) $ if there exist $f_n \in \op{conv}(\mathcal F)$ such that $f_n(x) \to f(x)$ for all x in the domain of $f$, as $n \to \infty$. 
If the class $(\mathcal{F}) $ is VC-subgraph, then $\overline{\op{conv}}(\mathcal{F}) $ is a VC hull class of functions. 
\end{definition}
A bounded VC-major class is a VC-hull class. Since VC-hull classes are defined with respect to the sequential closure of the convex hull ($\overline{\op{conv}}(\mathcal{F})$) of another function class $\mathcal{F}$, we frequently refer to $\mathcal{F}$ as the \textit{generating} VC-subgraph class for its corresponding VC-hull class. VC-hull classes provide a constructive definition for a VC-major class in relation to a VC-subgraph class, and satisfy the following bound on the entropy integral of the covering numbers: 
\begin{theorem}[Theorem 2.6.9 of \cite{van1996weak}; originally \cite{ball1990entropy}]\label{thm-dudley-ball-pajor}
Suppose there is a class of functions $\mathcal{F}$, with measurable square integral envelope $F$ with bounded second moments, that is VC-subgraph of dimension $V$, such that 
$D(\epsilon \norm{F}_2, \mathcal{F}, \norm{\cdot}_2  ) \leq C \left(\frac{1}{\epsilon}\right)^V.$
	Then, for $\overline{\op{conv}}(\mathcal{F}) $, the closure of the convex hull of $\mathcal{F}$ (e.g. the VC-hull class that is generated by $\mathcal{F}$), there exists a universal constant $K$ depending on $C$ and $V$ only such that 
	$$\log  D(\epsilon \norm{F}_2, \mathcal{F}, \norm{\cdot}_2  ) \leq K \left(\frac{1}{\epsilon}\right)^{2V/(V+2)}$$
\end{theorem}
Working with VC-major (equivalently, VC-hull) classes allows us to use stronger stability results such as the following stability result on the stability of composition of the class of monotone functions and VC-major function classes, though at the expense of introducing a universal constant in the Dudley entropy integral.
	\begin{lemma}[Proposition 4.2 of \cite{dudley1987universal}]\label{lemma-vcmajor-stability-monotone-composition}
		If $\mathcal H$ is a VC-major class for the generating class $\mathcal{C}$, and $\Und$ denotes the set of all nondecreasing functions from $\mathbb{R}\mapsto [0,1]$, and 
		$$ \mathcal{F} \defeq \{  u \circ h \colon h \in \mathcal{H}, u \in \Und \}, $$ 
		then $\mathcal{F}$ is a major class for the monotone derived class $\mathcal{D}$ of $\mathcal{C}$. Therefore if $\mathcal{H}$ is a VC-major class, so is $\mathcal{F}$.
	\end{lemma}
}

{
	These stability results allow us to prove, e.g. \Cref{prop:vcmonotone}, that it is sufficient to restrict to optimizing over the set of worst-case weights (with additional structure).
\proof{Proof of \Cref{prop:vcmonotone}}
The result follows from \Cref{prop:monotoneweights} by applying \Cref{lemma-vcmajor-stability-monotone-composition}.
 \qed
\endproof
}
{
	We introduce the uniform convergence setup we use to provide tail inequalities. We will apply a standard chaining argument with Orlicz norms and introduce some notations from standard references, e.g. \cite{vershynin2018high,pollard1990empirical,wainwright2019high}. A function $\phi: [ 0,\infty) \to [ 0,\infty)$ is an Orlicz function if $\phi$ is convex, increasing, and satisfies $\phi(0) = 0, \phi(x) \to \infty$ as $x\to \infty$. For a given Orlicz function $\phi$, the Orlicz norm of a random variable $X$ is defined as $ \norm{ X }_\phi = \inf \{ t> 0 \colon \E[ \Phi(\norm{X}\mid t ) ] \leq 1 \}  $. The Orlicz norm $\norm{Z}_\Phi$ of random variable $Z$ is defined by $$\norm{Z}_\Phi = \inf \{  C > 0 \colon \E[ \Phi( Z / C) ] \leq 1  \}. $$
	A constant bound on $\norm{Z}_\Phi$ constrains the rate of decrease for the tail probabilities by the inequality $ \pr( \vert Z \vert \geq t ) \leq  1 /\Phi(t/C) $ if $C = \norm{Z}_\Phi$. For example, choosing the Orlicz function $\Phi(t) = \frac{1}{5} \exp(t^2)$ results in bounds by subgaussian tails decreasing like $\exp(-C t^2)$, for some constant $C$.
}

We next introduce the tail inequalities that use a standard chaining argument to control uniform convergence over $\pi \in \Pi$ and appropriate reparametrizations of the weight functions.
First we define the following function classes conditional on all the data, $(X_{1:n},T_{1:n}, Y_{1:n})$. 
For $\pi$, we consider a shifted function class with an envelope function: let $ f_i(\pi) =  (\pi(T_i\mid X_i)-\frac{1}{m})  Y_i  $ where
$$\mathcal{F}(X_{1:n}, T_{1:n},Y_{1:n}) = \{ (f_1(\pi), \dots , f_{n}(\pi)) \colon \pi \in \Pi  \}. $$
Next we introduce function classes for the weight functions: the minimax-regret achieving functions $W \in \overline{\mathcal W}^\Gamma(\pi)$ may also be written as compositions of the nominal weight functions with a function $u$, $$W \circ u(\pi) = a_t^\Gamma(x)+u(y(\pi(t\mid x)-\pi_0(t\mid x))) (b_t^\Gamma(x)-a_t^\Gamma(x)),$$ where $ u \in \Ugamma(\pi)$, the class of nondecreasing functions in the index $y (\pi(t\mid x) - \pi_0(t\mid x) )$ for $\pi_0 = \frac{1}{m}$ and a fixed policy $\pi$, defined as the following: 
$$\Ugamma(\pi) = \left\{ 
\begin{matrix}
u(x,t,y): \mathbb{R} \mapsto [0,1] : 
u(y(\pi(t\mid x) - \pi_0(t \mid x)))	\text{ is monotonic nondecreasing } 
\end{matrix} \right\}.$$
Analogously, we let $\overline{\mathcal U}^\Gamma= \cup_{\pi \in \Pi} \; \Ugamma(\pi) $ denote the set of nondecreasing functions on the same index, but now ranging over $\pi \in \Pi$. 
Clearly, optimizing over $\overline{W}^\Gamma$ is equivalent to optimizing over $\overline{\mathcal U }^\Gamma$: 
\begin{corollary}\label{cor:monotoneweightsU}
	Let $\overline{\mathcal U }^\Gamma= \cup_{\pi \in \Pi} \; \Ugamma(\pi)$.
	Then, for any $\pi\in\Pi$,
	\begin{align*}
	{\overline R}_{\pi_0}(\pi;\overline{\mathcal U }^\Gamma)
	=\suma\sup_{ u \in\overline{\mathcal U }^\Gamma }{R}^{(t)}_{\pi_0}(\pi;W(u)),\quad
	\hat{\overline R}_{\pi_0}(\pi;\overline{\mathcal U }^\Gamma_n)
	=\suma\sup_{u \in\overline{\mathcal U }^\Gamma_n  }{\hat R}^{(t)}_{\pi_0}(\pi;W(u)).
	\end{align*}
\end{corollary}
This characterization is a consequence of \Cref{prop:monotoneweights} and its proof, which studies the linear change of variables from $W$ to $u$. 

{ 
	For this section, we consider maximal inequalities for the function classes for the enveloped policy class $\mathcal F$) and policy-optimal weight functions. Let $\epsilon_i \in \{-1,+1\}$, be iid Rademacher variables (symmetric Bernoulli random variables with value $-1,+1$ with probability $\frac 12$), distributed independently of all else. 
}
{
	\begin{lemma}[Uniform convergence of policy function $\pi$ over envelope class $\mathcal F $]\label{lemma-policy-uc}
Let \mbox{$f(\pi)\leq \norm{F}_2 \leq C$} be a bound on the envelope function for $f \in \mathcal F$. 
		Then for $n$ large enough, with probability $> 1 - \delta$,  
\begin{equation} \sup_{ f \in \mathcal F } \abs{ \frac{1}{n}\sum_{i=1}^n (f_i(\pi) - \E[f(\pi)] ) } \leq  \nicefrac{9}{2} C \sqrt{K} ({v+2} ) \sqrt{\frac{  \log(\nicefrac{5}{\delta}) }{{n} } }
\end{equation}
	\end{lemma}
	\proof{Proof.}
		We first bound the deviations uniformly over the policy class and introduce the following empirical processes, $$M = \sup_{ f \in \mathcal F } \abs{ \sum_{i=1}^n (f_i(\pi) - \E[f(\pi)] ) }, L = \sup_{ f \in \mathcal F }  \abs{ \sum_{i=1}^n \epsilon_i f_i(\pi)  }.$$ 
By a standard symmetrization argument, applying Jensen's inequality for the convex function $\Phi$ of the symmetrized process (e.g. Theorem 2.2 of \cite{pollard1990empirical}), we may bound the Orlicz norm of the maxima of the empirical process by the symmetrized process, conditional on the observed data: 
$$ \E[ \Phi(M)] \leq  \E[ \Phi(2L)]   $$
		Taking Orlicz norms with $\Phi(t) = \frac{1}{5} \exp(t^2)$, we apply a tail inequality on the Orlicz norm of the symmetrized process $\Phi \left( 2 L  \right)$, under the assumption of bounded outcomes. Applying Dudley's inequality to the symmetrized empirical process $L$, (e.g. Theorem 3.5 of \cite{pollard1990empirical}), we have that 
		\begin{equation}\label{duddleyinteq} \E_\epsilon\bracks{ \exp(L^2/J^2) \mid \mathcal D }\leq5~~\text{for}~~ J=9\magd{F}_2\int_0^1\sqrt{\log(D(\magd{F}_2\zeta, \mathcal F(X_{1:n}) ))}d\zeta, \end{equation}
Then, applying \Cref{thm-dudley-ball-pajor}, we have that there exists a universal constant $K$ (depending only on the VC-major dimension $v$ of the policy class), such that: 
$$ \log D(\norm{F}_2 \zeta, \mathcal{F}(X_{1:n},T_{1:n})  ) \leq K \left(\frac{1}{\zeta}\right)^{\frac{2v}{v+2}}$$
The corresponding Dudley entropy integral is bounded by $\int_0^1 \sqrt{  K \left(\frac{1}{\zeta}\right)^{2v/(v+2)} } d\zeta \leq \sqrt{K} \frac{v+2}{2}.$
		By Markov's inequality, we have that
		$$ \mathbb{P}\left( \frac{1}{n} L > t \right) \leq 5 \exp(-t^2n^2/ \norm{L}^2_2 ) = 5 \exp( -t^2 n /J^2 C^2  ) $$
		so that therefore,
		$$ \frac{1}{n}M \leq \frac{ \nicefrac{9}{2} C \sqrt{K} ({v+2} )\sqrt{\log(\nicefrac{5}{\delta})} }{\sqrt{n} } $$
		\qed
	\endproof
}
{
	\begin{lemma}[Uniform convergence of weight functions $u(y(\pi(t\mid x) - \pi_0(t\mid x)))$ over $\Ugamma(\pi)$]\label{lemma-weight-uc}
		With probability $\geq 1 - p$, we have that 
		$$\sup_{u \in \Ugamma(\pi) }   \frac{1}{n} \abs{ \sum_{i=1}^{n}  u(Y_i(\pi(T_i\mid X_i)-\pi_0(T_i\mid X_i) ) )  - \E[u(Y(\pi(T\mid X)-\pi_0(T\mid X) ) )]   } \leq 9 \sqrt{ \frac{ {2\pi \log(\nicefrac{1}{p})} }{ {n} } } $$
	\end{lemma}
}
\proof{Proof.}
{
	We define the maxima of the empirical process (and its symmetrization), $H^t,S^t$, for the weight function $u \in \Ugamma(\pi)$, which maximizes over $u$ for a fixed $\pi$: \begin{align*}
	H&= \sup_{ u \in \Ugamma(\pi)}  \abs{ \sum_{i=1}^{n}  u(Y_i(\pi(T_i\mid X_i)-\pi_0(T_i\mid X_i)))  - \E[u(Y(\pi(T\mid X)-\pi_0(T\mid X) ) )] } \\
	S&= \sup_{u \in \Ugamma(\pi) }  \abs{ \sum_{i=1}^{n} \epsilon_i u(Y_i(\pi(T_i\mid X_i)-\pi_0(T_i\mid X_i))) }
	\end{align*}
	Taking Orlicz norms and symmetrizing as in the proof of \Cref{lemma-policy-uc}, we have that $\E[ \Phi(H)] \leq \E[ \Phi(2S)]$.
	We show that the entropy integral (log of the covering numbers) is bounded
using the VC-hull property of %
the class of non-decreasing functions taking values on $[0,1]$ ultimately is VC-hull but not VC-subgraph \cite{van1996weak}. 
	$\Ugamma(\pi)$ is in fact included in the symmetric convex hull of $\mathcal I$, $\Ugamma(\pi) \subseteq \overline{\op{conv}}(\mathcal I)$. (This follows since taking differences of indicator thresholds recovers any interval, e.g. Example 3.6.14 of \cite{gine2016mathematical}). 
}
{
	We apply \Cref{thm-dudley-ball-pajor} (relating the log-covering numbers to the entropy integral for VC-hull classes and their generating VC-subgraph classes), and using a result from Sec. 3 of \cite{van1996new} to extract an explicit bound for this class of functions: 
	$$\log D(
	\zeta, \Ugamma(\pi)(X_{1:n}, T_{1:n}, Y_{1:n})) \leq \frac{1}{\zeta} \log(\frac{1}{\zeta}).$$ 
	The Dudley entropy integral is in turn bounded by $\int_0^1 \sqrt{\frac{1}{\zeta} \log(\frac{1}{\zeta})} d\zeta \leq \sqrt{2\pi}$.
	Next, we apply Theorem 3.5 of \cite{pollard1990empirical}, in order to bound mgf $\E\left[ \exp\left( \frac{1}{81} \frac{1}{2\pi} \frac{S^2}{n } \right) \right] \leq 5$ 
	such that we can use the subgaussian tail bound $ \pr\left[ \frac{1}{n} S \geq t \right] \leq 5 \exp \left(-\frac{  t^2  }{162 \pi} n \right)  $.
	Therefore, with probability $\geq 1 - p$, we have that 
	$$ \frac{1}{n}S \leq {9} \sqrt{ \frac{ 2\pi\log(\nicefrac{1}{p})}{ {n} } } $$
}
\qed
\endproof
An analogous result holds for the restriction of the process to a specific treatment partition $t$.
\begin{corollary}
			With probability $\geq 1 - p$, we have that 
	$$\sup_{u \in \Ugamma(\pi) }   \frac{1}{n} \abs{ \sum_{i=1}^{n}  u(Y_i(\pi(T_i\mid X_i)-\pi_0(T_i\mid X_i) ) ) \mathbb{I}[T_i =t] - \E[u(Y(\pi(T\mid X)-\pi_0(T\mid X) )) \mathbb{I}[T =t] ] } \leq  9\sqrt{ \frac{ {2\pi \log(\nicefrac{1}{p})} }{ {n} } }  $$
\end{corollary}
{
Next, we use the previous results to obtain a uniform convergence result for the minimax weight functions 
when we optimize jointly over policy functions $\pi$ and weight functions $u(y(\pi(t\mid x) - \pi_0(t\mid x))) \in \overline{\mathcal U }^\Gamma$. Under \Cref{prop:monotoneweights}, the weight function class remains monotone, even under composition with a VC-major policy class: however the dimension of the resulting class is not explicit from the stability result.
	\begin{lemma}[Uniform convergence of $u(y(\pi(t\mid x) - \pi_0(t\mid x)))$ over $\overline{\mathcal U }^\Gamma$]\label{lemma-weight-uc-pi}
		With probability $\geq 1 - p$, for some universal constant $K_{ u \circ \pi }$ that only depends on $v_{ u \circ \pi }$ (the VC-major dimension of the composition class) we have that 
		$$\sup_{u \in \overline{\mathcal U }^\Gamma}   \frac{1}{n} \abs{ \sum_{i=1}^{n}  u(Y_i(\pi(T_i\mid X_i)-\pi_0(T_i\mid X_i) ) ) 
		- \E[u(Y(\pi(T\mid X)-\pi_0(T\mid X) ) )]
	} \leq \nicefrac{9}{2}\sqrt{K_{ u \circ \pi }}({v_{ u \circ \pi }+2})\sqrt{    \frac{  \log(\nicefrac{1}{p}) }{ {n} } }$$
	\end{lemma}
	\proof{Proof.}
We first apply \Cref{lemma-vcmajor-stability-monotone-composition} (a stability result for the composition of monotone function classes with VC-major classes) under Assumption~\ref{assumption:vc1}. Then, applying \Cref{thm-dudley-ball-pajor}, there exists a universal constant $K$ (depending only on $v=v_{ u \circ \pi }$ (the VC-major dimension of the composition class in \Cref{prop:monotoneweights}), such that: 
		$$ \log D( 
		 \zeta , \overline{\mathcal U }^\Gamma( X_{1:n}, T_{1:n}, Y_{1:n})  ) \leq K \left(\frac{1}{\zeta}\right)^{\frac{2v}{v+2}}$$
The result follows by the typical chaining argument (e.g. \Cref{lemma-weight-uc}), but instead bounding
		the Dudley entropy integral by $\int_0^1 \sqrt{  K \left(\nicefrac{1}{\zeta}\right)^{\frac{2v}{v+2}} } d\zeta \leq \sqrt{K} \frac{v+2}{2}.$
		\qed
\endproof
 }
{
		\subsection{Proof of \Cref{unifconv}} 
	\proof{Proof of \Cref{unifconv}}
The proof of uniform convergence over $\pi \in \Pi, W \in \overline{\mathcal W}^\Gamma$ follows by decomposing the regret, then applying the tail inequalities of the previous section.
	\paragraph{Regret Decomposition}
The following lemma allows us to study the minimax regret via uniform convergence arguments. 
\begin{lemma}\label{lemma-supdiffsup}
	\begin{equation}
	\sup_{y \in S}h(y) - \sup_{y \in S}g(y)   \leq	\sup_{y \in S} \{ h(y) - g(y)\} %
	\end{equation}
\end{lemma}
\proof{Proof.}
To see this, consider $y_1^* \in \arg\max h(y), y_2^*  \in \arg\max g(y)$ and $y^* \in \arg\max h(y) - g(y)$: then
$$ h(y_1^*) - g(y_2^*) \leq  h(y_1^*) - g(y_1^*) \leq  h(y^*) - g(y^*)$$
\qed
\endproof
	We use \Cref{prop:monotoneweights} and \Cref{lemma-supdiffsup} in the following minimax regret decomposition where $\pi_{CR} = \frac 1m$: 
}
{
	\begin{align*} 
	&
	\sup_{ {  \pi \in \Pi } }
	\left\{ \sup_{ W \in {\mathcal{W}^\Gamma}(\pi) } \hat{R}_{\pi_{0}}(\pi , W) - \sup_{ W \in {\mathcal{W}^\Gamma}(\pi)  }  R_{\pi_{0}}(\pi,  W  ) \right\} 
	\\&
	\sup_{ {  \pi \in \Pi } }
	\left\{ \sup_{ W \in {\overline{\mathcal{W}}^\Gamma}(\pi) } \hat{R}_{\pi_{0}}(\pi , W) - \sup_{ W \in {\overline{\mathcal{W}}^\Gamma}(\pi) }  R_{\pi_{0}}(\pi,  W  ) \right\} 
	\\
	&\leq
	{  \sup_{ { \pi \in \Pi, W \in {\overline{\mathcal{W}}^\Gamma}(\pi)  } }  \hat{R}_{\pi_{0}}(\pi ,W ) -  R_{\pi_{0}}(\pi, W  ) }    \\
	&\leq \sup_{ { \pi \in \Pi, W \in {\overline{\mathcal{W}}^\Gamma}(\pi) } }
	\{  \hat{R}_{\pi_{\CR}}(\pi , W ) -  R_{\pi_{\CR}}(\pi,  W  ) \} + 
	\underbrace{ \sup_{ {W \in {\overline{\mathcal{W}}^\Gamma}(\pi_0)   } }
	\{  \hat{R}_{\pi_{\CR}}(\pi_0 , W) -  R_{\pi_{\CR}}(\pi_0,  W  ) \} }_{ \textcircled{\raisebox{-0.9pt}{3}} }
	\end{align*}
	Then, using subadditivity of the supremum, that $\mathcal{W}^\Gamma$ is a product uncertainty set, and the elementary decomposition ${\frac{a}{b} - \frac{c}{d}} = {a} \frac{{b-d} }{{b} {d}} + \frac{{a-c}}{{d}}$,
	we further decompose the minimax regret:
	\begin{align*}& \sup_{ { \pi \in \Pi, W \in {\overline{\mathcal{W}}^\Gamma}(\pi) } }
	\{  \hat{R}_{\pi_{\CR}}(\pi , W ) -  R_{\pi_{\CR}}(\pi,  W  ) \} \\ 
	& \leq {  \sup_{ \pi \in \Pi, W \in {\overline{\mathcal{W}}^\Gamma}(\pi)  }  \left\{ \suma
		\frac{ \E_n [ (\pi(T\mid X) - \frac{1}{m}) YW \mathbb{I}[T=t] ] )}{E_n [ W \mathbb{I}[T=t] ] } 
		- \frac{ { \E[ (\pi(T\mid X) - \frac{1}{m}) YW \mathbb{I}[T=t] ]}}{\E[ W \mathbb{I}[T=t] ] }  \right\} }\\
	& \leq 
\sup_{ \pi \in \Pi, W \in {\overline{\mathcal{W}}^\Gamma}(\pi)  }     %
 \suma \frac{  (\E_n - \E) ( (\pi(T\mid X) - \frac{1}{m})  YW \mathbb{I}[T=t] ) }{   \sumexpregdenom   } %
 \\
& + \sup_{ \pi \in \Pi, W \in {\overline{\mathcal{W}}^\Gamma}(\pi)  }
\suma\frac{ \E_n [  (\pi(T\mid X) - \frac{1}{m})  YW \mathbb{I}[T=t] ] )}{ \E_n [ W \mathbb{I}[T=t] ]}  \frac{  (\E_n - \E) (W \mathbb{I}[T=t] ) }{   \sumexpregdenom}   
\\
& 
\leq 	
\sup_{ \pi \in \Pi, W \in {\overline{\mathcal{W}}^\Gamma}(\pi)  }     
\underbrace{    { (\E_n - \E) (  (\pi(T\mid X) - \nicefrac{1}{m})  YW  ) }}_{\textcircled{\raisebox{-0.9pt}{1}}
}
+
\abs{ B} \suma  \sup_{W \in {\overline{\mathcal{W}}^\Gamma}(1)} \underbrace{\abs{  (\E_n - \E) (W \mathbb{I}[T=t] ) }}_{\textcircled{\raisebox{-0.9pt}{2}}
 } 
	\end{align*} 
}
{
	The last inequality follows by applying submultiplicativity of the supremum (for absolute values), and since $\E[ W \mathbb{I}[T=t]] = 1$.
The upper bound $\sup_{ \pi \in \Pi , W \in {\overline{\mathcal{W}}^\Gamma(\pi)} }   \abs{ \frac{ \E_n [   (\pi(T\mid X) - \frac{1}{m})  YW \mathbb{I}[T=t] ] )}{ \E_n [ W \mathbb{I}[T=t] ]}  } \leq B$ follows since this term simply evaluates the minimax regret over ${\overline{\mathcal{W}}^\Gamma}(\Pi)$: due to weight normalization, it is \textit{deterministically} bounded by $B$ under Assumption~\ref{asn-bounded-outcomes}. We now apply the tail inequalities of the previous section to the maximal processes of $\textcircled{\raisebox{-0.9pt}{1}}, \textcircled{\raisebox{-0.9pt}{2}}, \textcircled{\raisebox{-0.9pt}{3}}$, in this order.
}
{
	\paragraph{$\textcircled{\raisebox{-0.9pt}{1}}$ Reducing a bound on product function class to the individual function classes. }
		{
			Recall the weight functions are re-parametrized with respect to $u$: throughout this analysis, for brevity, we denote this by $W_{i}(u(\pi))$: 
\begin{equation*}W_{i}(u(\pi)) = a_{T_i}^\Gamma(X_i) +( b_{T_i}^\Gamma(X_i) - a_{T_i}^\Gamma(X_i)) u_{T_i}(Y_i (\pi(T_i \mid X_i) - \frac{1}{m}  )).\end{equation*}
		}
	Now define $(Q, P)$ for the quantities for the empirical process for the product function class and its symmetrized version: 
	\begin{align*}
&Q = \sup_{f \in \mathcal{F} , W\in { \overline{\mathcal{W}}^\Gamma}(\pi)  } \abs{ \sum_{i=1}^n (f_i(\pi) W_i(u(\pi))) - R_{\pi_{CR}} (\pi,W) )  }, \;\;
P = \sup_{f \in \mathcal{F} , W\in { \overline{\mathcal{W}}^\Gamma}(\pi)   }  \abs{ \sum_{i=1}^n \epsilon_i f_i(\pi) W_i(u (\pi) ))   }
\end{align*}
		By a symmetrization argument (Theorem 2.2 of \cite{pollard1990empirical}), we have that 
	$$ \textstyle \E \Phi(Q) \leq \E[ \Phi(2P) ] $$
	We now reformulate the maximal inequality over the product function class in terms of Orlicz norms on each function class $\mathcal F , \overline{\mathcal W }^\Gamma$ separately, using the fact that observation $f W = \frac{1}{4} ( f+W)^2 -\frac{1}{4} ( f-W)^2)$. 
	For the weight function $W_i(u(\pi))$, we will use the contraction map $\lambda(s) = \nicefrac{1}{2 \max_x b(x) - \min_x a(x)} \min(1, s^2)$. 
	We then decompose the terms including the product of $f, W$ to the sums of squares of $f, W$, optimize over $W \in \overline{\mathcal W }^\Gamma$ rather than $W \in {\Wgamma}(\pi)$, and then apply a contraction result in order to use results on convergence over $\pi \in \Pi, u \in  \overline{\mathcal U }^\Gamma$. We next apply inequality 5.5. of \cite{pollard1990empirical}, which decomposes the maximal inequality over the addition of function classes,
 $ \E_\epsilon \Phi( \underset{f \in \mathcal{F}, \overline{\mathcal{W}}^\Gamma(\pi) }{\sup} \abs{ \epsilon \cdot (f + W) } ) \leq \frac{1}{2} \E_\epsilon \Phi\left( 2 \underset{{f \in \mathcal{F} }}{\sup} \abs{ \epsilon \cdot f } \right) +  \frac{1}{2}  \E_\epsilon \Phi\left( 2 \underset{{\overline{\mathcal{W}}^\Gamma(\pi) }}{\sup}  \abs{ \epsilon \cdot W } \right).$ 
	\begin{align*}
	&\E[ \Phi(2P) ] \\
	&\leq   \E_\epsilon \Phi \left( \sup_{ f \in \mathcal{F} , \overline{\mathcal{W}}^\Gamma(\pi) } \frac{1}{2} \abs{ \epsilon \cdot (f(\pi)+ (a^\Gamma + (b^\Gamma-a^\Gamma) u))^2 } \right) +\E_\epsilon \Phi \left( \sup_{ f \in \mathcal{F} , \overline{\mathcal{W}}^\Gamma(\pi) } \frac{1}{2} \abs{ \epsilon \cdot (f(\pi)- (a^\Gamma + (b^\Gamma-a^\Gamma) u))^2 } \right)  \\
	& \leq  {  \E_\epsilon \Phi \left( \sup_{ f \in \mathcal{F} , \overline{\mathcal{W}}^\Gamma(\pi)} \abs{\frac{1}{2} \epsilon \cdot (f(\pi)\pm  (b^\Gamma-a^\Gamma) u )^2} \right)} + {\frac{1}{2}   \E_\epsilon \Phi \left(\sup_{ f \in \mathcal{F} , \overline{\mathcal{W}}^\Gamma(\pi) } 2 \abs{ \epsilon \cdot a^\Gamma (f(\pi)\pm  (b^\Gamma-a^\Gamma) u )} \right)}  \\
	&  \leq{3} \E_\epsilon \Phi \left( 8\sup_{f \in \mathcal{F} }  \abs{ \sum_{i=1}^n \epsilon_i f_i(\pi)  }  \right) 
+ \frac{1}{2} \E_\epsilon \Phi \left( 4 \frac{1}{\nu} \sup_{f \in \mathcal{F} }  \abs{ \sum_{i=1}^n \epsilon_i f_i(\pi)  }\right)  \\
	&	+ {3 } \E_\epsilon \Phi \left( 8 \frac{1}{\nu}   (\Gamma - \frac{1}{\Gamma}) \sup_{ u \in \overline{\mathcal{U}}^\Gamma }  \abs{ \sum_{i=1}^{n} \epsilon_i u(Y_i (\pi(T_i \mid X_i)-\frac{1}{m}) ) } \right)
	+ \frac{1}{2} \E_\epsilon \Phi \left( 4 \frac{1}{\nu^2}   (\Gamma - \frac{1}{\Gamma})  
	\sup_{ u \in  \overline{\mathcal{U}}^\Gamma}  \abs{ \sum_{i=1}^{n} \epsilon_i u(Y_i (\pi(T_i \mid X_i)-\frac{1}{m}) ) }
	 \right)  \label{eq:prod-class-decomposition}
	\end{align*}
	The last inequality follows from a Lipschitz contraction result (see e.g. Theorem 5.7 of \cite{pollard1990empirical}).
	From the above decomposition, it remains to apply the tail inequalities of \cref{lemma-policy-uc,lemma-weight-uc,lemma-weight-uc-pi} and a contraction argument separately for the function classes on $\mathcal F, \overline{\mathcal U }^\Gamma$.
}
{
	\paragraph{Applying tail inequalities on maxima of individual function classes $\Pi, \Ugammand$: }
}
{
	For $n$ large enough,
	with probability greater than $1 - p_1$, where $p_1 = \frac{p}{6}$, 
	\begin{align*} \E[ \Phi(2P)] 
	& \leq 18(12+\nicefrac 1\nu)  (B\sqrt{K_{\pi }} ({v_{ \pi }+2} ) +2 \frac{1}{\nu} (\Gamma  - \frac{1}{\Gamma}) \sqrt{K_{ u \circ \pi }} ({v_{ u \circ \pi }+2})) \sqrt{ \frac{\log(\nicefrac{30}{p})}{ {n} }}
	\end{align*}
}
{
	$\textcircled{\raisebox{-0.9pt}{2}}$ We next bound the maximal deviations of the term $$ \suma \sup_{u \in {\Ugamma}(1)} \abs{ \frac 1n \sum_i \mathbb{I}[T_i = t]W(t, X_i, Y_i) - \E[ \mathbb{I}[T = t]W(t, X,Y) ] }.$$ 
Note that studying uniform convergence of $\textcircled{\raisebox{-0.9pt}{2}}$, $\textcircled{\raisebox{-0.9pt}{3}}$, we can restrict attention to nondecreasing weights which are nondecreasing in a fixed policy, ${\Ugamma}(1)$. We apply the tail inequality of \Cref{lemma-weight-uc} with a contraction argument, and obtain
	a bound on the maxima of the absolute value deviation by an argument of Remark 8.1.5 of \cite{vershynin2018high}:
	note that the zero function is an element of the class of non-decreasing functions on $\mathbb{R}$, and apply Dudley's inequality to the increment process $\abs{ \hat D_t - 0 }$.
	Choosing $p_2 = \frac{p}{3 m }$, and taking a union bound over the event that each bound holds for each treatment partition $t$, we obtain the high probability bound that 
	$$ \suma \sup_{u \in \Ugamma(1)} \abs{ \frac 1n \sum_i \mathbb{I}[T_i = t]W(t, X_i, Y_i) - \E[ \mathbb{I}[T = t]W(t, X,Y) ] } \leq   \frac{  18m\nicefrac 1\nu(\Gamma - \frac{1}{\Gamma}) \sqrt{\log(\nicefrac{15 m }{p})} }{\sqrt{n} }$$
		$\textcircled{\raisebox{-0.9pt}{3}}$ Lastly, we bound
		$ \sup_{u \in \Ugamma(\pi_0) } \abs{ \hat{R}_{\pi_{CR}}(\pi_0 , W(u(\pi_0)) ) -  R_{\pi_{CR}}(\pi_0, W(u(\pi_0)) ) },$ follows from the tail inequality of \Cref{lemma-weight-uc}, such that for $n$ large enough,
		, with probability greater than $1-p_2$, where $p_2 = \frac{p}{3}$,
	$$ \sup_{u \in \Ugamma(\pi_0) } \abs{ \hat{R}_{\pi_{CR}}(\pi_0 , W(u(\pi_0)) ) -  R_{\pi_{CR}}(\pi_0, W(u(\pi_0)) ) } \leq   \frac{ {18} B \nicefrac 1\nu(\Gamma - \frac{1}{\Gamma}) \sqrt{  \log(\nicefrac{15}{p})} }{ \sqrt{n} }$$
	Putting together the above bounds on terms $ \textcircled{\raisebox{-0.9pt}{1}}, \textcircled{\raisebox{-0.9pt}{2}}, \textcircled{\raisebox{-0.9pt}{3}}$
 we have that with probability $\geq 1 - p$,: 
\begin{align*}  &\sup_{\pi\in\Pi}\abs{
	\hat{\overline R}_{\pi_0}(\pi;\mathcal W_n^\Gamma)
	-
	\overline R_{\pi_0}(\pi;\mathcal W^\Gamma)
}
\\
& \leq  18(12+\nu^{-1})  (B\sqrt{K_{\pi }} ({v_{ \pi }+2} ) +\nu^{-1} (\Gamma  -  \Gamma^{-1} ) (2 \sqrt{K_{ u \circ \pi }} ({v_{ u \circ \pi }+2}) + B + m ) ) \sqrt{ \frac{\log(\nicefrac{15m}{p})}{ {n} }} 
\end{align*} 
}
The statement follows by collecting constants that depend on $v$.
\qed
\endproof
\subsection{Proof of \Cref{improvethm}}
\proof{Proof of \Cref{improvethm}}
{
			We analyze uniform convergence for the true propensity weights, assumed to be in the uncertainty set, $W^* \in \mathcal{U}$. We use the tail inequalities of \cref{lemma-policy-uc}, as well as standard Hoeffding inequalities for the sample expectations, with the \textit{true} inverse propensity weights $W^*_t(X_i,Y_i)$. Define  
	$$ \hat D_t^* = \E_n[(\piAX - \nicefrac{1}{m})W^* \indic{T=t}]. $$ 
	First consider an analogous regret decomposition as in the proof of \Cref{unifconv}: 
\begin{align*}	
	\sup_{ { \pi \in \Pi } }&
\{ \hat{R}_{\pi_{0}}(\pi , W^* ) -  R_{\pi_{0}}(\pi,  W^*  ) \}   \\
\leq \sup_{ { \pi \in \Pi } }& 
	\{  \hat{R}_{\pi_{\CR}}(\pi , W^* ) -  R_{\pi_{\CR}}(\pi,  W^*  ) \} + 
\left(  \hat{R}_{\pi_{\CR}}(\pi_0 , W^*) -  R_{\pi_{\CR}}(\pi_0,  W^*  ) \right) 
\end{align*}
Note that the second term can be bounded by application of Hoeffding's inequality, such that with probability $\geq 1 -p_3$,
$$\abs{  \hat{R}_{\pi_{CR}}(\pi_0 , W^* ) -  R_{\pi_{CR}}(\pi_0, W^*) } \leq  \nicefrac{B}{\nu}  \sqrt{ \frac{ { \log(\nicefrac{2}{p_3}) } }{2{n}} } $$
Next, we bound the regret deviation uniformly over $\pi$:
\begin{align*}	
\sup_{ { \pi \in \Pi } }&
\{ \hat{R}_{\pi_{\CR}}(\pi , W^* ) -  R_{\pi_{\CR}}(\pi,  W^*  ) \}   \\
&\leq 
 \sup_{ \pi \in \Pi } \frac{1}{n}   \sum_i \frac{  (\pi(T_i\mid X_i) - \frac{1}{m} ) W_i^* Y_i }{   \E[ \hat D_{T_i}] }  -  R_{\pi_{0}}(\pi,  W^*  ) 
 + \frac{1}{n} 
\sum_i  \frac{  (\pi(T_i\mid X_i) - \frac{1}{m} ) W_i^* Y_i }{    
 \E[ \hat D_{T_i} ]
}   \frac{ \E[ \hat D_{T_i}] - \hat D_{T_i}  }{\hat D_{T_i}  } \\
& \leq
 \sup_{ \pi \in \Pi }\left\{ \frac{1}{n}   \sum_i  (\pi(T_i\mid X_i) - \frac{1}{m} ) W_i^* Y_i -  R_{\pi_{0}}(\pi,  W^*  ) \right\} 
 + \frac{B}{\nu}  \sum_i  \frac 1n\frac{ \E[ \hat D_{T_i}] - \hat D_{T_i}  }{    \hat D_{T_i} } 
\end{align*}
We apply \Cref{lemma-policy-uc} (e.g. a standard chaining argument with bounded envelope function $WY \leq \nicefrac{B}{\nu}$) to bound the first term. 
Therefore, we have that with high probability greater than $p_2$, the first term is bounded by: $$	 \sup_{ \pi \in \Pi }\left\{ \frac{1}{n}   \sum_i  (\pi(T_i\mid X_i) - \frac{1}{m} ) W_i^* Y_i -  R_{\pi_{0}}(\pi,  W^*  ) \right\}   \leq  {  \frac{9B}{2\nu} 
\sqrt\frac{ {\log(\nicefrac{5}{p_2})} }{{n} } }. $$
We then bound the second term, $\frac{B}{\nu}  \sum_i  \frac 1n\frac{ \E[ \hat D_{T_i}] - \hat D_{T_i}  }{   \hat D_{T_i} } $:  instead of summing the second term over treatments $t$, observe that for $n_t = \sum_i \mathbb{I}[T_i = t]$,
\begin{align*}
\nicefrac{B}{\nu}  \sum_i  \frac 1n\frac{ \E[ \hat D_{T_i}] - \hat D_{T_i}  }{   \hat D_{T_i} } = \nicefrac{B}{\nu}  \frac 1n \suma  n_t \frac{ \abs{\hat D_{t}   -1 } }{   \hat D_{t} } 
\end{align*}
}
	{ 
We proceed conditionally on the event that $\frac{n_t}{n} \in [ \frac{1}{2}\rho_t, \frac{3}{2}\rho_t ], \;\; \forall t \in \{ 0, \dots, m-1 \}$, where $\rho_t = \pr(T=t)$ is the marginal probability of treatment. By Hoeffding's inequality, $\pr( \abs{ \frac{n_t}{n}  -\rho_t} \geq \rho_t/2 )  \leq 2 \exp ( - \frac{1}{2} \nu^2 \rho_t^2 n )$, so it suffices to choose $p_4 \in [0,1]$ such that $ \frac{1}{\nu} \sqrt{ \frac{\log(\nicefrac{2m}{p_4})}{2 n} } \leq \rho_t^2/2, \forall t \in \{0,\dots,m-1 \} $ (after taking a union bound over the $m$ treatment groups). Next, we bound $\frac{ \abs{\hat D_{t}   -1 } }{   \hat D_{t} }$: by Hoeffding's inequality,%
$$ \pr( \abs{ \hat D_{t}-1} \geq \epsilon )  \leq 2 \exp ( -2 \nu^2 \epsilon^2 n ) $$ 
For $p_1 \in [0,1]$ such that $ \frac{1}{\nu} \sqrt{ \frac{\log(\nicefrac{2m}{p_1})}{2 n} } \leq 1  $ then with probability at least $1-p_1$, $\frac{1}{\hat D_{t}} \leq 2$ and $\frac{ \abs{(1-\hat D_{t}) 	}  }{\hat D_{t}} \leq \frac{2}{\nu} \sqrt{ \frac{\log(\nicefrac{2}{mp_1})}{2 n} }, \forall t \in \{0, \dots, m -1 \}$ (again taking a union bound over $t \in \{0, \dots, m-1\}$). 
}
{
Now combining the above tail inequalities and applying the union bound, we have that for $p_1, p_2, p_3, p_4 = \frac \delta4$ for $p > 0$, with high probability greater than $1-p$, 
\begin{align*} \sup_{ { \pi \in \Pi } }  \{ \hat{R}_{\pi_{0}}(\pi , W^* ) -  R_{\pi_{0}}(\pi,  W^*  ) \} & \leq  \frac{B}{\nu} \sqrt{ \frac{\log(\nicefrac{8}{p_3})}{2{n}} }   + 36 \frac{B\sqrt v }{\nu}  \frac{ \sqrt{\log(\nicefrac{20}{p})} }{\sqrt{n} } +   \frac{3}{\nu} \sqrt{ \frac{\log(\nicefrac{8m}{p_1})}{2 n} }\\
& \leq  \frac{1}{\nu}( {B}(1 + \frac{9}{2}K(v+2)) + 3  ) \sqrt{\frac{ 2\log(\nicefrac{\max(8m,20)}{\delta}) }{{n}}}
\end{align*}
Lastly, the proof follows by noting that by assumption of well-specification, $W^*_t \in \mathcal{W}_t$, so there exists $\ccpscalar_t > 0, \forall t \in \mathcal T $ such that $\frac{ {W^*_t}}{\ccpscalar_t}\in \mathcal{W}_t$, and we have that therefore $\hat{R}_{\pi_0} (\pi, W^*) \leq  \sup_{W^* \in \mathcal{W} }  \hat{R}_{\pi_0} (\pi, W^*)$. 
And, in the statement, we have further folded all $v$-dependent constants into one.
	}
\endproof
\subsection{Proof of \Cref{prop:monotoneweightsLambda}}
\proof{Proof of \Cref{prop:monotoneweightsLambda}}
\label{proof-prop:monotoneweightsLambda} 
\blockedit{
		We prove that the budgeted uncertainty set solution has bounded entropy integral by first taking a partial Lagrangian dual with respect to the budget constraint, then invoking strong duality to study a partial maximization: we show
		the solution can be reparametrized to instead range over the space of nondecreasing functions on $[0,1]$, $\mathcal{U}$, for the fixed \textit{optimal} $\eta^*, \ccpscalar^*$. We then proceed to argue that the structural result implies, using the equivalence of the linearized fractional program and the fractional program, that we may then correspondingly optimize over the values of $\eta, \ccpscalar$, and the space of nondecreasing functions. This implies that it is sufficient to restrict the optimization to the set of nondecreasing functions (which satisfy the budget constraint), and we may optimize over a set of restricted complexity. This, for example, allows us to leverage the same stability results as in the proof of \Cref{unifconv} to obtain the same regret guarantees.

We first analyze the linearized budgeted linear program in \Cref{pbm-dual-budgeted} (that is, post-Charnes-Cooper transformation) for $\hat{\overline{Q}}(r;\mathcal{W}_n^{\Gamma,\Lambda})$. Throughout, we presume that $\Gamma$ is some fixed input and write $a,b$ for $a^\Gamma, b^\Gamma$. We also analyze the problem within a single treatment component, and reindex $i = 1, \dots, n$ to be counting conditional on a treatment component.
		\begin{equation*}\begin{aligned}
	\hat{\overline{Q}}(r;\mathcal{W}_n^{\Gamma,\Lambda})=
		{\max}_{\ccpscalar \geq 0,w\geq0,d}~&~  \sum_{i=1}^n w_i r_i
		\\\text{s.t.}~&~ \sum_{i=1}^n d_i \leq \Lambda \ccpscalar,~~\sum_{i=1}^n w_i = 1
		\\&~{a_i \ccpscalar} \leq w_i \leq {b_i }\ccpscalar~~\forall\;i=1,\dots,n
		\\&~d_i \geq w_i - \tilde W_i \ccpscalar~~\forall\;i=1,\dots,n
		\\&~d_i \geq \tilde W_i \ccpscalar - w_i~~\forall\;i=1,\dots,n
		\end{aligned}\end{equation*}
In the following, we condense the linearized representation for the absolute value variable $d_i$ and write $d_i = \abs{ w_i - \tilde{W_i}  \ccpscalar }$ for brevity.
First, we take the Lagrangian partial dual, dualizing the normalized budget constraint $\sum_i d_i \leq \Lambda \ccpscalar$ with Lagrange multiplier $\eta$: 
		\begin{align*}
	\hat{\overline{Q}}(r;\mathcal{W}_n^{\Gamma,\Lambda})=	\min_{ \eta \geq 0 }  \max_{w,d, \ccpscalar} \{ \sum_i w_i r_i + \eta(\Lambda \ccpscalar - \sum_i d_i)  \colon a \ccpscalar \leq w \leq b \ccpscalar, \sum_i w_i = 1,  d_i = \abs{ w_i - \tilde{W_i} \ccpscalar}  \} 
		\end{align*}
		We consider a partial maximization, and substitute with the transformation $u_i = \frac{w_i-a_i \ccpscalar}{\ccpscalar(b_i-a_i)}, u\in[0,1]$.  
		Define 
		\begin{align*} {m}(u,\ccpscalar,\eta) &\defeq
 \sum_i r_i (\ccpscalar (b_i - a_i) {u_i}  + \ccpscalar a_i ) + \eta^*(\Lambda \ccpscalar- \sum_i d_i)  \\
 \mathcal{S}(\ccpscalar) &\defeq \left\{  \sum_i \ccpscalar (b_i - a_i)  u_i + \ccpscalar a_i= 1,  d_i = \abs{ w_i(u) - \tilde{W_i} \ccpscalar}, 0 \leq u_i \leq 1, i = 1, \dots, n  \right\}
\end{align*}
		so that 
		$$ \hat{\overline{Q}}(r;\mathcal{W}_n^{\Gamma,\Lambda})= 	\min_{ \eta \geq 0 }  \max_{t > 0, u \in  \mathcal{S}(\ccpscalar)} {m}(u,\ccpscalar,\eta) $$
		By a standard min-max theorem, we may interchange the min and maximum, and by strong duality (with the Slater point of $u$ such that $w_i(u) = \tilde{W}\ccpscalar$), there exists a saddle point pair $(u^*, \ccpscalar^*), \eta^*$ that are best-responses to each other such that $ \hat{\overline{Q}}(r;\mathcal{W}_n^{\Gamma,\Lambda}) = {m}(u^*, v^*,\eta^*).$ 
		We argue further that $\hat{\overline{Q}}(r;\mathcal{W}_n^{\Gamma,\Lambda}) = \max_{u \in S (\ccpscalar^*)} {m}(u, \ccpscalar^*,\eta^*)$; e.g. we may fix a partial best response of $(u,\ccpscalar^*)$ and $\eta^*$, and recover the optimal solution when we optimize over $u$. (We show this by contradiction: Suppose not: that $\tilde{u}^*  \in \arg\max_u {m}(u, \ccpscalar^*,\eta^*)$ is such that ${m}(\tilde{u}^*, \ccpscalar^*,\eta^*) > {m}(u^*, \ccpscalar^*,\eta^*).$ This contradicts strong duality. On the contrary, ${m}(\tilde{u}^* , \ccpscalar^*,\eta^*) < {m}(u^*, \ccpscalar^*,\eta^*)$ is not possible since $u^*$ is feasible for $\ccpscalar^*, \eta^*$ and therefore achieves a better objective value; so this contradicts definition of $\tilde{u}^*  \in \arg\max_u {m}(u, \ccpscalar^*,\eta^*)$.)
		Therefore, by the preceding argument, 
				$$ \hat{\overline{Q}}(r;\mathcal{W}_n^{\Gamma,\Lambda})=  \max_{{u \in \mathcal{S}(\ccpscalar^*)}}{m}(u,\ccpscalar^*,\eta^*) $$
We further simplify and drop terms from the \textit{parametric objective} ${m}(u,\ccpscalar^*,\eta^*)$ that are constant given $\eta^*, \ccpscalar^*$ and therefore do not vary with $u$, 
		such that we recover the \textit{globally optimal} $u^*$ by optimizing the reformulated objective $m'(u,\ccpscalar^*,\eta^*)$:
\begin{align*} 
 {m}'(u,\ccpscalar^*,\eta^*) &\defeq
\sum_i r_i (\ccpscalar^* (b_i - a_i) {u_i}  ) - \eta^*\sum_i d_i)  \\
u^* &\in  \argmax_{u \in  \mathcal{S}(\ccpscalar^*)} \; {m'}(u,\ccpscalar^*,\eta^*)
\end{align*} 
		We next prove that we can \textit{further} reparametrize optimization of the objective function $\overline{m}'(u,\ccpscalar^*,\eta^*)$ over $u \in \mathcal{S}(\ccpscalar^*)$ to the class of $u$ vectors that is nondecreasing in $r$, $$\mathcal{U} = \{ u: \mathbb R \mapsto [0,1] , u \text{ monotonically nondecreasing} \}.$$ 

We prove the following technical result, which establishes a structural result on the globally optimal $u^*(\ccpscalar, \eta)$ which establishes that is of bounded complexity. 
		\begin{lemma}[Nondecreasing parametrization of optimal $u$ for budgeted uncertainty set]\label{lemma-budgeted-nondecreasing}
Fix $\ccpscalar, \eta \geq 0$: then the correspondingly optimal rescaled weight function $  u^*(\ccpscalar, \eta) $, defined as the solution to the optimization problem,
$$ u^*(\ccpscalar, \eta) \in \arg\max \{\ccpscalar\sum_i (b_i - a_i) \left(r_i   - \eta \op{sgn}(w_i(u) > \tilde{W}\ccpscalar )\right)  u_i  \colon  0 \leq u \leq 1, \sum_i \ccpscalar(b_i - a_i)  u_i +  \sum_i a_i= 1 \},
$$
is non-decreasing in the coefficient index vector $r$. Therefore, $  u^*(\ccpscalar, \eta) $ is nondecreasing in $r$ for all $\ccpscalar, \eta$.
		\end{lemma}
	\proof{Proof of Lemma \ref{lemma-budgeted-nondecreasing}}
By the preceding arguments, we have established the optimal subproblem solution can be written as the following program:
			\begin{align*}
\hat{\overline{Q}}(r;\mathcal{W}_n^{\Gamma,\Lambda}) = \max_{u \in\mathcal{S}(\ccpscalar) } \max_{\ccpscalar >0 }  \min_{\eta \geq 0}&
	\sum_i \ccpscalar(b_i - a_i) \left(r_i   - \eta \op{sgn}(w_i(u) > \tilde{W}\ccpscalar )\right)  u_i  \\
	& 0 \leq u \leq 1\\
	&\sum_i \ccpscalar(b_i - a_i)  u_i = 1- \sum_i a_i
	\end{align*}
		The idea is that given the optimal dual variable $\eta^*$ and scaling factor $\ccpscalar^*$, the problem reduces to a similar problem as the fractional knapsack problem: it is sufficient to sort first on the multipliers $r_i$; then fill the knapsack lexicographically in order of distance $\abs{w_i(u) - \tilde{W} \ccpscalar^*}$ (since the $\eta^*$ penalty is fixed and identical for all units).
		We will prove the reparametrization over $\mathcal{S}(\ccpscalar^*) \cap \mathcal{U}$ by contradiction. 
				Suppose not: that the optimal solution, $u^*$ had indices $i, i'$ such that $r_i > r_{i'}$ but $u_i < u_{i'}$. We  enumerate the following cases that exhaust the possible orderings of $u_i, u_{i'}$ relative to $\tilde{W}_{i} \ccpscalar^*, \tilde{W}_{i'} \ccpscalar^*$: 
		\begin{itemize}
			\item $w_i(u_i) <\tilde{W}_i \ccpscalar^*,  w_{i'}(u_{i'}) < \tilde{W}_{i'} \ccpscalar^*$ or $w_i(u_i) >\tilde{W}_i \ccpscalar^*,  w_{i'}(u_{i'}) > \tilde{W}_{i'}\ccpscalar^*$:
			For any same-ordered set we could generate a contradiction by increasing $u_i$ without generating a change in sign that changes the $\eta^*$ coefficient. 
			\item $w_i(u_i) >\tilde{W}_i \ccpscalar^*,  w_{i'}(u_{i'}) < \tilde{W}_{i'} \ccpscalar^*$: increasing $u_i$ cannot change sign of $\eta^*$.
			\item $w_i(u_i) =\tilde{W}_i \ccpscalar^*$: We need only consider a simultaneous perturbation increasing $u_i$ and moving $u_{i'}$ such that $d_{i'}(u_{i'})$ is decreasing; either such a perturbation increases $u_{i'}$ and overall increases the objective, or decreases $u_{i'}$ (which is offset by the increase due to $r_i > r_i'$, and offsets the increase in $d_i(u_i)$. 
		\end{itemize} 
\qed
	\endproof

Note that the characterization of Lemma \ref{lemma-budgeted-nondecreasing}, which states that the optimal $\ccpscalar,\eta$-parametrized solution $u^*(\ccpscalar,\eta)$ is nondecreasing in $r$, in fact characterizes the structure of the optimal \textit{set} of $u(\ccpscalar,\eta)$ \textit{for all} $\ccpscalar, \eta$ since the index for monotonicity, $r$, is independent of the parameters $\ccpscalar, \eta$. Of course, the particular optimal solution $u^*(\ccpscalar,\eta)$ may change in $\ccpscalar,\eta$. 
As a consequence, $$ \hat{\overline{Q}}(r;\mathcal{W}_n^{\Gamma,\Lambda})=  \max_{u \in \mathcal{S}(\ccpscalar^*) \cap \mathcal{U}}{m}(u,\ccpscalar^*,\eta^*) $$
Combining this structural result with the preceding arguments, we establish that we can equivalently search over scalars $\ccpscalar, \eta > 0$, and $u \in \mathcal{S}(\ccpscalar) \cap \mathcal{U}$.
		\begin{align*}
		\hat{\overline{Q}}(r;\mathcal{W}_n^{\Gamma,\Lambda}) = \max_{u \in\mathcal{S}(\ccpscalar) \cap \mathcal{U}} \max_{\ccpscalar >0 }  \min_{\eta \geq 0}&
		\sum_i \ccpscalar(b_i - a_i) \left(r_i   - \eta \op{sgn}(w_i(u) > \tilde{W}\ccpscalar )\right)  u_i  \\
		& 0 \leq u \leq 1\\
		&\sum_i \ccpscalar(b_i - a_i)  u_i = 1- \sum_i a_i
		\end{align*}
We note that by the equivalence of the linear-fractional programs and linearized program, e.g. via the primal variables $W, U = \frac{W-a}{(b-a)}$ on the one hand and the scalarized $w = W \ccpscalar, \ccpscalar = \sum_i W, u = \frac{w-a \ccpscalar}{\ccpscalar(b-a)}$ on the other hand, (and the implied transformations on $d$), our structural result that it is equivalent to optimize over $u(\ccpscalar,\eta)$ nondecreasing  implies that $U^*(\ccpscalar,\eta) = \frac{u^*(\ccpscalar,\eta) }{\ccpscalar}, U \in [0,1]$  is \textit{also} a monotonically nondecreasing function in $r$. (Multiplying by the scalar $\ccpscalar>0$ simply induces an isomorphism to the \textit{same} set of monotonically nondecreasing functions in $r$). Using this final transformation, we show that our structural result holds also for the original primal problem.
\begin{align*}
\hat{\overline{Q}}(r;\mathcal{W}_n^{\Gamma,\Lambda}) = \underset{U \in \mathcal{U}}{\max} \left\{  \frac{\sum_i U_i (b_i -a_i) r_i+ a_i r_i}{\sum_i U_i (b_i -a_i) + a_i } \colon  \sum d_i(U_i) \leq \Lambda \right\} 
\end{align*}
To contextualize this characterization, we remark that this is weaker than \Cref{thm-norm-wghts-soln} as this does not provide us with an algorithmic solution: nonetheless, proving this result that it is sufficient to optimize over $\mathcal{U}$, even in the primal nonconvex fractional formulation, is sufficient to establish uniform convergence. 
		Finally, we specialize the analysis to the setting for our estimator, where $r_i = \pi( T_i \mid X_i) - \pi_0(T_i \mid X_i)Y_i$, which introduces a dependence on $\pi(X_i)$. (Note that the dependence is only on $X$ through the function $\pi$, which is of restricted complexity.) Since we only required the VC-major property of $u(r)$, applying \Cref{lemma-vcmajor-stability-monotone-composition} is sufficient to verify that the VC-major property holds when we also range the policy $\pi\in \Pi$.
}
\endproof
\subsection{Proof of \Cref{cor-budgeted-minimax}}
\proof{Proof of \Cref{cor-budgeted-minimax}}
The proof is similar to that of \Cref{lemma-lipschitzness}: we study sensitivity analysis in the dual of the linearized linear program, in order to isolate an additive approximation error term of the sample budget constraint from its population counterpart; we control the latter uniformly over the space of weights by our previous tail inequality. Since we optimize in the sample based on an \textit{sample expectation estimate} of the L1 budget constraint, we recall the definitions of ${\mathcal W}^{\Gamma,\Lambda}_n(\mathbb P_n)$ and ${\mathcal W}^{\Gamma,\Lambda}_n(\mathbb P)$:
\begin{align*} 
\mathcal{\weightuncertainty}_n^{\Gamma,\Lambda}(\mathbb P_n)& 
=\left\{ W \in \mathbb{R}^n_+ \colon  ~\text{s.t.}~	\frac{1}{\vert \mathcal{I}_t \vert }\sum_{i \in \mathcal{I}_t }\fabs{ W_i - \tilde W_i} \leq {\Lambda_t},~ a_i^\Gamma  \leq W_i \leq   b_i^\Gamma~\forall i  \right\}\\
\mathcal{\weightuncertainty}_n^{\Gamma,\Lambda}(\mathbb P)
& =\left\{ W \in \mathbb{R}^n_+ \colon  ~\text{s.t.}~		\E[ \fabs{ W(T,X,Y) - \tilde W(T,X)}\mid T=t ] \leq {\Lambda_t},~ a_i^\Gamma  \leq W_i \leq   b_i^\Gamma~\forall i  \right\}
\end{align*} 

Now, we use \Cref{prop:monotoneweightsLambda} to equivalently parametrize the optimization over the set of weight functions which include the \textit{nondecreasing} component $u( y(\pi(t\mid x) - \pi_0 (t\mid x )))$, and introduce the corresponding \textit{nondecreasing sample-budgeted} uncertainty set, $ \overline{ \mathcal W}^{\Gamma,\Lambda}_n(\mathbb P_n)$:
 \begin{align*}
\overline{\mathcal W}^{\Gamma,\Lambda}(\pi;\mathbb P_n)
&=\braces{W(t,x,y)\colon 
	\begin{array}{c}
	W(t,x,y)=a_t^\Gamma(x)+u(y(\pi(t\mid x)-\pi_0(t\mid x))) \cdot (b_t^\Gamma(x)-a_t^\Gamma(x)),\\
	\text{$u(y(\pi(t\mid x) - \pi_0(t\mid x ) )):\Rl\to[0,1]$ is monotonic nondecreasing}, \\
	\frac{1}{\vert \mathcal{I}_t \vert }\sum_{i \in \mathcal{I}_t }\fabs{ W(T_i, X_i, Y_i) - \tilde W(T_i, X_i, Y_i)} \leq {\Lambda_t}
	\end{array}}
\end{align*}
In analogy to \Cref{cor:monotoneweights}, we may define the union over the policy class $\overline{\mathcal W}^{\Gamma,\Lambda}(\mathbb P_n) =  \cup_{\pi \in \Pi} \overline{\mathcal W}^{\Gamma,\Lambda}(\pi;\mathbb P_n)$. The next corollary, a consequence of the nondecreasing optimal solution characterization of \Cref{prop:monotoneweightsLambda} states that we recover the optimal regret by optimizing over the restricted class of budgeted weights. 
\begin{corollary}
$$\hat{\overline R}_{\pi_0}(\pi;\mathcal W^{\Gamma,\Lambda}(\mathbb P_n)) = \suma \sup_{W\in \overline{\mathcal W}^{\Gamma,\Lambda}(\mathbb P_n) } \hat{R}_{\pi_0}^{(t)} (\pi; W)$$
\end{corollary}

We show that $\hat{\overline R}_{\pi_0}(
\pi
;\mathcal W^{\Gamma,\Lambda}(\mathbb P_n))$ 
and
$
\hat{\overline R}_{\pi_0}( \pi
;\mathcal W^{\Gamma,\Lambda}_n(\mathbb P))$ are close for the two policies of interest in the minimax regret bound: the sample-optimal 
$\hat{\overline{\pi}} \defeq \hat{\overline{\pi}}(\Pi,{\mathcal{W}}^{\Gamma,\Lambda}_n(\mathbb P_n),\pi_0)$ and population-optimal \mbox{$ \pi^* \in \arg\inf_{\pi\in\Pi}\overline R_{\pi_0}(\pi;\mathcal W^{\Gamma,\Lambda}(\mathbb P))$} policies.
The result will follow by applying this bound with the triangle inequality.

In the following, we denote $r_i = Y_i (\pi(T_i\mid X_i) - \pi_0 (T_i \mid X_i))$ for brevity, and apply the Charnes cooper transformation. Define $\tilde{u} = u \ccpscalar$ as the corresponding transformation for $u$ in the change of variables $W = a + (b-a) u$, and note that this preserves monotonicity of $\tilde{u}$ for all $\ccpscalar$. Denote the uncertainty set on $w, \ccpscalar$ and implicitly, nondecreasing $u$ as $ \overline{\mathcal S} (w,\ccpscalar, \tilde{u}; \pi)$:  
\begin{align*}
\overline{\mathcal S} (w,\ccpscalar, \tilde{u}; \pi )&= 
\left\{\begin{matrix}
{ \sum_i w_i = 1 ;\;\; \ccpscalar_{T_i} a_i^\Gamma \leq w \leq b_i^\Gamma \ccpscalar_{T_i},  \forall i = 1, \dots,n }\\
w = a_i^\Gamma \ccpscalar_{T_i }+ (b_i^\Gamma - a_i^\Gamma) \tilde{u}_i , \forall i = 1, \dots,n \\
\tilde u(y (\pi(t\mid x) - \pi_0(t\mid x))) \text{ monotonically nondecreasing } \\
\ccpscalar_t \geq 0, \forall t
\end{matrix} \right\} \\
 \overline{\mathcal S} (w,\ccpscalar, \tilde{u} )& = \cup_{\pi \in \Pi} \overline{\mathcal S} (w,\ccpscalar, \tilde{u}; \pi )
\end{align*} 
\begin{align*} 
&\hat{\overline{R}}_{\pi_0}(
\pi
;\mathcal W^{\Gamma,\Lambda}(\mathbb P_n))\\
& = \max \left\{ \suma  \frac{\sum_i r_i W_i \mathbb{I}[T_i = t] }{\sum_i W_i \mathbb{I}[T_i = t]} \colon w, \ccpscalar, \tilde u \in \overline{ \mathcal S} \right\} \\
& = \max   \left\{ \suma  {\sum_i r_i w_i \mathbb{I}[T_i = t] }  
\colon 
(w, \ccpscalar, \tilde u) \in \overline{\mathcal S} (w,\ccpscalar, \tilde{u} ) , \sum_{i \in \mathcal{I}_t }\abs{ {w_i}- \ccpscalar_{T_i} \tilde W_i} \leq \ccpscalar_{T_i} {\Lambda_t}, \forall t 
\right\} 
\\
&= \min_{\eta_t \geq 0, \forall t} 
\underset{ 
(w, \ccpscalar, \tilde u) \in \overline{\mathcal S} (w,\ccpscalar, \tilde{u} ) \;
}{\max} 
\left\{ \suma  {\sum_i r_i w_i \mathbb{I}[T_i = t] } + \suma \eta_t (\ccpscalar_{t} {\Lambda_t} -\sum_{i \in \mathcal{I}_t }\abs{ {w_i}- \ccpscalar_{t} \tilde W_i } ) 
\right\}\\
&= \underset{ 
w, \ccpscalar, \tilde u \in \overline{\mathcal S} (w,\ccpscalar, \tilde{u} ) \;
}{\max}  \left\{ \suma  {\sum_i r_i w_i \mathbb{I}[T_i = t] } + \suma \eta_t^*(\mathbb P_n) (\ccpscalar_{t} {\Lambda_t} -\sum_{i \in \mathcal{I}_t }\abs{ {w_i}- \ccpscalar_{t} \tilde W_i } ) 
\right\}
\end{align*} 
for optimal dual variables $\eta^*_t(\mathbb P_n)$, by strong LP duality (existence of the saddle point).
Similarly, for the corresponding optimal dual variable  $\eta^*_t(\mathbb P)$ for the population budget-constrained uncertainty set,
\begin{align*} 
&\hat{\overline R}_{\pi_0}(
\pi
;\mathcal W^{\Gamma,\Lambda}(\mathbb P_n))\\
&=\underset{ 
	w, \ccpscalar, \tilde u \in \overline{\mathcal S} (w,\ccpscalar, \tilde{u} ) \;
}{\max}  \left\{ \suma  {\sum_i r_i w_i \mathbb{I}[T_i = t] } + \suma \eta^*_t(\mathbb P) (\ccpscalar_{t} {\Lambda_t} -	\E[ \fabs{ w(T,X,Y) - \tilde W(T,X)\ccpscalar_{t} }\mid T=t ] \leq {\Lambda_t} \ccpscalar_{t} ) 
\right\}
\end{align*} 
By \Cref{lemma-supdiffsup}, we combine objectives and obtain a lower bound since we optimize over the same feasible set: 
\begin{align*}
& \overline R_{\pi_0}(\pi;\mathcal W^{\Gamma,\Lambda}_n(\mathbb P_n))- \overline R_{\pi_0}(\pi;\mathcal W^{\Gamma,\Lambda}_n(\mathbb P)) \\
& \leq  \underset{ w, \ccpscalar \in \overline{\mathcal S}}{\max}
\ccpscalar_t \suma  {\Lambda_t} ( \eta_t^*(\mathbb{P}_n) - \eta_t^*( \mathbb P) )  + \max(\eta^*_t(\mathbb P), \eta^*_t(\mathbb P_n) ) \suma  (\sum_{i \in \mathcal{I}_t }\abs{ {w_i}- \ccpscalar_{t} \tilde W_i }   - 	\E[ \fabs{ w(T,X,Y) - \tilde W(T,X)\ccpscalar_{t} }\mid T=t ] )\\
& 
\leq \max_{w \in \overline{\mathcal S}(\ccpscalar^*)} \ccpscalar_t^* \suma  {\Lambda_t} ( \eta_t^*(\mathbb{P}_n) - \eta_t^*( \mathbb P) )  + \max(\eta^*_t(\mathbb P), \eta^*_t(\mathbb P_n) ) \suma  (\sum_{i \in \mathcal{I}_t }\abs{ {w_i}- \ccpscalar_{t}^* \tilde W_i }   - 	\E[ \fabs{ w(T,X,Y) - \tilde W(T,X)\ccpscalar_{t}^* }\mid T=t ] )
\end{align*}
Note that $\ccpscalar_t \in \frac{1}{n_t} [ \nu, 1 ] $ by definition. Next, we argue that the optimal dual variables are bounded
by first noting that the optimal primal solution is finite and bounded on $[B, -B]$ by the self-normalized property of the estimator and Assumption~\ref{asn-bounded-outcomes}. Moreover, the constraints on $W$, for a fixed $\Gamma$, imply bounds on how far \textit{feasible} $W$ can be from their nominal values. So, we have a bound which the optimal dual variables must satisfy. Let $$\overline{\Lambda}_t = \max( \frac{1}{\vert \indt \vert}
\sum_{i \in \indt } \max( \tilde W_i -a^\Gamma_i, b_i^\Gamma-\tilde W_i ), 
\E[ \max( \hat W -a^\Gamma, b^\Gamma-\hat W )]) 
$$ denote the maximal total deviation of weights, induced by the uncertainty set on $W^\Gamma$. Let $$\underline{w}_i = \ccpscalar_{T_i} (b_i^\Gamma \mathbb{I}[r_i < 0] + a_i^\Gamma \mathbb{I}[r_i > 0]), \overline{w}_i = \ccpscalar_{T_i} (b_i^\Gamma \mathbb{I}[r_i > 0] + a_i^\Gamma \mathbb{I}[r_i < 0])$$ achieve the minimal and maximal feasible rescaled primal objectives, respectively. Now, we have the bounds that 
$
 \suma \eta_t^* + \sum_i r_i \underline{w}_i   \geq -B$ and $
  \suma \eta_t^* \overline{\Lambda}_t + \sum_i r_i \overline{w}_i   \leq B
$
 which admits a naive componentwise bound that 
 $ \eta_t^* \geq -B -  \sum_i r_i \underline{w}_i,    \eta_t^*   \leq \frac{B -  \sum_i r_i \overline{w}_i }{ \min_t \overline{\Lambda}_t}, \forall t $.

 Therefore, since $\eta_t^* \geq 0$, 
 we obtain the following bound: $$\eta_t \leq  \max (\abs{ \frac{B -  \sum_i r_i \overline{w}_i }{ \min_t \overline{\Lambda}_t}}, \abs{-B -  \sum_i r_i \underline{w}_i} ) \leq \frac{e 2 \nu^{-1} B \Gamma}{ \min_t {\Lambda}_t \wedge 1 } .$$ Applying this bound on $\eta^*$: 
\begin{align*}& 2  B \Gamma \nu^{-1} ( \max_t \nicefrac{1}{p_t} \frac{1}{n}  \suma  {\Lambda_t}  +  \underbrace{\max_{w \in \overline{S}(\ccpscalar^*)} \suma  (\sum_{i \in \mathcal{I}_t }\abs{ {w_i}- \ccpscalar_{t} \tilde W_i }   - 	\E[ \fabs{ w(T,X,Y) - \tilde W(T,X)\ccpscalar_{t} }\mid T=t ] )}_{  \textcircled{\raisebox{-0.9pt}{1}} } )
\end{align*} 
It remains to study uniform convergence of $\textcircled{\raisebox{-0.9pt}{1}}$ when we optimize over $w$ in the set of restricted complexity (recall that monotonicity over $u$ is equivalent to monotonicity over $\tilde{u}$; or we may equivalently reparametrize in $W$ for the fixed scaling $\ccpscalar$). We do so by a Lipschitz contraction argument and applying our tail inequality from \Cref{lemma-weight-uc-pi}. Note that the absolute value function is globally $1$-Lipschitz; and the envelope function on $W(u)$ is bounded by $(b-a) u  \leq  \nu^{-1} (\Gamma - \Gamma^{-1}) $. Now, by Lipschitz contraction (Theorem 5.7 of \cite{pollard1990empirical}), applying \Cref{lemma-weight-uc-pi}, and taking a union bound over the number of treatments, we obtain the final bound that, with high probability $\geq 1 -p$, 
\begin{align*}
& \hat{\overline R}_{\pi_0}(
\pi;\mathcal W^{\Gamma,\Lambda}(\mathbb P_n)) - \hat{\overline R}_{\pi_0}(
\pi;\mathcal W^{\Gamma,\Lambda}(\mathbb P)) \\
& \leq  \frac{2 \nu^{-1} B \Gamma}{ \min_t {\Lambda}_t \wedge 1 } \left(  \frac{\max_t \nicefrac{1}{p_t}\suma  {\Lambda_t}}{n}   + 18 m K^{\Pi} \nu^{-1}(\Gamma - \Gamma^{-1})  \sqrt{ \frac{ { \log(\nicefrac{5m}{p})} }{ {n} } }  \right) 
\end{align*} 

The result follows by applying this bound twice, at the sample-optimal and population-optimal policies, and taking a union bound over the event of this bound holding with high probability and the previous minimax regret bound of \Cref{unifconv}, and the triangle inequality. 
\qed
\endproof
\subsection{Proof of \Cref{lemma-lipschitzness}.}
\proof{Proof of \Cref{lemma-lipschitzness}.}
{
	In the following, we first consider the optimization problem within a single treatment partition, reindexing $i=1, \dots, n$ to enumerate the elements of a generic treatment partition. The lemma follows by applying the same analysis to each treatment partition separately. 
	We aim to bound the approximation error incurred by optimizing over an uncertainty set derived from \textit{estimated} propensities, $\hat e_t(X)$ which may differ from the oracle values $\tilde{e}_t(X)$. 
}
\blockedit{
	Recall the weight bounds derived from the oracle nominal propensities, with $\tilde{W}= \nicefrac{1}{\tilde{e}_t(x)}$, are $ a = 1 + \frac{1}{\Gamma}(\tilde{W} - 1), b = 1 + {\Gamma}(\tilde{W} - 1)$; for this section, we define $\delta^a_i, \delta^b_i$ as the \textit{perturbations} of the sample weights from the oracle bounds $a, b$: $$\hat{\delta}_i^{a,\Gamma} = 1 + \frac{1}{\Gamma} \left( \nicefrac{1}{\hat{e}_{T_i}(x)} - 1\right) - a_i, \hat{\delta}_i^{b,\Gamma} = 1 + {\Gamma} \left( \nicefrac{1}{\hat{e}_{T_i}(x)} - 1\right) - b_i$$
	
	Observe that the dual of the primal program,\begin{equation}\label{lemma-boundedperturbations-primal}\sup_w \left\{ \sum_i w_i y_i :  \ccpscalar
	\cdot(a_i+\delta^a_i )\leq w_i \leq \ccpscalar \cdot(b_i+\delta^b_i), \; \sum_i w_i =1   \right\}  
	\end{equation}  for a fixed $\ccpscalar$ scaling, and a generic multiplier $r$, is the program: 
	\begin{equation}\label{lemma-boundedperturbations-primal} \inf_{\lambda, u\geq 0, v\geq 0} \left\{\lambda + \ccpscalar \cdot \left( -\sum_i (a_i+\delta^a_i) u_i + \sum_i (b_i+\delta^b_i) v_i  \right) : \lambda - u_i + v_i \geq _i, \forall i = 1... n  \right\} 
	\end{equation}
	and since $u, v \geq 0$, we again observe (as in the proof of \Cref{thm-norm-wghts-soln}) that by complementary slackness, $v = (r_i - \lambda)_+, u = (\lambda - r_i)_+$. We make the corresponding substitution and proceed to define the partial Lagrangian relaxation. Denote $g_{\delta_a, \delta_b}(\ccpscalar;\lambda,u,v)$, as the \textit{objective} function with given $\ccpscalar$, and $\delta^a,\delta^b$ perturbations to the weights:  
	$$\inf_{\lambda, u\geq 0, v\geq 0} g_{\delta^a, \delta^b}(\ccpscalar;\lambda,u,v) =  \inf_{\lambda, u\geq 0, v\geq 0} \left\{\lambda + \ccpscalar\cdot \left( -\sum_i (a_i+\delta^a_i) (\lambda - r_i)_+ + \sum_i (b_i+\delta^b_i)(r_i - \lambda)_+ \right) \right\} $$
	As a consequence of Lemma~\ref{lemma-supdiffsup}, $ \abs{\inf f - \inf g} \leq \sup\abs{f -g }$. Furthermore, we may optimize over the restrictions of the dual variables to compact sets: since $\lambda$ is a quantile of the coefficients, $\lambda \in [ \min_i r_i, \max_i {r_i}]$. We also have that $ \ccpscalar \in [\frac{1}{n}, \frac{1}{\nu n} ]$ under Assumption~\ref{asn-overlap} (strong overlap), and the constraint that $\sum_i w_i = 1$, so that 
	$\ccpscalar \cdot \sum_i (b_i+\delta^b_i) \leq \sum_i w_i  \leq \ccpscalar \cdot \sum_i (a_i+\delta^a_i) $.
	
	We invoke strong LP duality which holds with bounded optimal value (assuming bounded outcomes); strict feasibility and boundedness implies that the problem cannot be primal infeasible or primal unbounded. Let $\{\ccpscalar_{a,b}^*,(\lambda_{a,b}^*,u_{a,b}^*, v_{a,b}^*)\} \in \arg\min g_{\delta^a, \delta^b}(\ccpscalar;\lambda,u,v)$. Therefore, compactness of the feasible region gives that the optimal primal and dual variables, $\{\ccpscalar_{a,b}^*,(\lambda_{a,b}^*,u_{a,b}^*, v_{a,b}^*)\}$, $ \{ \ccpscalar_{0,0}^*,(\lambda_{0,0}^*,u_{0,0}^*, v_{0,0}^*) \}$, are also pairs of optimal best responses for the min/max partial Lagrangian duals of the perturbed and nominal problem. In the following, let $ S = \{ (\lambda, u, v) : \lambda\in[-B,B], u \in[0,2B], v\in [0,2B] \} $ denote the compact restriction. 
	\begin{align*}
	&\abs{ \overline{R}_{\pi_0}(\pi, \tilde{\mathcal{W}} ) -\overline{R}_{\pi_0}(\pi, \hat{\mathcal{W}})  }
	=\abs{ \sup_{\ccpscalar> 0}  \left\{ \inf_{\lambda, u \geq 0, v \geq 0 } g_{\delta^a, \delta^b}(\ccpscalar;\lambda,u,v)\right\} - \sup_{\ccpscalar' > 0}\left\{ \inf_{\lambda, u \geq 0, v \geq 0 } g_{00}(\ccpscalar';\lambda,u,v) \right\} } \\
	& = \abs{  \inf_{ (\lambda, u, v) \in S } \left\{ \sup_{\ccpscalar\in [\frac{1}{n}, \frac{1}{\nu n}]} g_{\delta^a, \delta^b}(\ccpscalar;\lambda,u,v)\right\}
		-\inf_{  (\lambda', u', v') \in S  } \left\{   \sup_{ \ccpscalar' \in[\frac{1}{n}, \frac{1}{\nu n}] }g_{00}(\ccpscalar';\lambda',u',v') \right\} } \numberthis \label{lemma-boundedperturbations-eqn1} \\
	& = \abs{    \sup_{\ccpscalar\in [\frac{1}{n}, \frac{1}{\nu n}]} \left\{\inf_{ (\lambda, u, v) \in S  } g_{\delta^a, \delta^b}(\ccpscalar;\lambda,u,v)\right\}
		- \sup_{ \ccpscalar' \in[\frac{1}{n}, \frac{1}{\nu n}] }\left\{ \inf_{ (\lambda', u', v') \in S  }  g_{00}(\ccpscalar';\lambda',u',v') \right\} } 
	\numberthis \label{lemma-boundedperturbations-eqn2} \\
	&  \leq  \sup_{\ccpscalar\in  {\{\ccpscalar_{00}^*, \ccpscalar_{ab}^*\}} } \abs{ \inf_{ (\lambda, u, v) \in S  } g_{\delta^a, \delta^b}(\ccpscalar;\lambda,u,v)
		-  \inf_{ (\lambda', u', v') \in S  }  g_{00}(\ccpscalar;\lambda',u',v')  }
	\numberthis \label{lemma-boundedperturbations-eqn3}
	\\
	& \leq \max_{j,k \in \{00, ab\} } \abs{ g_{\delta^a, \delta^b}( \ccpscalar_j^*; \lambda_k^*, u_k^*, v_k^*) -  g_{00}( \ccpscalar_j^*; \lambda_k^*, u_k^*, v_k^*)  } 
	\numberthis \label{lemma-boundedperturbations-eqn4}
	\end{align*}
	In the above, the equality of \Cref{lemma-boundedperturbations-eqn1} follows since without loss of generality, we can restrict attention to bounded feasible regions for the variables. In \Cref{lemma-boundedperturbations-eqn2}, we swap the order of the sup and inf since strong duality holds with equality. In \Cref{lemma-boundedperturbations-eqn3}, restricting the supremum over $\ccpscalar$ to the best responses $\ccpscalar_{00}^*, \ccpscalar_{ab}^*$ doesn't change the optimal value; that $\lambda^*, u^*, v^*$  and $\ccpscalar^*$ are best responses is a consequence of von Neumann's minimax theorem, since g is bilinear in its arguments $\ccpscalar$ and $\lambda, u, v$.
	\Cref{lemma-boundedperturbations-eqn4} holds since $\lambda^*_{0,0}, u^*_{0,0}, v^*_{0,0}$ {were optimal for} $g_{0,0}$ (resp., for $g_{\delta^a, \delta^b}$) and we expand the feasible set.
	
	Combining $g_{\delta^a, \delta^b}$ and $g_{0,0}$, we can now bound the perturbation incurred based on possible values of $\ccpscalar^*, \lambda^*$: 
	\begin{align*}
	& =    \max_{j,k \in \{00, ab\} } \abs{  \ccpscalar_j^* \cdot \left( -\sum_i \delta^a_i (\lambda_k^* - r_i)_+ +  \sum_i \delta^b_i (r_i - \lambda_k^*)_+\right) } \\
	& \leq  \max_{\ccpscalar \in \{\ccpscalar_{a,b}^*, \ccpscalar_{0,0}^* \} } \ccpscalar\cdot  (\norm{\delta^a}_1 + \norm{\delta^b}_1)  (2\max_i r_i ) \text{ since the optimal } \lambda^* \text{ is bounded} \\
	& \leq \frac{ 2\max r_i (\norm{\delta^a}_1 + \norm{\delta^b}_1) }{n} \\
	&= (\max_i Y_i ) (\Gamma + \nicefrac{1}{\Gamma}) \frac{1}{n}\sum_i \abs{\frac{1}{\hat{e}_t(X_i)}- \frac{1}{\tilde{e}_t(X_i)} } %
	\end{align*}
	The bound on the range for $\ccpscalar$ follows since for $\ccpscalar \in  \{\ccpscalar_{a,b}^*, \ccpscalar_{0,0}^* \}$, we have that $\ccpscalar \leq \max\{ \frac{1}{\sum_i (a_i + \delta^a_i) }, \frac{1}{\sum_i a_i  }  \} \leq \frac{1}{n}   $ since the bounds $\alpha_i+a_i$ and $ \alpha_i$ are inverse probabilities. 
}
{ 
	We simply apply the above argument for each group, under the product uncertainty set assumption. Define the treatment-conditional partial dual objective, computed for data from treatment partition $T=t$, as $g_{\delta^a, \delta^b}(\ccpscalar;\lambda,u,v;t)$. We apply the above bound for every treatment partition $T=t$, which holds \textit{deterministically} for all $\pi$, with the multiplier $r = (\pi - \pi_0 )Y$.
	\begin{align*}
	&	\hat{\overline{R}}_{\pi_0}(\pi, \tilde{\mathcal{W}}^\Gamma_n)  -\hat{\overline{R}}_{\pi_0}(\pi, \hat{\mathcal{W}}^\Gamma_n)  
	\\&=
	\abs{ \sup_{W\in \mathcal{W}^\Gamma(\tilde{e}_T)}\suma \frac{ \E_n[   (\piAX-\piAXbaseline) YW   \ttmentindic]}{ \E_n[ W  \ttmentindic ]}  - \sup_{W\in \mathcal{W}^\Gamma(\hat{e})}
		\suma \frac{ \E_n[   (\piAX-\piAXbaseline) YW   \ttmentindic ]}{ \E_n[ W \ttmentindic]} }\\
	&\leq  \suma \abs{\sup_{W(\cdot,\cdot;t)\in \mathcal{W}^\Gamma_t(\tilde{e}_T)}\frac{ \E_n[   (\piAX-\piAXbaseline) YW   \ttmentindic]}{ \E_n[ W  \ttmentindic ]}  - 
		\sup_{W(\cdot,\cdot;t)\in \mathcal{W}^\Gamma_t(\hat{e})}
		\frac{ \E_n[   (\piAX-\piAXbaseline) YW   \ttmentindic ]}{ \E_n[ W \ttmentindic]} }\numberthis \label{lemma-boundedperturbations-eqn-} \\
	&= \suma
	\abs{ \sup_{ \ccpscalar > 0}  \left\{ \inf_{\lambda, u \geq 0, v \geq 0 } g_{\delta^a, \delta^b}(\ccpscalar;\lambda,u,v;t)\right\} - \sup_{\ccpscalar' > 0}\left\{ \inf_{\lambda, u \geq 0, v \geq 0 } g_{00}(\ccpscalar';\lambda,u,v; t) \right\} }
	\\
	&= 2B (\Gamma + \nicefrac{1}{\Gamma}) \frac{1}{n}\sum_{i=1}^n \abs{\frac{1}{\hat{e}_{T_i}(X_i)}- \frac{1}{\tilde{e}_{T_i}^*(X_i)} }
	\end{align*}
	Here, \ref{lemma-boundedperturbations-eqn-} follows by the product set structure of the uncertainty set and application of the triangle inequality.
}
\endproof
\subsection{Proof of \Cref{thm:reformgenpolicy}}

\proof{Proof of \Cref{thm:reformgenpolicy}.}
Given convex $\mathcal S\subseteq\R n$, notice that its conic hull is $K=\{\sum_{i=1}^k\alpha_iu_i\colon k\in\mathbb N,\alpha_i\geq0,u_i\in\mathcal S\}=\bigcup_{\psi\geq0}(\psi\mathcal S)$. Let $r\in\R n$. Given that $\mathcal S$ has a non-empty interior, a Charnes-Cooper transformation followed by strong duality yields
\begin{align*}
\sup_{u\in\mathcal S}\frac{\sum_{i=1}^nr_iu_i}{\sum_{i=1}^nu_i}
&=\sup_{\substack{u/\psi\in\mathcal S,\,\psi\geq0,\\\sum_{i=1}^nu_i=1}}{\sum_{i=1}^nr_iu_i}=\sup_{\substack{u\succeq_{K}0,\\\sum_{i=1}^nu_i=1}}{\sum_{i=1}^nr_iu_i}=\inf_{\substack{\lambda\succeq_{K^*}r}}\lambda.
\end{align*}
The statement of the proposition proceeds by applying this for each treatment level $t$.
\qed\endproof

\section{Optimization Algorithm Details}
\subsection{Subgradient Approach Refinements}\label{apx-opt-refinements}
{
	We describe some additional changes to the subgradient method optimization procedure of \ref{sec-subgrad} which improve the optimization by specializing to the unique case of our problem. Further refinements are possible with e.g. homotopy methods for LPs; we leave this to future work. 
	In the case that we are optimizing over a series of increasing $\Gamma$ parameters, 
	$1=\Gamma_0<\Gamma_1<...<\Gamma_m$,
	we can use the nested property of the corresponding uncertainty sets to provide additional checks on the optimization. 
	\begin{enumerate}
		\item We include a warm start for optimization for $\Gamma_{k+1}$ with $\Gamma_k$ as one of the random initializations: therefore we are guaranteed an initialization that does well for similar $\Gamma$. 
		\item For each proposed optimal policy returned by the optimization, which we denote as $\bar{\pi}(\Gamma_k)$ for a policy optimized over $W_n^{\Gamma_k}$, we
		check the achieved objective value of previous policies, $\bar{R}(\bar{\pi}(\Gamma_k), \Gamma_i), i < k$. If for some $i$,  $\bar{R}(\bar{\pi}(\Gamma_k), \Gamma_i) < \bar{R}(\bar{\pi}(\Gamma_k), \Gamma_k)$, we set the policy to the previous policy, $\pi_k^* = \pi_i^*$.
	\end{enumerate}
	We find empirically that including these refinements stabilizes the optimization when optimizing over a nested series of $\Gamma$ parameters, as we anticipate a decision-maker would do in practice, given a feasible range of plausible $\Gamma$ values. 
}
\subsection{Optimal Confounding-Robust Trees} 
We next consider the function class consisting of axis-aligned decision tree policies where each leaf is assigned a constant probability of treatment. 
Decision tree policies are advantageous due to their simplicity and interpretability. Our optimal confounding-robust tree (OCRT) presented below determines the best confounding-robust decision tree via global optimization using mixed-integer optimization. 
Our approach is to combine the dual linear program formulation of $\hat{\overline R}_{\pi_0}(\pi;\mathcal W_n^\Gamma)$ in eq.~\eqref{pbm-dual} with a mixed-integer formulation of this class of decision trees, following the formulation of \citet{bd17}, along with a special heuristic to find a good warm start.

A decision tree (with maximal splits) of a fixed depth $D$ can be represented by an array labeled by a set of nodes, split into a set of branching nodes $\mathcal{K}_B$ and leaf nodes $\mathcal{K}_L$. The space of decision tree policies is parametrized by $\Theta = \{  \{\alpha_{k_b}, \beta_{k_b}\}_{k_b \in \mathcal{K}_B}, \{ c_k \}_{k\in \mathcal{K}_L} \}$, where $\alpha_{k_b}, \beta_{k_b} \in \mathbb{R}^p$ parametrize the split at branching node $k_b$, which directs units to the left branch if $\alpha^\intercal x < \beta$, and to the right branch otherwise. The policy assignment probability is parametrized by $c_{k_b}  \in [0,1]$ for $k_b\in \mathcal{K}_L$. We consider axis-aligned splits such that $\alpha_{k_b}$ is a unit vector. 

We let the binary assignment variables $z_{ik}$ track assignment of data points $i$ to leaves $k \in \mathcal{K}_L$ subject to the requirement that every instance is assigned to a leaf node according to the results of axis-aligned splits $\alpha_{k_b}^\intercal x< \beta_{k_b}$, for splits occurring at $k_b \in \mathcal{K}_B$ branch nodes. The binary variables $d_k$ track whether a split occurs at node $k_b \in \mathcal{K}_B$. The binary variable $l_k$ tracks whether a leaf is empty or not. The policy optimization determines both the partitions of the covariates 
governing assignment to terminal leaf nodes and the variables $c_k$ for $k\in \mathcal{K}_L$ governing probability of treatment assignment in the leaf nodes. We denote $\text{par}(k)$ as the parent of node $k$, $A(k)$ as the set of all ancestors of node $k$, and the subsets $A_L(k) \cup A_R(k) = A(k)$ denote the sets of ancestor nodes where the instance was split to the left or right, respectively. In this section, we assume that the covariates are rescaled such that each covariate lies in $[0,1]$. 
{
	We introduce additional constraints to encode our dual objective in the optimal classification tree framework. We define the policy assignment probability for treatment $T=t$, $P_i^t
	= \sum_{k \in {\mathcal{K}}_L} z_{ik} c_k^t $ where $c_k^t$ is the policy assignment probability of leaf node $k \in {\mathcal{K}}_L$ of assigning treatment $t$, and $z_{ik}$ describes whether or not instance $i$ is assigned to leaf node $k$, enforced with the additional set of auxiliary big-$M$ constraints for the product of a binary variable and continuous variable; for each set of such product variables $P_i^t$.
}
{
	\begin{align*} &  \quad p_{i,k}^t \leq z_{ik}; \quad p_{i,k}^t \leq c_k^t; \quad p_{i,k}^t \geq c_k^t + z_{ik} - 1 & \; \forall i = 1,\dots,n; \forall t \in  \mathcal{T} ,  k \in \mathcal{K}_L\\
	&  P_i^t = \sum_{k \in {\mathcal K }_L} p_{i,k}^t &  \forall a \in \mathcal A, \; \forall i = 1,\dots,n\\
	& \suma c_k^t = 1&  k \in \mathcal{K}_L\\
	& p_{i,k}^t\in [0,1] &  \forall t \in  \mathcal{T}, \; \forall i =1,\dots,n, k \in \mathcal{K}_L \\
	& c_k^t \in [0,1] &  \forall t \in  \mathcal{T}, \; \forall k \in \mathcal{K}_L\\
	& P_i^t \in [0,1] & \forall t \in  \mathcal{T}, \;  \forall i=1,\dots,n
	\end{align*}
}
The combined formulation for policy optimization with confounding-robust optimal trees is as follows: 
{
	\begin{subequations}
		\begin{alignat}{2}
		\min & \; \suma  \lambda_{t}  &\\ 
		\text{s.t.  }&v_i - u_i + \lambda_{T_i} \geq Y_i (P_i^{T_i} - \pi_0^{T_i}) & , \forall i \in \mathcal{I}_t \label{prob-optt-dual1} \\
		& \sum_{ i\in \mathcal{I}_t }- b^\Gamma_i v_i + a^\Gamma_i  u_i \geq 0, 
		&\forall   t \in \mathcal{T} \label{prob-optt-dual2}&\\
		&  \quad p_{i,k}^t \leq z_{it}; \quad p_{i,k}^t \leq c_k^t; \quad p_{i,k}^t \geq c_k^t + z_{ik} - 1 & \forall  t \in \mathcal{T}, \; \forall i = 1,\dots,n, k \in \mathcal{K}_L, \label{prob-optt-pol2}\\
		&  P_i^t = \sum_{k \in \mathcal{K}_L} p_{i,k}^t &  \forall t \in \mathcal{T}, \; \forall i = 1,\dots,n \label{prob-optt-pol1}\\
		& \suma c_k^t = 1& k \in \mathcal{K}_L\\
		& c_k^t \in [0,1] &  \forall t \in \mathcal T, \; \forall k \in \mathcal{K}_L\\
		&a_{m}^\intercal( x_i + \epsilon) \leq b_{m} + (1-z_{ik}) & \forall i =1,\dots,n, \forall k \in \mathcal{K}_B, \forall m \in A_L(k) \label{prob-optt-splleft} \\
		&a_{m}^\intercal (x_i+\epsilon) \leq b_{m} - (1+\epsilon_{max})(1-z_{ik})\quad& \forall i = 1,\dots,n,  \forall k \in \mathcal{K}_B, \forall m \in A_R(k) \label{prob-optt-splright}\\ 
		& {\textstyle\sum_{k \in \mathcal{K}_L}} z_{ik} = 1 \label{prob-optt-zleaf}&\forall k \in \mathcal{K}_B\\
		& {\textstyle\sum_{i=1}^n z_{ik}} \geq N_{min} l_t & \forall  i = 1,\dots, n \label{prob-optt-minleaf} \\
		& {\textstyle\sum_{j=1}^p} a_{jt} = d_t \label{prob-optt-splitindic} &\\
		& 0 \leq b_k \leq d_k & \forall  k \in \mathcal{K}_B \label{prob-optt-splitindiccons} \\&  d_t \leq d_{\text{par}(k)}& \forall k \in \mathcal{K}_B \setminus \{1\} \label{prob-optt-splitcons} \\
		& l_{U(k)} \geq d_{(\text{par}(k))} & k \in \mathcal{K}_B \setminus 1 \label{prob-optt-nonemptyleafifsplit} \\
		& l_{k} \leq d_{\text{par}(m)}& \forall m \in \mathcal{T}_B, t \in[D(k_b), U(k_b)] \label{prob-optt-splitleaf-consistency}\\
		&  l_{k} \geq d_{\text{par}(t)}  & \forall k \in \mathcal{K}_L   \label{prob-optt-splitleaf-consistency-2}\\
		&z_{ik}, l_k \in \{0,1\} &i = 1,\dots,n , \forall k \in \mathcal{K}_L \\
		& a_{jk}, d_k \in \{0,1\}& j=1,\dots,p, \forall k \in \mathcal{K}_B\\
		& p_{i,k}^t\in [0,1] &  \forall t \in  \mathcal{T}, \; \forall i =1,\dots,n, k \in \mathcal{K}_L \\
		& c_k^t \in [0,1] &  \forall t \in  \mathcal{T}, \; \forall k \in \mathcal{K}_L\\
		& P_i^t \in [0,1] & \forall t \in  \mathcal{T}, \;  \forall i=1,\dots,n\\
		&  u, v\geq 0&
		\end{alignat}
	\end{subequations}
}
{ 
	Constraints (\ref{prob-optt-pol1}, \ref{prob-optt-pol2}) set the policy assignment variable $P^{t}_i \in [0,1]$, which is the sum of products $p_{i,k}^t=z_{ik} c_k $ over leaf nodes. Our objective is specified via the dual formulation, and constraints (\ref{prob-optt-dual1}, \ref{prob-optt-dual2}) encode the constraints from the dual of the inner maximization subproblem. Constraints (\ref{prob-optt-splleft}, \ref{prob-optt-splright}) enforce that if a node is in a leaf (as indicated by $z_{ik}$), it satisfies the splits at ancestor nodes. Constraint (\ref{prob-optt-zleaf}) enforces that each instance is in a leaf node, while constraint \ref{prob-optt-minleaf} enforces a size constraint on leaf membership. }Constraints (\ref{prob-optt-splitindic}, \ref{prob-optt-splitcons}, \ref{prob-optt-splitindiccons}) enforce consistency constraints between $d$, indicating whether a split occurs at leaf node $k$, and split variables $a_{jk}, b_{k}$. $\{D(k)\}_{k\in \mathcal{K}_B}$ denotes the set of leaf nodes of smallest index which can be reached from splits at $k$, and similarly $\{U(k)\}_{k\in \mathcal{K}_B}$ denotes the set of reachable leaf nodes of largest index. Constraints (\ref{prob-optt-nonemptyleafifsplit}, \ref{prob-optt-splitleaf-consistency}, \ref{prob-optt-splitleaf-consistency-2}) enforce that leaves are non-empty only if splits do occur in the relevant ancestor nodes.

For the mixed-integer linear program, we provide a warm start for the optimization via a recursive partitioning-based approach which incrementally optimizes directly the robust risk, over iterative refinements of either the constant all-treat or all-control policy, described in Sec.~\ref{algos-trees} of the EC.
\subsection{Recursive Partitioning: MIP Warm Start}\label{algos-trees}
We provide a heuristic recursive-partitioning based scheme for optimizing policy risk over the space of limited-depth decision trees recursively, analogous to CART's recursive partitioning approach \citep{cart84}. Such an approach is used to obtain a warm start for the MIP of the optimal confounding-robust tree. The algorithm initializes by assigning the same treatment $\tau_0$ to all, and iteratively refines the treatment assignment by recursive partitioning, seeking univariate splits which minimize the minimax risk. The candidate split threshold for each covariate is determined by iteratively re-evaluating the minimax risk for incremental changes to the policy, maintaining the invariant that the base policy is set by the leaves above a node in the tree. Using specialized data structures such as B-trees allows for $O(log(N))$ efficient updates for maintaining and updating the sorted list of multipliers $Y_iT_i(\pi_i-\pi_0)$, and manipulating pre-computed cumulative sums of the initial sorted order allows for efficient re-computation of the optimization solution. We note that such an approach is possible only for the unbudgeted uncertainty set $\mathcal{U}_n^\Gamma$, since incorporating the uncertainty budget would couple the risk across tree levels. 
		\begin{algorithm}[t!]
        \caption{ Greedy Recursive Partitioning (\textbf{Partition})}
        \label{alg-greedy-rec-part}
	\begin{algorithmic}[1]
		\State Input: partition $S_{L,l} = \{ (X_{i_1},T_{i_1},Y_{i_1}) \}$, depth $\Delta$, preliminary assignment $\tau^{\Delta-1} \in[m]^n$
		\For{$d \in [p]$} \text{(find best partition index)}: 
		\State $[i] \gets$ Get the sorted indices of ($\{X_{i,d}\}$)
		\State  $i_j^*, v_j^* \gets$ Find the best dimension and threshold to split $x_{j^*} < x_{i_j^*, j^*}$
		\State  $i_{j,rev}^*, v_{j,rev}^* \gets$ Find the best dimension and threshold to split $x_{j^*} > x_{i_j^*, j^*}$
		\EndFor
		\State $j^* \gets \argmin_j i^*_j,$ $\quad i^* \gets  i^*_{j^*}$, $\quad\theta^* \gets \frac{X_{(i^*),j^*}+X_{(i^*+1),j^*}}{2}$
		\State $\pi(X)$ $\gets$ $x_{j^*} \leq \theta^*$ if $v_j^* < v_{j,rev}^*$ else $x_{j^*} \geq \theta^*$ 
		\If{(continue recursing)}: 
		\State $S_L \gets X_{[0:i^*]}, S_R \gets X_{[i^*:\vert S \vert ]}$
		\State update $\tau_0$, the candidate treatment assignment
		\State $\hat{\Pi}_L \gets$ \textbf{Partition}$(S_L, \tau', \Delta+1)$, $\hat{\Pi}_R \gets$ \textbf{Partition}$(S_R, \tau_0, \Delta+1)$
		\EndIf 
		\Return ($\pi(X)$ , $\hat{\Pi}_L$, $\hat{\Pi}_R$ )
	\end{algorithmic}
\end{algorithm}
In comparison to other approaches using tree-based approaches for estimating causal effects \citep{wager2017estimation} or for personalization \citep{k16}, which consider splits based on impurities related to the expected mean squared error of causal effects on a separate sample of data from that used to estimate the causal effects within leaves, or determine the optimal treatment, our recursive partitioning heuristic simultaneously determines the partition and the policy treatment assignment within the partition. In making greedy splits, changes in the objective function are assessed as a result of changing the policy assignment within $S_{L, l}$, and the optimal split location and sense ($\mathbb{I}[ a^\intercal X < b ]$ or $\mathbb{I}[ a^\intercal X > b ]$) are determined by changes in the policy assignment within $S_{L, l}$. However, in general, the optimal such policy assignment, determined incrementally from the assignments $\{ S_{L-1, l}\}_{l \in \mathcal{T}_L}$, depends additionally on the assigned policy for other nodes at the same level as well. 
\section{WHI Case Study details}\label{apx-WHI}
{ 
	\begin{table}
		\setlength{\tabcolsep}{5pt}
		\centering
\caption{Policy regret for WHI, under different $\lambda$ scalarizations}
	\begin{adjustbox}{angle=270}
		\setlength{\tabcolsep}{3pt}
		\label{tbl-whi-lambdas-crlogit}
		\begin{tabular}{rrrrrrrrrrrrrrrrrrrrr}
			\hline
			$\lambda$ &   $\substack{\log(\Gamma) \\= 0.05}$ &   0.075 &   0.1 &   0.12 &   0.14 &   0.16 &   0.18 &   0.2 &   0.225 &   0.25 &   0.275 &   0.3 &   0.325 &   0.35 &   0.4 &   0.45 &   0.5 &   0.75 &   1.0 &   2.0 \\
			\hline
			-0.50 &                    0.26 &    0.07 & -0.15 &  -0.10 &   0.07 &   0.10 &   0.08 &  0.06 &    0.04 &   0.02 &    0.01 &  0.01 &    0.02 &   0.02 &  0.03 &   0.05 &  0.07 &   0.08 &  0.06 &  0.00 \\
			-0.57 &                    0.59 &    0.39 &  0.00 &   0.00 &   0.00 &   0.00 &   0.00 &  0.00 &    0.00 &   0.00 &    0.00 &  0.00 &    0.00 &   0.00 &  0.00 &   0.00 &  0.00 &   0.00 &  0.00 &  0.00 \\
			-0.64 &                    0.33 &    0.33 &  0.13 &   0.05 &  -0.00 &   0.01 &   0.02 &  0.01 &   -0.01 &  -0.02 &   -0.03 & -0.03 &   -0.02 &  -0.01 &  0.01 &   0.03 &  0.05 &   0.00 &  0.00 &  0.00 \\
			-0.71 &                    0.33 &    0.12 & -0.16 &  -0.44 &  -0.45 &  -0.27 &  -0.04 &  0.11 &    0.11 &   0.11 &    0.11 &  0.11 &    0.11 &   0.11 &  0.11 &   0.11 &  0.11 &   0.09 &  0.06 &  0.00 \\
			-0.79 &                    0.86 &    0.66 &  0.18 &  -0.23 &  -0.40 &  -0.47 &  -0.51 &  0.00 &    0.00 &   0.00 &    0.00 &  0.00 &    0.00 &   0.00 &  0.00 &   0.00 &  0.00 &   0.00 &  0.00 &  0.00 \\
			-0.86 &                    0.39 &    0.35 &  0.16 &  -0.17 &  -0.41 &  -0.54 &  -0.62 & -0.67 &   -0.47 &  -0.09 &    0.05 &  0.04 &    0.04 &   0.05 &  0.07 &   0.09 &  0.10 &   0.08 &  0.05 &  0.00 \\
			-0.93 &                    0.24 &    0.23 &  0.25 &   0.08 &  -0.21 &  -0.40 &  -0.47 & -0.36 &   -0.13 &   0.02 &    0.07 &  0.08 &    0.07 &   0.05 &  0.03 &   0.02 &  0.01 &   0.00 &  0.00 &  0.00 \\
			-1.00 &                    0.18 &    0.19 &  0.13 &   0.28 &   0.06 &  -0.10 &  -0.42 & -0.44 &   -0.07 &   0.06 &    0.05 &  0.02 &    0.01 &   0.02 &  0.04 &   0.07 &  0.09 &   0.00 &  0.00 &  0.00 \\
			-1.07 &                    0.22 &    0.17 &  0.08 &  -0.02 &  -0.23 &  -0.32 &  -0.47 & -0.47 &   -0.15 &   0.03 &    0.05 &  0.02 &    0.00 &  -0.01 &  0.01 &   0.03 &  0.05 &   0.00 &  0.00 &  0.00 \\
			-1.14 &                    0.16 &    0.14 &  0.12 &   0.12 &  -0.24 &  -0.27 &  -0.19 & -0.12 &   -0.06 &   0.00 &    0.00 &  0.00 &    0.00 &   0.00 &  0.00 &   0.00 &  0.00 &   0.00 &  0.00 &  0.00 \\
			-1.21 &                    0.08 &    0.06 &  0.02 &   0.03 &   0.48 &   0.30 &   0.14 & -0.11 &    0.02 &   0.06 &    0.06 &  0.06 &    0.06 &   0.05 &  0.06 &   0.07 &  0.07 &   0.00 &  0.00 &  0.00 \\
			-1.29 &                   -0.01 &   -0.02 & -0.05 &  -0.09 &  -0.11 &  -0.12 &  -0.10 & -0.14 &   -0.19 &  -0.22 &   -0.22 & -0.21 &   -0.19 &  -0.17 &  0.00 &   0.00 &  0.00 &   0.00 &  0.00 &  0.00 \\
			-1.36 &                   -0.07 &   -0.09 & -0.12 &  -0.15 &  -0.23 &  -0.29 &  -0.16 & -0.08 &   -0.04 &  -0.05 &   -0.11 &  0.00 &    0.00 &   0.00 &  0.00 &   0.00 &  0.00 &   0.00 &  0.00 &  0.00 \\
			-1.43 &                   -0.18 &   -0.21 & -0.22 &  -0.22 &  -0.21 &  -0.21 &   0.21 &  0.33 &    0.44 &   0.51 &    0.50 &  0.45 &    0.38 &   0.00 &  0.00 &   0.00 &  0.00 &   0.00 &  0.00 &  0.00 \\
			-1.50 &                   -0.37 &   -0.37 & -0.37 &  -0.36 &  -0.32 &  -0.30 &  -0.30 & -0.19 &    0.02 &   0.09 &    0.07 &  0.03 &    0.01 &  -0.00 &  0.00 &   0.01 &  0.02 &   0.00 &  0.00 &  0.00 \\
			\hline
		\end{tabular}
	\end{adjustbox} 
	\end{table}
}

\paragraph{Calibration for WHI}
\begin{figure}\label{fig:whi-joyplot}
	\centering
\includegraphics{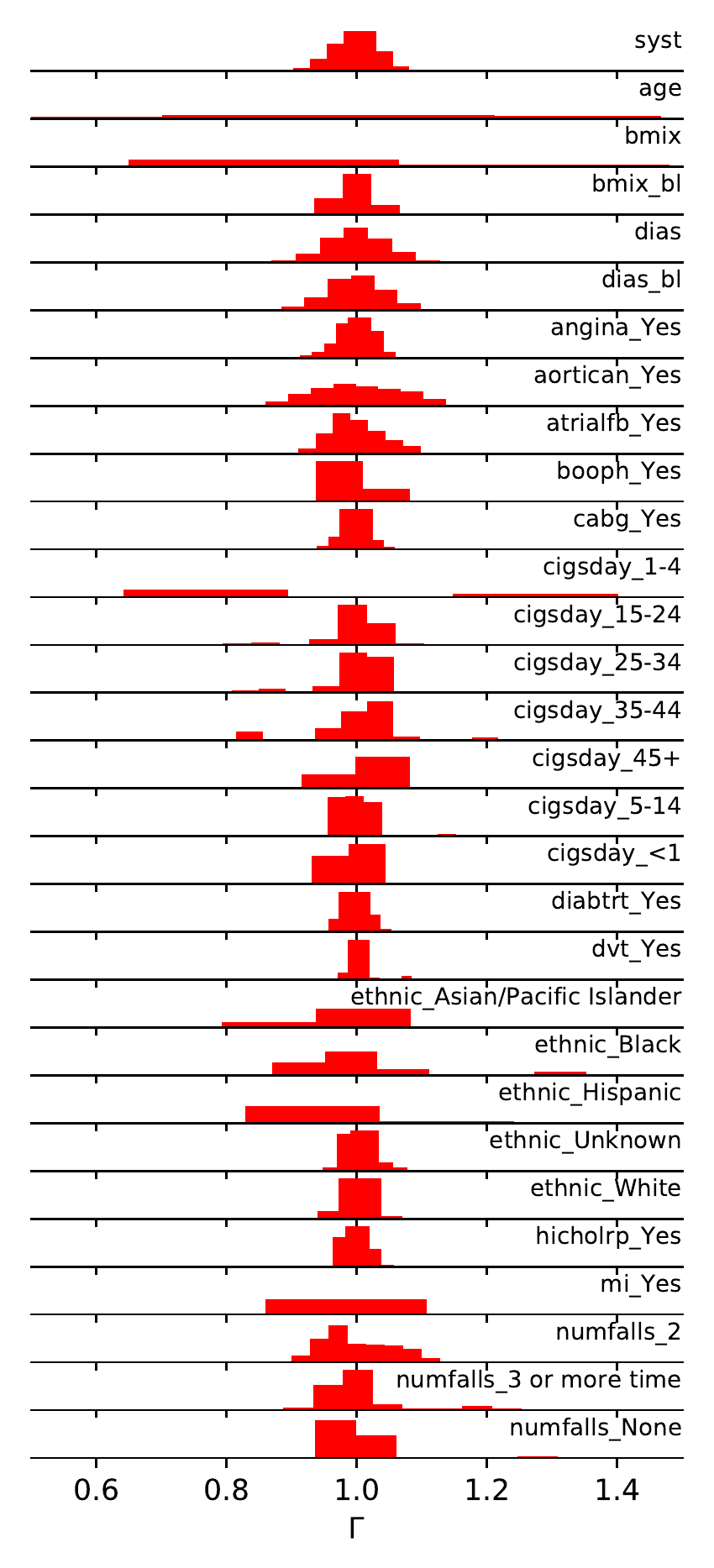}
\caption{ Density comparison of odds ratios induced by training propensities with dropped covariates (one per line, in order). x-axis is the odds ratio, while y-axis (for each subplot) is a density plot; fixed y-scale $y \in [0,10]$ for all subplots. Note that most of the probability mass is within $\Gamma\in [0.8, 1.2]$, with the exception of a few covariates with wider distributions of informativity.}
\end{figure}
\end{document}